\documentclass[preprint,12pt]{elsarticle}

\usepackage[utf8]{inputenc} 
\usepackage{geometry}
\usepackage{amsmath}
\usepackage{graphicx}
\usepackage{xcolor}
\usepackage{natbib}
\usepackage{media9}
\usepackage{tikz}
\usepackage{tcolorbox}
\usepackage{graphicx}
\usepackage[flushleft]{threeparttable}
\usepackage{float}
\usepackage{booktabs}
\usepackage{subcaption}

\DeclareMathOperator*{\argmax}{arg\,max}

\usepackage[font=small,skip=0pt]{caption}
\usepackage{multirow}

\usepackage{amssymb}
\usepackage{pifont}
\usepackage{array}
\newcolumntype{L}[1]{>{\raggedright\let\newline\\\arraybackslash\hspace{0pt}}m{#1}}
\newcolumntype{C}[1]{>{\centering\let\newline\\\arraybackslash\hspace{0pt}}m{#1}}
\newcolumntype{R}[1]{>{\raggedleft\let\newline\\\arraybackslash\hspace{0pt}}m{#1}}

\usepackage[T1]{fontenc}
\newcommand*{\dittotikz}{--\textquotedbl--}

\usepackage{setspace}

\usepackage{hyperref}
\hypersetup{
colorlinks={true},
linkcolor={Blue3},
urlcolor={Blue3},
citecolor={Blue3},
pdfstartview={FitH},
pdfauthor={Author},
pdftitle={Title},
pdfsubject={Subject}
plainpages=false
}

\newcommand{\cmark}{\ding{51}}%

\newcommand{\STAB}[1]{\begin{tabular}{@{}c@{}}#1\end{tabular}}

\newcommand{\tcolB}{\textcolor{black}}
\newcommand{\tcolK}{\textcolor{black}}
\newcommand{\rrw}{\rightarrow}

\usepackage{upgreek}
\newcommand{\indCourier}{c} 
\newcommand{\setCourier}{\mathcal{C}}
\newcommand{\indOrder}{o}
\newcommand{\indRestaurant}{e} 
\newcommand{\indCustomer}{u} 

\newcommand{\dimGridm}{m}
\newcommand{\dimGridn}{n}
\newcommand{\timePrepare}{\pi}
\newcommand{\timeDeliver}{\delta}
\newcommand{\timeDeliverPrev}{\uptau} 
\newcommand{\distance}{d}
\newcommand{\distanceAvail}{\tilde{\distance}}
\newcommand{\distanceDepot}{\mu}
\newcommand{\distanceRestaurant}{\eta}

\newcommand{\paramOrderDistn}{\lambda}
\newcommand{\discount}{\gamma}

\newcommand{\QValue}{Q}
\newcommand{\QNetParams}{\theta}

\newcommand{\setState}{\mathcal{S}}
\newcommand{\setAction}{\mathcal{A}}
\newcommand{\setReward}{\mathcal{R}}

\newcommand{\indState}{s}
\newcommand{\indStateBig}{S}
\newcommand{\indAction}{a}
\newcommand{\indActionBig}{A}
\newcommand{\indTime}{t}
\newcommand{\indRewardBig}{R}

\newcommand{\update}{Y}
\newcommand{\stepSize}{\upalpha} 
\newcommand{\ddqnUpdateIter}{C}
\newcommand{\indTransition}{i}
\newcommand{\ISWeight}{w}
\newcommand{\PERTransitionProb}{P}
\newcommand{\priority}{p}
\newcommand{\PERexponenta}{\alpha}
\newcommand{\PERexponentb}{\beta}
\newcommand{\PERBuffer}{M}

\newcommand{\actionAssign}{x}
\newcommand{\actionReject}{\bar{\actionAssign}}
\newcommand{\actionReturnDepot}{\nabla}
\newcommand{\actionMoveToRestaurant}{\Delta}

\newcommand{\funcReward}{r}

\newcommand{\ERMemory}{\mathcal{M}}
\newcommand{\stepsToUpdateTargetNet}{\mathcal{U}}
\newcommand{\batchSize}{\mathcal{B}}

\newcommand{\softUpdateParam}{\tau}

\newcommand{\indHour}{h}
\newcommand{\orderCountDaily}{N}
\newcommand{\paramOrderDistnPercentage}{\Lambda}

\def\tsc#1{\csdef{#1}{\textsc{\lowercase{#1}}\xspace}}
\tsc{WGM}
\tsc{QE}
\tsc{EP}
\tsc{PMS}
\tsc{BEC}
\tsc{DE}
\journal{Knowledge-Based Systems}

\begin{document}
\let\WriteBookmarks\relax
\def\floatpagepagefraction{1}
\def\textpagefraction{.001}

\title {A Deep Reinforcement Learning Approach for the Meal Delivery Problem}

\author[1]{Hadi Jahanshahi\corref{cor1}}
\ead{hadi.jahanshahi@ryerson.ca}


\address[1]{Data Science Lab, Ryerson University, Toronto, Ontario, M5B 2K3, Canada}

\author[1]{Aysun Bozanta}
\ead{aysun.bozanta@ryerson.ca}

\author[1]{Mucahit Cevik}
\ead{mcevik@ryerson.ca}

\author[2,3]{Eray Mert Kavuk}
\address[2]{Faculty of Computer and Informatics Engineering, Istanbul Technical University, Istanbul,
34467, Turkey}
\author[2]{Ay\c{s}e Tosun}
\author[3]{Sibel B. Sonuc}
\address[3]{Getir Perakende Lojistik A.S., Istanbul, 34337, Turkey}
\author[3]{Bilgin Kosucu}
\author[1]{Ay\c{s}e Ba\c{s}ar}

\cortext[cor1]{Corresponding author}

\begin{abstract}
We consider a meal delivery service fulfilling dynamic customer requests given a set of couriers over the course of a day.  
A courier's duty is to pick up an order from a restaurant and deliver it to a customer. 
We model this service as a Markov decision process and use deep reinforcement learning as the solution approach.
We experiment with the resulting policies on synthetic and real-world datasets and compare those with the baseline policies. 
We also examine the courier utilization for different numbers of couriers. 
In our analysis, we specifically focus on the impact of the limited available resources in the meal delivery problem. 
Furthermore, we investigate the effect of intelligent order rejection and re-positioning of the couriers. 
Our numerical experiments show that, by incorporating the geographical locations of the restaurants, customers, and the depot, our model significantly improves the overall service quality as characterized by the expected total reward and the delivery times.
Our results present valuable insights on both the courier assignment process and the optimal number of couriers for different order frequencies on a given day. 
The proposed model also shows a robust performance under a variety of scenarios for real-world implementation. 

\end{abstract}

\begin{highlights}
    \item Our MDP model for the courier assignment task characterizes on-demand meal delivery service.
    \item We tailor deep reinforcement learning algorithms to address the problem in a dynamic environment.
    \item We incorporate the notion of order rejection to reduce the number of late orders.
    \item We investigate the importance of intelligent repositioning of the couriers during their idle times.
\end{highlights}

\maketitle
\newpage

\section{Introduction}
Meal delivery is a rapidly growing business area and globally projected to reach \$200 billion in terms of gross revenue by 2025~\citep{Statista}. 
Especially in urban areas, many people may not find time to cook for a variety of reasons, including long working hours and heavy traffic, making them order takeout frequently. 
With the increase of such demand, meal delivery service, which started as facilitating orders from each restaurant's own web page, has led to the emergence of services such as UberEats, Doordash, and Grubhub that bring orders from restaurants to the customers.
Increasing demand over the years has enabled this market to thrive. 
The revenue growth of the platform-to-customer food delivery market has increased 27\% from 2019 to 2020~\citep{Statista}. 
The number of users is expected to amount to 965.8 million all around the world by 2024 in the platform-to-consumer delivery segment~\citep{Statista}.

In addition to the challenges due to increasing demand, meal delivery services also pose significant operational difficulties. 
Most delivery platforms guarantee their customers' short delivery times with the objective of serving fresh foods. 
Providing a reliable and fast service enables the company to attract more users, leading to revenue and market share growth. 
However, especially in the peak hours of the day, i.e., at noon or in the evening, timely delivery becomes a challenging task.
Accordingly, courier shifts and order assignments should be defined based on the expected number of orders and the couriers' working hours. 
Another concern is the increasing number of couriers, which can be a prohibitive operational cost item while operating such a company. 
Accordingly, it is important to determine the ideal number of couriers that should be ready for each hour of the day to fulfill the customers' demands while minimizing the cost at the same time. 

These operational challenges and increasing market growth have lead research communities to explore possible solutions for the meal delivery problem and its variants. 
In recent studies, different approaches have been presented for the meal delivery problem~\citep{Mao2019,  pinto2020network, reyes2018meal, Ulmer2020deadline, yildiz2019provably}.
In our study, we consider a limited number of available couriers fulfilling customer demands by picking up the meals from requested restaurants and delivering them to the customers as fast as possible. 
\textcolor{black}{Our overall objective for the meal delivery problem is to maximize total profit by minimizing the expected delays, postponing or declining orders with low/negative rewards, and assigning the orders to the most appropriate couriers. 
We apply various deep reinforcement algorithms to solve our model, many of which have not been considered for the meal delivery and courier routing problems. 
We conduct detailed numerical analysis with a varying number of couriers in the system to determine the ideal number for the working hours of the day.}

The main characteristics and novelty of our work can be summarized as follows.
\begin{itemize}
   \item The primary contribution of our study is to propose a novel MDP model for the meal delivery problem and showing its use in practical settings through our numerical analysis with the datasets obtained from the day-to-day operations of a meal delivery company.
   \item Our model enables examining system improvements that can contribute to profitability and service quality. 
   For instance, in a typical setting, when a courier delivers an order, if there is no new order in the system, then the courier directly returns to the depot. 
   In our proposed model, couriers may head to the restaurants instead of going back to the depot. 
   By doing this, we aim to make the couriers learn the restaurants with a higher probability of having a new order so that the delivery times are shortened.
  In addition, we include a reject action in the action space, which corresponds to choosing not to deliver an order that is deemed to be far away from the current courier locations. 
  This action allows trading immediate rewards with future rewards and helps us achieving higher expected total rewards.
  Furthermore, our model allows assigning multiple orders to the couriers provided that the long run return of that assignment is higher than other options. 
  This model specification has only been considered in a few other studies in the literature~\citep{reyes2018meal, Steever2019, yildiz2019provably}.

  \item We conduct numerical experiments with different numbers of couriers to determine the ideal number of couriers at each hour of a given day for the purpose of meeting the customer demands while minimizing the costs. 
  We also present the courier utilizations to explore the rate at which each courier is assigned a new order, which helps to observe workload distribution among couriers.
  This consideration has important practical implications and, to the best of our knowledge, has not been examined in previous studies on the meal-delivery problems. 

  \item Our detailed numerical study helps exploring the performances of eight deep Q-Networks (DQN) algorithms for the meal delivery problem. 
  We conduct our experiments using both synthetic and real-world datasets to better assess the generalizability of our results. 
  We also perform hyperparameter tuning to identify the ideal parameters for our proposed model and provide a detailed discussion of these parameters. 
  Accordingly, our observations contribute to future studies that apply DQN and its variants to other problems.
    
\end{itemize}

The outline of the rest of the paper is as follows. 
In Section~\ref{LR}, we present the related literature. 
We define our problem and solution methodology in Section~\ref{method}. 
Subsequently, the results are presented in Section~\ref{results}. 
The paper concludes with a summary, limitations, and future research suggestions in Section~\ref{conc}.

\section{Literature review} \label{LR}

We primarily review the recent studies on the meal delivery problem and its variants. 
We first discuss courier routing/ assignment literature and then other delivery tasks (e.g., delivery robots) that share similar characteristics. Afterwards, we discuss the methodologies used for these problems, specifically reinforcement learning and approximate dynamic programming.

\subsection{Courier routing and assignment}
There is a vast literature on courier routing and assignment.
\citet{reyes2018meal} studied the meal-delivery routing problem (MDRP) and proposed a rolling-horizon repeated matching approach to solving large-scale MDRPs. 
They observed that the uncertainty of meal preparation time has a significant impact on the system reliability. 
\citet{yildiz2019provably} assumed perfect information about the order-arrival in their meal delivery problem. 
They considered different objectives, including cost minimization, click-to-door minimization, and the combination of both, and solved their model using a combination of row and column generation.
They noted that their model takes five minutes to assign a courier, making it difficult to be implemented for real-time decision-making settings.

\citet{chen2019Can} considered the problem of maximizing the revenue from the served orders given a limited number of couriers and a specific schedule. 
They formulated a Markov decision process (MDP) model for their problem and used multi-agent reinforcement learning to solve their models. 
They assessed the generalizability of their approach under different circumstances, e.g., zone distribution variations, different grid sizes, and various order distributions. 
They also applied their methods to the pickup-service platform of Alibaba to show the model scalability in practice. 
\citet{Steever2019} studied the virtual food court delivery problem (VFCDP) in which a customer can order from multiple restaurants at a time. 
They used a mixed-integer linear programming formulation in a simulation environment for their problem. 
They evaluated the model performance in terms of earliness, freshness, waiting time, and couriers' miles traveled using the datasets for Auburn and Buffalo cities. 

\citet{Mao2019} investigated the effect of delivery time on the future demands of the customers. 
They incorporated the couriers' knowledge and experience in their model. 
Based on their analysis of a Chinese online delivery platform in Hangzhou, they found that late deliveries can severely affect the customers' future demands. 
On the other hand, faster deliveries provide a positive impact on the number of upcoming orders but not as significant as the negative effect of the late orders. 
\citet{Ulmer2020deadline} defined the restaurant meal delivery problem (RMDP) as a fleet of couriers that serve dynamic orders with known locations and unknown food preparation time, where the objective is to minimize order delays. 
They formulated an MDP model for the problem in which the orders can be postponed and bundled.

\subsection{Autonomous delivery}
Autonomous electrical vehicles and drones have started to be considered as alternatives for couriers for same-day deliveries as well as long-distance deliveries. 
The concept of using these vehicles for deliveries lead to many studies. 
\citet{jennings2019} investigated three research questions about Sidewalk Automated (or Autonomous) Delivery Robots (SADRs); namely, what are the existing regulations in the US?, what are the technical capabilities of the SADRs?, and do SADRs save time and cost compared to other alternatives? 
Their results showed that SADRs lead to cost and time savings in some circumstances. 
\citet{chen2022deep} studied an assignment problem in which they aimed to find an optimal decision of whether to assign a new order to a drone or vehicles. 
Drones are faster than vehicles, but they have limited capacity, whereas the former has more capacity but may be stuck in urban traffic. 
They used a deep Q-learning approach, and their obtained policies outperformed considered benchmark policies. 
Another research topic on the use of electric vehicles and drones is when to recharge these vehicles~\citep{alkanj2020}.
\citet{pinto2020network} considered a problem in which drones perform meal delivery operations. 
They investigated how the strategic location of the charging stations affects the overall performance of the system, where the maximum time the customers are willing to wait is limited. 

\subsection{Reinforcement learning approaches}
\textcolor{black}{\citet{mnih2015human} proposed deep Q-networks, which laid the foundations for many deep RL methodologies that followed.
Several studies investigated strategies to overcome the slow convergence (or divergence) issues associated with DQNs.
These strategies include Double DQNs~\citep{Hasselt2016}, which focus on action value overestimation in DQNs, prioritized experience replay~\citep{Schaul2016}, which aim to address low sampling efficiency by strategically choosing the samples in DQN training, dueling network architectures~\citep{wang2016}, and noisy DQN~\citep{fortunato2017noisy} among others.
\citet{hessel2018rainbow} provided an empirical analysis that compare the performance of various combinations of these strategies.
}

The reinforcement learning (RL) approaches have been used for fleet management problems with both same-day delivery and long-distance delivery, as well as variants of vehicle routing problems.
\citet{chen2022deep} proposed a deep Q-learning approach to solve a same-day delivery problem to determine whether a drone or a vehicle should deliver an upcoming order. 
\citet{zhou2019multi} used an extension of Double DQN with a soft update to solve a large-scale order dispatching problem. 
In their formulation, all agents work independently with the guidance of a shared policy. 
They compared their algorithm to a variant of Double DQN, nearest-distance order dispatching, decentralized multi-agent deep reinforcement learning, and centralized combinatorial optimization for three cities under different traffic conditions. 
Their numerical study showed that their approach outperformed others with respect to accumulated driver income and order response rate. 
\citet{lin2018efficient} proposed two new contextual multi-agent reinforcement algorithms, namely contextual DQN and contextual actor-critic algorithm. 
In addition, they compared their proposed approaches with the rule-based algorithms, Q-learning, SARSA, DQN, and actor-critic algorithms.
They noted that their methods are superior to the existing RL and rule-based methods. 

The RL approach is also frequently used for vehicle routing problems, which shares many similarities to the courier routing and assignment. 
\citet{vera2019deep} proposed an actor-critic network algorithm and compared it with various benchmark algorithms, including Clarke-Wright savings heuristic and sweep heuristic. 
They ran their experiments for a varying number of customers and vehicle capacities. 
They further showed that a policy generated by their algorithm works better for the routing problem involving a fleet of vehicles with heterogeneous capacities. 
\citet{peng2019deep} presented a dynamic attention neural network model with dynamic encoder-decoder architecture for a vehicle routing problem and compared it with a non-dynamic attention model for different configurations. 
Their results highlighted the improved performance of the dynamic attention over its non-dynamic version.

\subsection{Approximate dynamic programming approaches}
Approximate Dynamic Programming (ADP) has been considered for many problems, including ride-sharing, vehicle-routing, and courier assignment tasks, which share many similarities by their nature.
\citet{alkanj2020} defined a ride-hailing system in which their objective is to assign the proper vehicle to a given trip, decide on the re-charging time, and re-position the vehicle when it is idle. 
Their formulation is comprehensive in that they also determined the trip price and the fleet size for the system. 
They used ADP techniques to solve their MDP model and suggested that using hierarchical aggregation leads to high-quality results and fast convergence. 
Moreover, their dynamic pricing solution increased the revenue by $13\%$. 

\citet{ulmer2019offline} proposed an offline ADP approach by incorporating online service request predictions into their methodology, where the stochasticity of the requests contributes to the uncertainties in the decisions. 
They showed that temporal and spatial anticipation of service requests outperforms the baseline policies. 
In another study, \citet{ulmer2019preemptive} used ADP to improve the performance of the same-day delivery system. 
They showed that a preemptive depot return, i.e., returning to the depot when some packages are currently on-board to address additional new requests that are received during the same day, increases the number of delivered orders per day. 

\citet{Ulmer2020M-VFA} studied the capacitated customer acceptance problem with stochastic requests (CAPSR), in which a dispatcher receives an order with a specific address, required capacity, and the revenue. 
The task is to maximize the revenue through instant acceptance or rejection of the orders. 
They used meso-parametric value function approximation that incorporates features of both routing and knapsack formulation.
They showed that their approach is superior to the benchmarks for the CAPSR. 

\subsection{Most related work}
Table~\ref{tab:LR_table} shows the relative positioning of our work with respect to the previous studies in the literature. 
We specifically provide the research objectives, proposed solution approaches, and model specifications in this table. 
In terms of the modeling objectives, our main consideration is to solve the meal delivery problem, which is similar to the order dispatching problem and closely associated with the vehicle routing problem (VRP). \textcolor{black}{The main purpose of the vehicle routing problem is to determine the optimal routes for the vehicles visiting more than one location in terms of time or distance. 
In order-dispatching problems, the couriers pick up orders from a centralized location (i.e., warehouse/depot) or different stores and deliver them to target customers. 
In this regard, it can be considered as a special case of the vehicle routing problem.
While the meal delivery problem is similar to those for finding the optimal routes and delivering orders to target customers, it differs in certain aspects. 
For instance, in the case of meal delivery, the origin of the order (i.e., a restaurant) is determined explicitly by the customer.
In addition, meal delivery problem has a more dynamic nature than VRP, and inherent stochasticity (e.g., frequent order arrivals and uncertainty in food preparation/delivery times) necessitates a framework that enables real-time decision making. 
Furthermore, unlike generic VRP, a courier typically does not visit multiple locations during the delivery as 
strict delivery time restrictions that should also consider the food preparation time constitute an important dimension of the problem.}

\begin{table}[!ht]
    \centering
    \caption{Summary of the most relevant studies in the literature}
    \renewcommand{\arraystretch}{1.5}
    \resizebox{\textwidth}{!}{\begin{tabular}{l|ccc|cccc|ccccc}
    \toprule
    &\multicolumn{3}{c}{\textbf{Objective}}&\multicolumn{4}{|c}{\textbf{Solution Approach}}&\multicolumn{5}{|c}{\textbf{Model Specification}}\\
    \cline{2-4}\cline{5-8}\cline{9-13}
     \textbf{Reference} & \textbf{VR} & \textbf{OD} & \textbf{MD} & \textbf{IP} & \textbf{RL} & \textbf{DRL} & \textbf{ADP} & \STAB{\rotatebox[origin=c]{90}{\textbf{Deadlines}}} & \STAB{\rotatebox[origin=c]{90}{\textbf{Timing}}} & \STAB{\rotatebox[origin=c]{90}{\textbf{Prepositioning }}} & \STAB{\rotatebox[origin=c]{90}{\textbf{Rejection}}} &
     \STAB{\rotatebox[origin=c]{90}{\textbf{Assignment$^\textbf{+}$}}}\\
      \hline
     \citet{alkanj2020} & & \cmark & & & & & \cmark & \cmark & \cmark & \cmark & \cmark &\\
     \citet{chen2022deep} & &\cmark & & & &\cmark & &\cmark &\cmark & &\cmark &\cmark\\
     \citet{chen2019Can} & &\cmark & & &\cmark & & &\cmark & & & &\\
     \citet{lin2018efficient} & &\cmark & & & &\cmark & & & & & &\cmark\\
     \citet{peng2019deep} &\cmark & & & & &\cmark & & & & & &\\
     \citet{reyes2018meal} & & &\cmark &\cmark & & &\cmark &\cmark &\cmark &\cmark & &\cmark\\
     \citet{Steever2019} & & &\cmark &\cmark & & & &\cmark &\cmark &\cmark & &\cmark\\
     \citet{Ulmer2020deadline} & & &\cmark & & & &\cmark &\cmark &\cmark & &\\
     \citet{Ulmer2020M-VFA} &\cmark & & & & & &\cmark &\cmark & & &\cmark &\\
     \citet{ulmer2019offline} &\cmark & & & & & &\cmark & &\cmark & &\cmark &\\
     \citet{ulmer2019preemptive} & &\cmark & & & & &\cmark &\cmark & & & &\\
     \citet{yildiz2019provably} & & &\cmark &\cmark & & & &\cmark &\cmark & & &\cmark\\
     \citet{zhou2019multi} & &\cmark & & & &\cmark & &\cmark & & & &\\
      \hline
     \textbf{Our Study}  & & &\cmark & & &\cmark & &\cmark &\cmark &\cmark &\cmark &\cmark\\
     \bottomrule
    \end{tabular}}
    \label{tab:LR_table}
    \footnotesize
    \begin{tablenotes}
    \item \quad VR: Vehicle Routing, \ OD: Order Dispatching, \ MD: Meal Delivery, \ IP: Integer Programming 
    \item \quad RL: Reinforcement Learning, \ DRL: Deep RL, \ ADP: Approximate Dynamic Programming 
    \end{tablenotes}
\end{table}

In terms of the solution methodology, integer programming has been the most commonly used technique for the meal delivery problems~\citep{reyes2018meal, Steever2019, yildiz2019provably}, and ADP has gained significant attention in recent years~\citep{Ulmer2020deadline}. 
Although (deep) RL approaches have been used for the vehicle routing and order dispatching problems~\citep{chen2022deep, chen2019Can, lin2018efficient, peng2019deep, vera2019deep, zhou2019multi}, to the best of our knowledge, various deep RL extensions (e.g., DDQN, dueling DDQN, and DDQN with prioritized experience replay) have not been considered for the meal delivery problem. 
In this study, we investigate a comprehensive list of DQN extensions together with the different hyperparameter settings that were not considered in previous studies.

In the ``Model Specification'' grouping of Table~\ref{tab:LR_table}, ``Deadlines'' implies the specified time window by which the order should be delivered. 
It is one of the most commonly used model specifications in meal delivery problems. 
It is an important consideration because meals should be delivered in a limited time to keep them fresh.
``Timing'' indicates if the delivery time affects any specific component of the model, especially reward. 
In general, the shorter the delivery time is, the higher reward the system will receive. 
``Prepositioning'' is defined as anticipating the next order origin and positioning the courier or vehicle accordingly before the order arrives in the system. 
We assume that if there is no new order and the courier is idle, they may move towards a restaurant rather than returning to the depot. 
It allows couriers to learn popular restaurants with a higher number of orders so that the overall delivery time is shortened. 
``Rejection'' is the flexibility of not delivering an order. 
Our model aims to learn rejecting the current order if its delivery is infeasible or the upcoming order is expected to come from a more convenient location.
Hence, this consideration enables not to lose time in delivering an unprofitable order. 
Accordingly, while other approaches fulfill both profitable and unprofitable orders, our model can fulfill more orders in the same time window. 
Furthermore, having the reject option might be particularly impactful when the number of available couriers is low.
Lastly, ``Assignment$^\textbf{+}$'' corresponds to assigning more than one task/order to a courier at a time. 
Our model has the flexibility to assign a new order to a courier who is currently delivering an order. 

\textcolor{black}{By incorporating the model specifications listed in Table~\ref{tab:LR_table}, our model can better mimic the real-life settings and take proactive and smart decisions. 
In meal delivery practice, companies commit to delivering foods within a specified time window. 
We handle this important consideration with the deadline specification.
Through the reward definition in our formulation, which links higher rewards to speedy deliveries, we also aim to ensure the freshness of the food. 
Moreover, our model aims to learn the next order origin and position the couriers based on this information, which enables shortening the delivery times. 
In addition, we incorporate smart order rejection based on the delivery time of the orders at hand and the features of the upcoming orders. 
Finally, our model can assign more than one task at a time to a courier. 
As such, all these specifications enhance customer satisfaction by decreasing delivery times and improving the overall quality of the system.}

\section{Methodology} \label{method}
In this section, we first present our MDP model for on-demand food delivery service along with the modeling assumptions. We then summarize different deep RL algorithms used to solve our model.

\subsection{Assumptions}
We consider a set of couriers $\indCourier \in \setCourier$, who are assigned an order, $\indOrder$, move to a restaurant, $\indRestaurant$, to pick it up, and deliver it to a customer, $\indCustomer$, over a region represented by $\dimGridm \times \dimGridn$ grid. 
The objective is to find a courier assignment policy that maximizes the total reward collected over time. 
The main assumptions of the problem are as follows.
\begin{itemize}
    \item We convert the real-world maps to the grids consisting of cells of size $500\ \text{meters} \times 500\ \text{meters}$ and assume that moving from one grid cell to the adjacent one approximately takes one minute. 
    
    \item Orders may come from each grid cell with a certain probability value.
    
    \item The restaurants' locations in the grid are fixed, and the number of orders from each restaurant varies.
    
    \item The preparation time of food for order $\indOrder$ is assumed to be between 5 to 15 minutes (i.e., $\timePrepare_{\indOrder} \in [5,15]$). 
    
    \item We assume that the company aims to deliver orders within 25 minutes; however, they still accept orders that take up to 45 minutes.
    
    \item The time unit in the system is ``minute'', and the system status and the collected rewards are updated every minute.
    
    \item If we assign an order to a courier that is returning to the depot, or moving towards a popular restaurant, then the courier can start their mission without returning to the depot or the restaurant. 
    In other words, whenever a courier is idle, we assume he is immediately available for the new orders.
    
    \item The time between two consecutive orders has an exponential distribution with the rate of $\paramOrderDistn_{\indTime}$, where $\indTime$ represents the time of the day. That is, order arrivals might change hourly and reflect rush hours and idle times.
    
    \item \textcolor{black}{
    If a customer places an order from multiple restaurants in a single order, we break that order into multiple orders and handle them separately. 
    For instance, when a customer orders from two different restaurants, we register them as two distinct orders with different origins and the same destination. 
    The model might assign them to a single courier or multiple couriers based on couriers' and restaurants' locations.}
\end{itemize}

\subsection{Problem formulation}
\begin{figure*}[!ht]
  \centering
  \medskip
  \begin{subfigure}[t]{.31\textwidth}
    \centering\includegraphics[width=\textwidth]{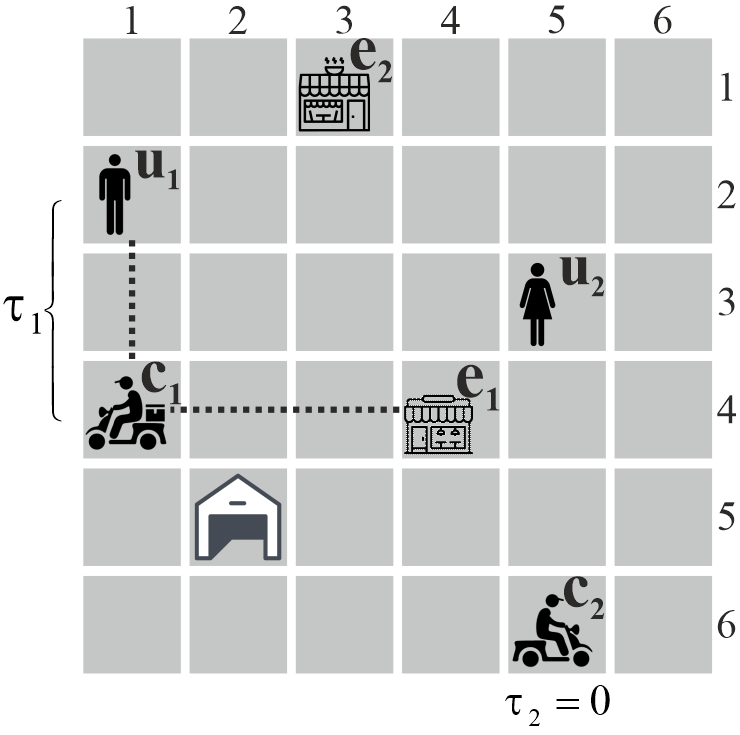}
    \caption{First scenario} \label{fig:typical_order1}
  \end{subfigure}\quad
\begin{subfigure}[t]{.31\textwidth}
    \centering\includegraphics[width=\textwidth]{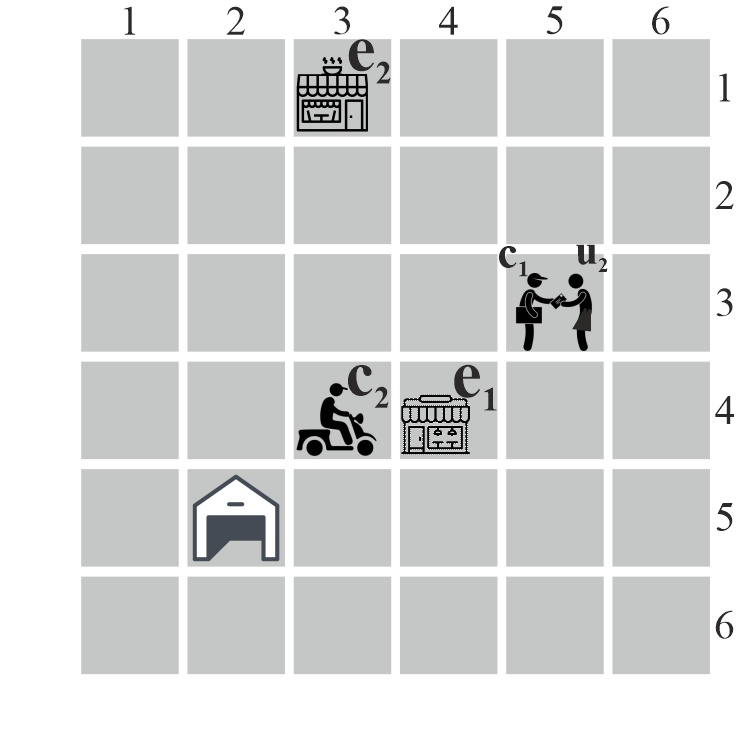}
    \caption{Second scenario} \label{fig:typical_order2}
  \end{subfigure}\quad
  \begin{subfigure}[t]{.31\textwidth}
    \centering\includegraphics[width=\textwidth]{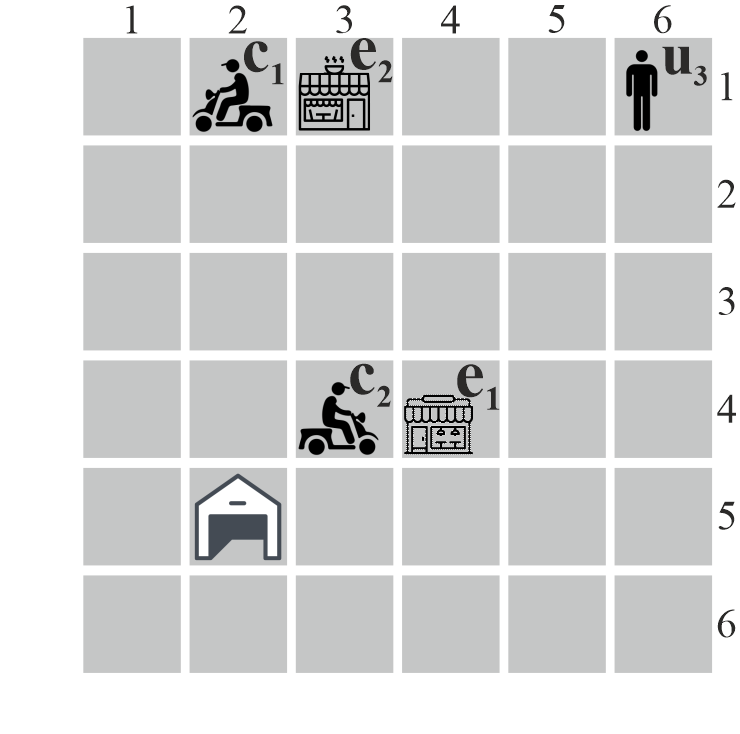}
    \caption{Third scenario} \label{fig:typical_order3}
  \end{subfigure}
  ~
    \caption{Typical orders in a system with two couriers.} \label{fig:typical_order}
\end{figure*}

In our problem setting, the order cycle starts with receiving an order, and then the company either accepts or rejects the order based on the time to deliver it. 
Ideally, the company aims to accept all orders with the caveat that some orders might lead to excessive delivery times. 
The company rejects orders only if the customer does not accept the longer waiting time. 
As an order is placed, the corresponding restaurant is notified, and a courier is dispatched from the depot to the restaurant to deliver the order. 
When the food is delivered, the courier returns to the depot. 
A typical order in the system is shown in Figure~\ref{fig:typical_order} on a $6 \times 6$ grid for three different order, customer and courier placements. 
We use this reference figure to explain the states, actions, and rewards in the system under different conditions.

\subsubsection{States}
The states in our model are defined based on the couriers' distances to the order origin, destination, and depot location. 
Adopting such state definition leads to substantially fewer states in the state space when compared to defining states using the exact locations of the couriers, order origin, and destination. 
We characterize the state space for our problem as follows.
\begin{align*}
    \setState = \{(\timeDeliver_{\indCourier}^{\indOrder}, \distanceDepot_{\indCourier}, \distanceRestaurant_{\indCourier}^{\indRestaurant}), \ \forall \indCourier, \indOrder, \indRestaurant\}
\end{align*}
Specifically, the state space, $\setState$, is composed of the following elements:
\begin{itemize}
    \item \textbf{Expected delivery time ($\timeDeliver_{\indCourier}^{\indOrder}$)} of the order $\indOrder$ for courier $\indCourier$ takes into account the distance between the restaurant (which is known for each order) and the customer ($\distance_{\indOrder}$), the preparation time of the order ($\timePrepare_{\indOrder}$), and time to deliver the previous order(s) ($\timeDeliverPrev_{\indCourier}$). 
    If no order is assigned to courier $\indCourier$, time to deliver the previous assigned order(s) is equal to zero (i.e., $\timeDeliverPrev_{\indCourier} = 0$).
    Accordingly, expected delivery time can be obtained as
    \begin{equation}
        \timeDeliver_{\indCourier}^{\indOrder} = \distance_{\indOrder} + \max{\{\timePrepare_{\indOrder}, \big( \timeDeliverPrev_{\indCourier} + \distanceAvail_{(\indCourier,\indOrder)} \big) \}}
    \end{equation}
    where $\distanceAvail_{(\indCourier,\indOrder)}$ is the distance between the restaurant and the courier's ``location of availability''. The location of availability is either the current location of a courier if he/she is idle or the location of the final customer if the courier has active order(s) at hand. 
    
    In Figure~\ref{fig:typical_order1}, we have two couriers, $\indCourier_1$ and $\indCourier_2$. 
    Courier $\indCourier_1$ is delivering the previous assigned order and is in the path between the restaurant $\indRestaurant_1$ and customer $\indCustomer_1$. 
    The time to deliver the previous order for that courier is equal to $\timeDeliverPrev_1 = 2$. 
    If the new order $\indOrder_2$ has the preparation time of 6, and we decide to assign it to courier $\indCourier_1$, then the expected delivery time $\timeDeliver_{\indCourier_1}^{\indOrder_2}$ is equal to 10 ($=4 + \max \{6 , (2 + 3)\}$), including two minutes to finish the current order, three minutes to go to restaurant $\indRestaurant_2$, an extra minute waiting time for the food to be prepared, and then 4 minutes to deliver it to the customer $\indCustomer_2$. 
    Similarly, the expected delivery time for the other courier is computed as $\timeDeliver_{\indCourier_2}^{\indOrder_2} = 4 + \max \{6, (0+7)\} = 11$.

    
    \item \textbf{Distance to depot ($\distanceDepot_{\indCourier}$)} indicates the Manhattan distance between a courier $\indCourier$ and the company's local depot. 
    This variable is activated only when the courier is idle, and the action prescribes \textit{returning to depot}. 
    In Figure~\ref{fig:typical_order2}, the distance to the depot is only feasible for courier $\indCourier_1$ as he delivered the order a moment ago and became idle in the system; therefore, $\distanceDepot_{\indCourier_1}$ is equal to 5 based on the Manhattan distance.
    
    \item \textbf{Distance to restaurants ($\distanceRestaurant_{\indCourier}^{\indRestaurant}$)} is the Manhattan distance between courier $\indCourier$ and restaurant $\indRestaurant$. 
    This state component becomes active when the courier is idle, and the action prescribes \textit{moving towards restaurant $\indRestaurant$}. 
    In Figure~\ref{fig:typical_order2}, the model only computes the distance to restaurants for courier $\indCourier_1$ who becomes idle in the system; therefore, the set $\distanceRestaurant_{\indCourier_1}$ includes $\distanceRestaurant_{\indCourier_1}^{\indRestaurant_1} = 2$ and $\distanceRestaurant_{\indCourier_1}^{\indRestaurant_2} = 4$. 
    The model considers the distance to each restaurant and, more importantly, learns the popularity of each restaurant and its expected long-run return. 
    Hence, the best action is likely to be the one that provides the highest gain through an intelligent repositioning of the couriers.
\end{itemize}

\subsubsection{Actions}
We have both reject ($\actionReject$) and accept ($\actionAssign$) decisions when a new order comes to the system.
We also define an action to take into account the case that when couriers finish their task, they should not necessarily come back to a depot; instead, they can go to popular restaurants to be ready for possible upcoming orders. Therefore, when a courier becomes idle, they can choose from the actions of returning to a depot ($\actionReturnDepot$) or moving towards restaurants ($\actionMoveToRestaurant$). Accordingly, action set for a given courier $\indCourier$ is defined as
\begin{equation}
    \setAction =  \{\actionAssign_{\indCourier}^{\indOrder}, \actionReject^{\indOrder}, \actionMoveToRestaurant_{\indCourier}^{\indRestaurant}, 
    \actionReturnDepot_{\indCourier}\}.
\end{equation}
where $\actionAssign_{\indCourier}^{\indOrder}$ and $\actionReject^{\indOrder}$ denotes assigning and rejecting order ${\indOrder}$, respectively, $\actionMoveToRestaurant_{\indCourier}^{\indRestaurant}$ corresponds to courier $\indCourier$ moving towards restaurant $\indRestaurant$, and $\actionReturnDepot_{\indCourier}$ corresponds to courier $\indCourier$ moving towards the depot.

At any time $\indTime$, the list of feasible actions, $\indAction_t \in \setAction$, is limited. 
Therefore, we define a triggering mechanism that filters the proper set of actions.
For instance, when we have a new order, we cannot select the action of \textit{return to depot} ($\actionReturnDepot$), or when a courier becomes idle, we cannot choose the action of \textit{reject} ($\actionReject$).
That is, our problem has an event-based environment. 
We can define the proper set of actions through regular monitoring of the system for new event occurrences. 
For instance, in Figure~\ref{fig:typical_order1}, the possible action set is $\{\actionReject^{\indOrder_2}, \actionAssign_{\indCourier_1}^{\indOrder_2}, \actionAssign_{\indCourier_2}^{\indOrder_2}\}$; in Figure~\ref{fig:typical_order2}, the possible action set for courier $\indCourier_1$ is $\{\actionReturnDepot_{\indCourier_1}, \actionMoveToRestaurant_{\indCourier_1}^{\indOrder_1}, \actionMoveToRestaurant_{\indCourier_2}^{\indOrder_1}\}$; in Figure~\ref{fig:typical_order3}, the possible action set, given the new order 3 and having two available couriers, is $\{\actionReject^{\indOrder_3}, \actionAssign_{\indCourier_1}^{\indOrder_3}, \actionAssign_{\indCourier_2}^{\indOrder_3}\}$.

\subsubsection{Rewards}
The immediate reward, $\funcReward(\indState,\indAction)$, in our system for taking an action $\indAction \in \setAction$ at a given state $\indState \in \setState$ is computed as follows.
\begin{equation}
\funcReward(\indState,\indAction) =
    \begin{cases}
      +45-\timeDeliver_{\indCourier}^{\indOrder} &  \text{if}\ \indAction = \actionAssign_{\indCourier}^{\indOrder} \\
      -15, &  \text{if}\ \indAction = \actionReject^{\indOrder} \\
      -\distanceDepot_{\indCourier}/10 &  \text{if}\ \indAction = \actionReturnDepot_{\indCourier}\\
      -\distanceRestaurant_{\indCourier}^{\indRestaurant}/10 &  \text{if}\ \indAction = \actionMoveToRestaurant_{\indCourier}^{\indRestaurant}\\
    \end{cases}
  \label{eq:reward}
\end{equation}

We assume that the company has a delivery time limit of 45 minutes for accepting the orders.
Therefore, a positive reward is incurred for a new assigned order if its delivery time is less than 45 minutes. 
Accordingly, shorter the delivery time is, higher the reward received. 
\textcolor{black}{For instance, if the time to deliver order $\indOrder_1$ for courier $\indCourier_3$ is 20 minutes ($\timeDeliver_{3}^{1} = 20$), then assigning $\indOrder_1$ to $\indCourier_3$ will result in a reward of 25 ($=45-20$).}
Additionally, we assume a fixed negative reward for rejecting orders. 
The penalty of $-15$ for rejection indicates that the system may still prefer to accept late orders if their delivery times do not exceed one hour. 
\textcolor{black}{In other words, if order $\indOrder_1$ takes 61 minutes to be delivered by courier $\indCourier_4$, the total reward will be -16 ($=45-61$), and the system prefer rejecting such order or assigning it to another courier rather than courier $\indCourier_4$.} 
Finally, we assign a small negative reward for moving to a depot or a restaurant in order to decrease the likelihood of unnecessary movements while the couriers are idle. 
\textcolor{black}{That is, a courier being idle does not incur any negative rewards, while inessential relocation when no order is assigned leads to a negative reward.
Given the concrete example in} Figure~\ref{fig:typical_order1}, if we assign the order $\indOrder_2$ to courier $\indCourier_1$, then the immediate reward is $+45-\timeDeliver_{\indCourier_1}^{\indOrder_2} = +45-10=+35$.
On the other hand, in Figure~\ref{fig:typical_order2}, where it is possible to choose between the actions of going to the depot or a restaurant, if we dispatch courier $\indCourier_1$ to the restaurant $\indRestaurant_2$, a penalty of $\frac{-4}{10}$ is incurred. 

\textcolor{black}{
We note that the delivery time implicitly incorporates various relevant factors such as distance and traffic.
Accordingly, adding a negative reward for rejected orders (i.e., for those taking more than 45 minutes) inherently takes into account the delivery distance as well. 
If an order takes a longer duration than the threshold based on its origin, destination, preparation time, and couriers' locations, the model tends to decline it. 
Additionally, grid-based representation of the geographic regions in our problem relates the time and distance since we assume that moving from one grid to another takes approximately one minute and one movement in the system. 
}

\subsection{Reinforcement learning algorithms}
Reinforcement Learning (RL) has three main components: agent, environment, and the interaction in between. 
The agent exists in a dynamic environment in which he/she gains knowledge through trial-and-error interactions. 
The agent selects an action, $\indActionBig \in \setAction$ based on the current state, $\indStateBig \in \setState$, at each iteration. 
The action might change the environment and typically result in a scalar reward, $\indRewardBig \in \setReward$. 
The agent's objective, on each step, is to choose the specific action that increases the total expected reward in the long run. 
For our meal delivery problem, we leverage the ability of reinforcement learning to explore the environment and determine a policy that optimizes long-run expected total rewards.

\subsubsection{Deep Q-Networks}
One of the most commonly used RL algorithms is Q-learning, which updates $\QValue$-values greedily while it only considers the maximum $\QValue$-value in the following state, that is,
\begin{align}
    \begin{split}
        \QValue_{new}(\indStateBig,\indActionBig) &= \QValue_{old}(\indStateBig,\indActionBig) + \stepSize \update \\
        \update &=  \indRewardBig + \discount \argmax_{\indAction} \QValue(\indStateBig^\prime, \indAction) - Q_{old}(\indStateBig,\indActionBig)
    \label{eq:Q-learning}
    \end{split}
\end{align}
where $\indRewardBig$ is the immediate reward, $\discount$ is the discount factor, and $\QValue$ is an approximation of state-action value pairs~\citep{Francois-Lavet2018}.

When the state space is large, Q-learning typically requires an excessively large number of iterations to fully explore the environment and, therefore, causes underfitting. 
\citet{mnih2015human} introduced Deep Q-Networks (DQN) as an effective algorithm for dealing with large state spaces. 
A multi-layer neural network estimates a vector of action values, $\QValue(\indStateBig,\indActionBig;\QNetParams)$, where $\QNetParams$ are the parameters of the network. 
They further improved DQN by augmenting experience replay memory and creating a separate target network. 
The former plays the role of memory storage with a predefined length from which we sample randomly during the training phase, whereas the latter postpones updating the weights of the neural network. 
These improvements remove the correlation between consecutive samples and also allow the neural network updates to use each step multiple times; hence, it addresses the fundamental instability problem in RL~\cite{Arulkumaran2017}.

\subsubsection{DQN Extensions}
\citet{Hasselt2016} showed that DQN induces an upward bias in certain cases. 
To alleviate this issue, they proposed Double Deep Q-Network (DDQN). 
DDQN decomposes the max operation in the target into action selection and evaluation. 
Therefore, an online network evaluates the greedy policy, and a target network estimates its value. 
DDQN replaces the update term ($\update$) in Equation~\ref{eq:Q-learning} with $\update^{\text{DDQN}}$, which can be obtained as
\begin{equation*}
    \label{eq:DDQN}
    \update^{\text{DDQN}} =  \indRewardBig + \discount \QValue(\indStateBig^\prime,\argmax_{\indAction} Q(\indStateBig^\prime, \indAction;\QNetParams); \QNetParams^{-}) 
\end{equation*}
where the network's parameters, $\QNetParams^{-}$, are updated every $\ddqnUpdateIter$ iterations with the assignment of $\QNetParams^{-} \leftarrow \QNetParams$. This modification leads to improved stability and better $Q$-value estimates~\cite{Francois-Lavet2018}. 

\citet{Schaul2016} suggested that instead of uniformly sampling from the experience replay memory, the RL algorithm needs to prioritize significant past events.
They proposed that the experiences with a higher temporal-difference (TD) error should be replayed more frequently. 
Let $\priority_{\indTransition} = \tiny \frac{1}{rank(\indTransition)}$ be the priority of transition $\indTransition$. 
Prioritized Experience Replay (PER) defines the probability of sampling transition $\indTransition$ and importance-sampling (IS) weights, $\ISWeight$, as 
\begin{equation*}
\label{eq:PER}
\PERTransitionProb(\indTransition)=\frac{\priority_{\indTransition}^\PERexponenta}{\sum_{k}p_k^\PERexponenta}; \quad \ISWeight_{\indTransition} = \Big(\frac{1}{\PERBuffer\cdot \PERTransitionProb(\indTransition)}\Big)^\PERexponentb
\end{equation*}
where exponent $\PERexponenta$ determines the extent of prioritization, with $\PERexponenta=0$ being a uniform distribution, i.e., no prioritization. Given the current experience replay size of $\PERBuffer$ with the maximum value of $\ERMemory$, the exponent $\PERexponentb$ supervises how much prioritization to apply. As the training is initially unstable, $\PERBuffer$ starts with a small number (e.g., chosen as $0.4$ in our case), and finally converges to one when importance sampling corrections matter towards the end. Accordingly, the weighted IS corrects the possible bias of PER. 

\citet{wang2016} proposed Dueling Double Deep Q-networks (D3QN) with the notion that estimating the value of all action choices is not necessary for all the states. 
Therefore, they separated the network based on two different estimators: one for the state-value function and the other for the state-dependent actions. 

Regarding updating the weights of a network, there are two common approaches: hard and soft update \citep{Hasselt2016}. 
While the target network parameters are updated and are prone to divergence in the hard update, soft update slowly changes the weights of the target network through the formula $\QNetParams^\prime \leftarrow \softUpdateParam \QNetParams + (1-\softUpdateParam)\QNetParams^\prime$, where $\softUpdateParam$ specifies the soft update rate. 
In this work, we explore DQN, DDQN, and their variants. Table~\ref{tab:parameters} shows the default parameter values for these algorithms. 
\renewcommand{\arraystretch}{1.3} 
\begin{table}[!ht]
    \caption{Default parameters used in the training phase}
    \centering
    \resizebox{0.8\textwidth}{!}{
    \begin{tabular}{lcl}
    \toprule
     & & \textbf{Default parameters} \\
     \midrule
    \textbf{Discount factor} & & $\discount = 0.9$  \\
    \textbf{Deep Network} & & The size of hidden units: $\{64, 128, 128, 64\}$ \\
    \textbf{Variations of DQN} & & \begin{tabular}[c]{@{}l@{}} PER: $\PERexponenta = 0.6$, $\PERexponentb =  0.4$
    \\ Soft update: $\softUpdateParam = 0.5 $ , batch size: $\batchSize = 128$ \\ Max. experience replay memory size: $\ERMemory = 20,000$ \end{tabular}\\
     \bottomrule
    \end{tabular}
    }
    \label{tab:parameters}
\end{table}

\subsection{Benchmark policies}
We consider two baseline policies in our analysis for benchmarking purposes.
\begin{itemize}
    \item \textit{Baseline policy 1}: 
    In a typical meal delivery service, if there is no other order in the system, the courier returns to the depot after delivering an order.
    That is, there is no policy of moving towards any restaurant. 
    In addition, in this policy, orders having delivery times of more than 45 minutes are rejected, which is compliant with the adopted practice in many meal delivery companies. 
    We adopt these rules as our first baseline policy, which we denote as P$_{45}$.
    \item \textit{Baseline policy 2}: 
    This policy is similar to the previous baseline with an important difference in the delivery time threshold. 
    This policy, named P$_{60}$, automatically rejects orders having an expected delivery time greater than 60 minutes.   
\end{itemize}
We denote the policies generated by the RL algorithms as $P_{RL}$ in the rest of the paper, with ``RL'' specifying the particular algorithm used to obtain the policy. 

\subsection{Dataset}
We use two types of datasets for our meal delivery problem. 
Specifically, we consider a synthetic dataset to compare DQN extensions and to determine the best performing RL algorithm for our model. 
In addition, we use datasets obtained from Getir, an online food and grocery delivery company that originated in Turkey and recently expanded its operations to the United Kingdom.
Getir dataset includes meal delivery information from three regions of Istanbul, Turkey.

\subsubsection {Synthetic dataset}
We create a synthetic region consisting of $10 \times 10$ grid. 
In this region, there are seven grid cells having restaurants, and the depot is placed in the middle cell (i.e., cell $[5,5]$). 
The number of orders that each restaurant receives per day differs from one restaurant to another. 
However, customers may be located in any cell with a uniform probability distribution. 

\subsubsection {Getir dataset}
\begin{figure*}[!ht]
  \centering
  \medskip
  \begin{subfigure}[t]{.3\textwidth}
    \centering\includegraphics[width=\textwidth]{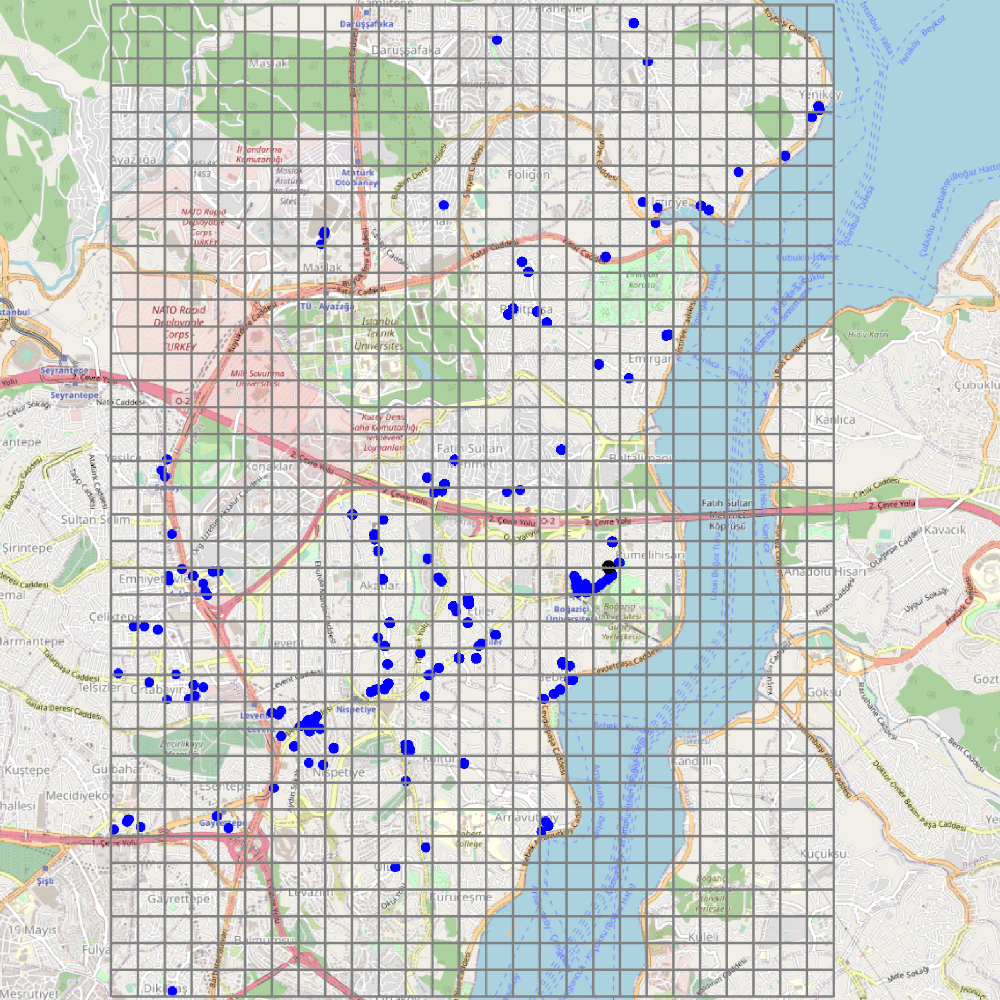}
    \caption{Hisarustu} \label{fig:Hisarustu_rest}
  \end{subfigure}\quad
\begin{subfigure}[t]{.3\textwidth}
    \centering\includegraphics[width=\textwidth]{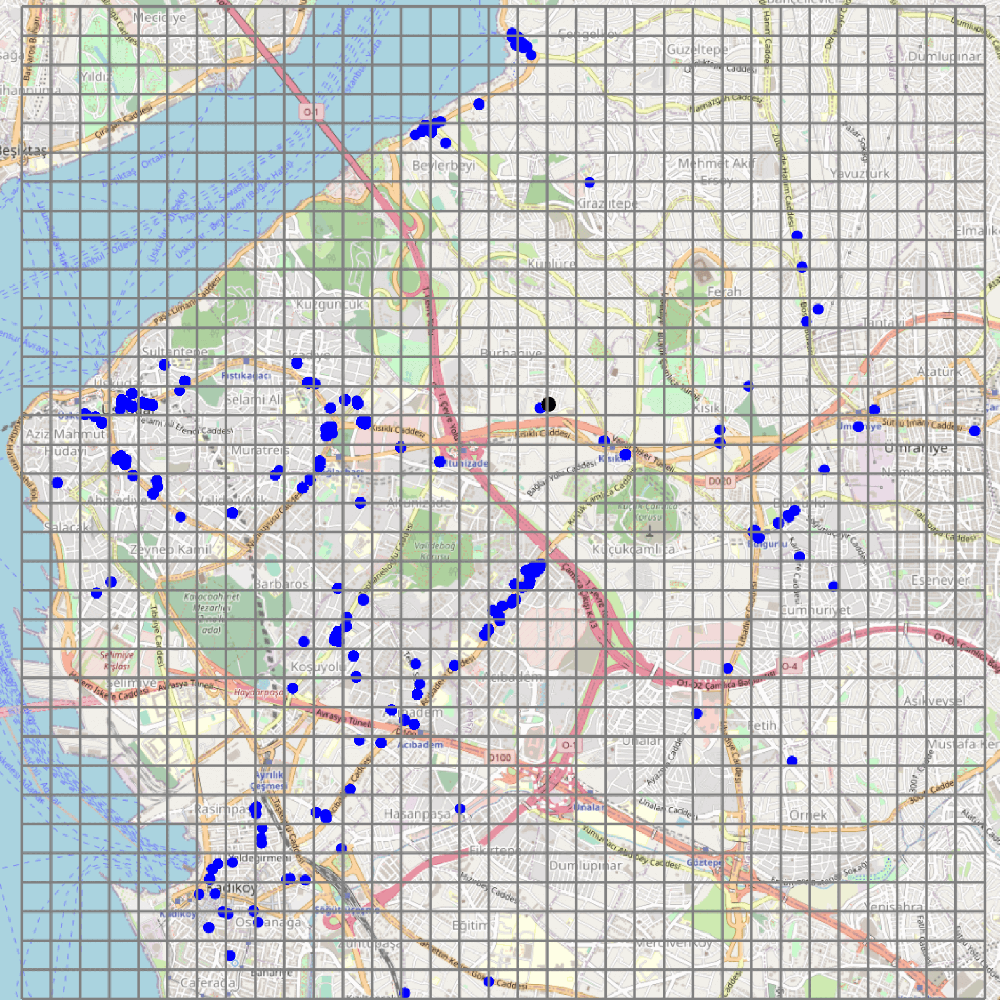}
    \caption{Uskudar} \label{fig:Uskudar_rest}
  \end{subfigure}\quad
  \begin{subfigure}[t]{.3\textwidth}
    \centering\includegraphics[width=\textwidth]{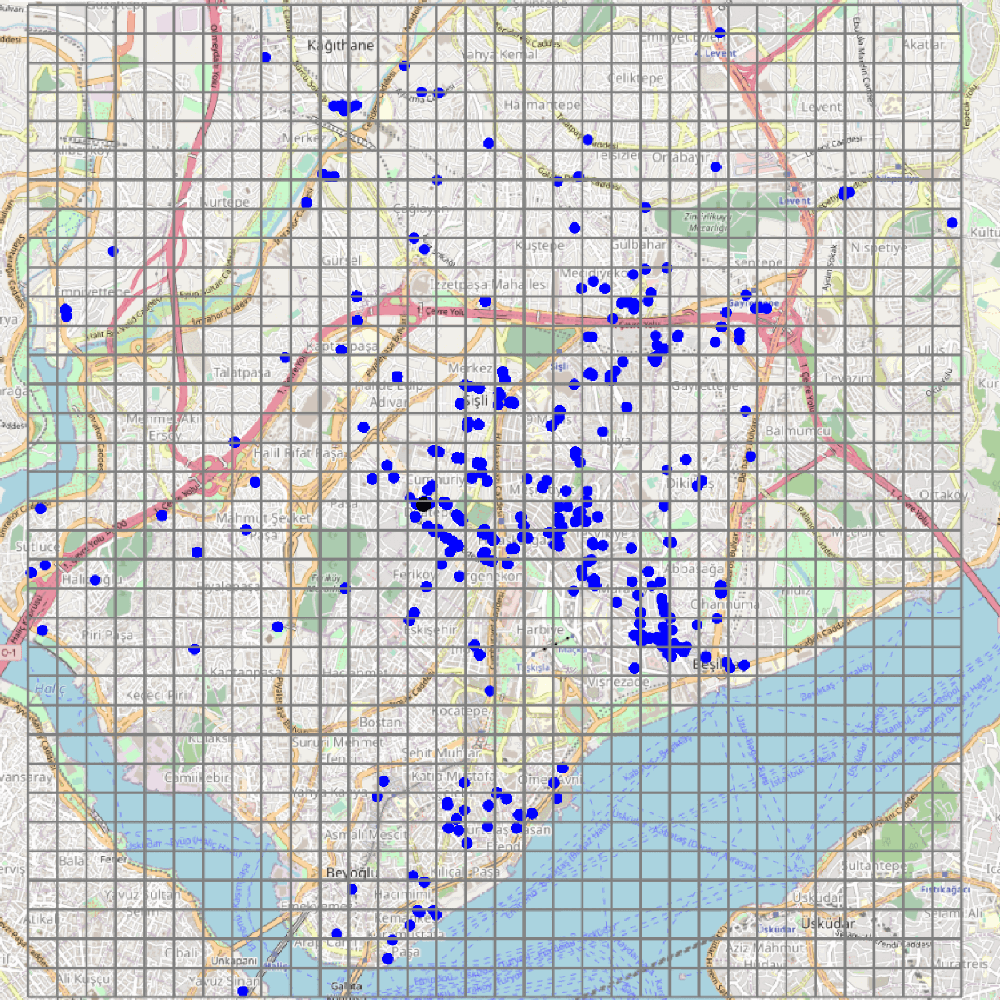}
    \caption{Bomonti} \label{fig:Bomonti_rest}
  \end{subfigure}
  ~
    \caption{Restaurants' locations. Blue dots present the restaurants whose orders are delivered at least one time by Getir's couriers. The legend is intentionally removed to mask the number of orders per the company's request.} \label{fig:regions_rest_and_depot}
\end{figure*}

We analyze meal delivery data from three different regions of Istanbul, namely, Bomonti, Uskudar, and Hisarustu, between October 2019 and April 2020. 
Each region has a different size and order density.
We consider Hisarustu as a low-density, Uskudar as a moderate density, and Bomonti as a high-density region. 
We convert each region's real map to the corresponding grids consisting of cells of size $500\ \text{meters} \times 500\ \text{meters}$. 
Table~\ref{tab:regions} shows the characteristics of each region. 
The number of active cells in the table includes the parts of the map for which we have at least one past order, either from a customer or a restaurant partner. 
Figure~\ref{fig:regions_rest_and_depot} shows the spatial distribution of restaurants across the region; however, note that the popularity of each restaurant --- the number of received orders for a given time --- may differ over time. 
\begin{table}[!ht]
    \caption{Characteristics of the regions}
    \centering
    \resizebox{0.8\textwidth}{!}{
    \begin{tabular}{cccrrr}
    \toprule
     & \textbf{Side} & \textbf{Size} & \textbf{\begin{tabular}[c]{@{}r@{}}\# of active \\ restaurants' \\ grid cells\end{tabular}} & \textbf{\begin{tabular}[c]{@{}r@{}}\# of active \\ customers' \\ grid cells\end{tabular}} & \textbf{\begin{tabular}[c]{@{}r@{}}Depot\\ location\end{tabular}} \\
     \midrule
    \textbf{Hisarustu} & European & $27\times37$ & 87  & 239 &  $[21,19]$ \\
    \textbf{Uskudar} & Anatolian & $34\times33$ & 94 & 405 & $[13, 18]$\\
    \textbf{Bomonti} & European & $34\times32$ & 153 & 373 &  $[17, 13]$\\
    \bottomrule
    \end{tabular}
    }
    \label{tab:regions}
\end{table}

\begin{figure*}[!ht]
  \centering
  \medskip
  \begin{subfigure}[t]{.3\textwidth}
    \centering\includegraphics[width=\textwidth]{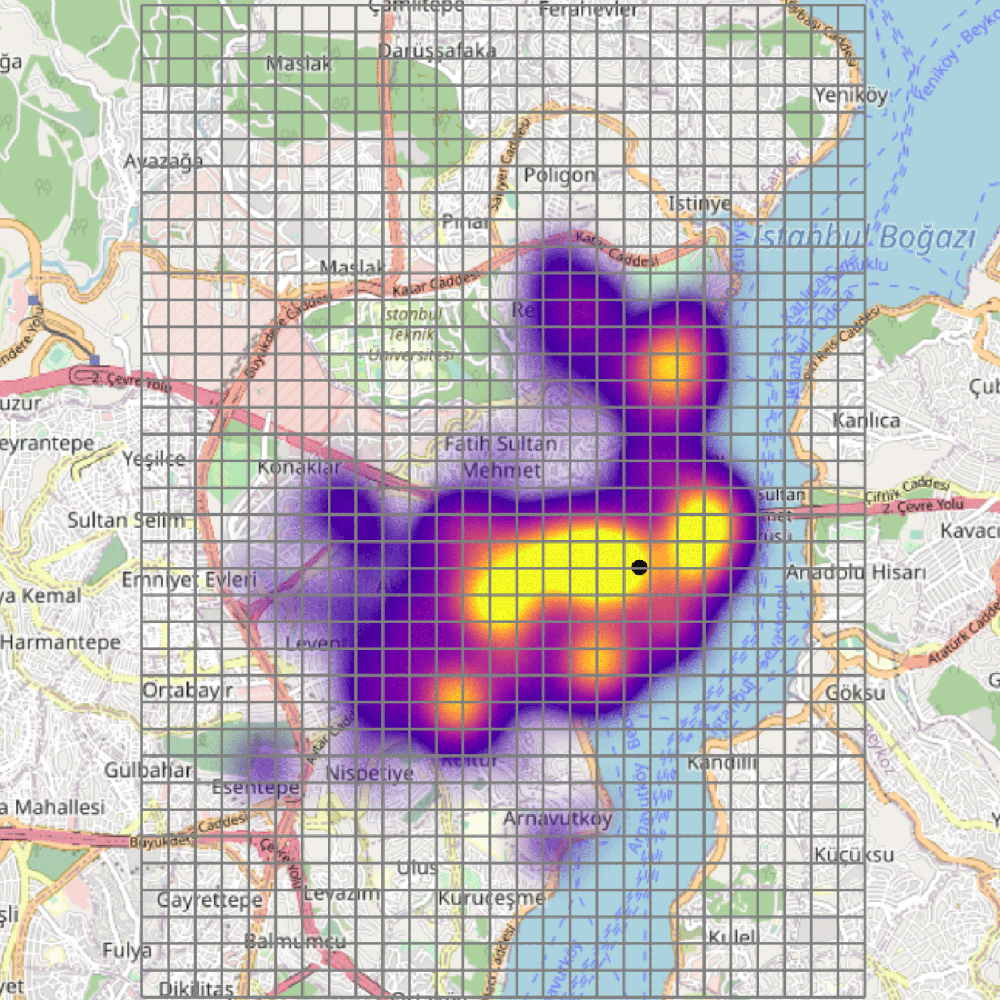}
    \caption{Hisarustu} \label{fig:Hisarustu_cust}
  \end{subfigure}\quad
\begin{subfigure}[t]{.3\textwidth}
    \centering\includegraphics[width=\textwidth]{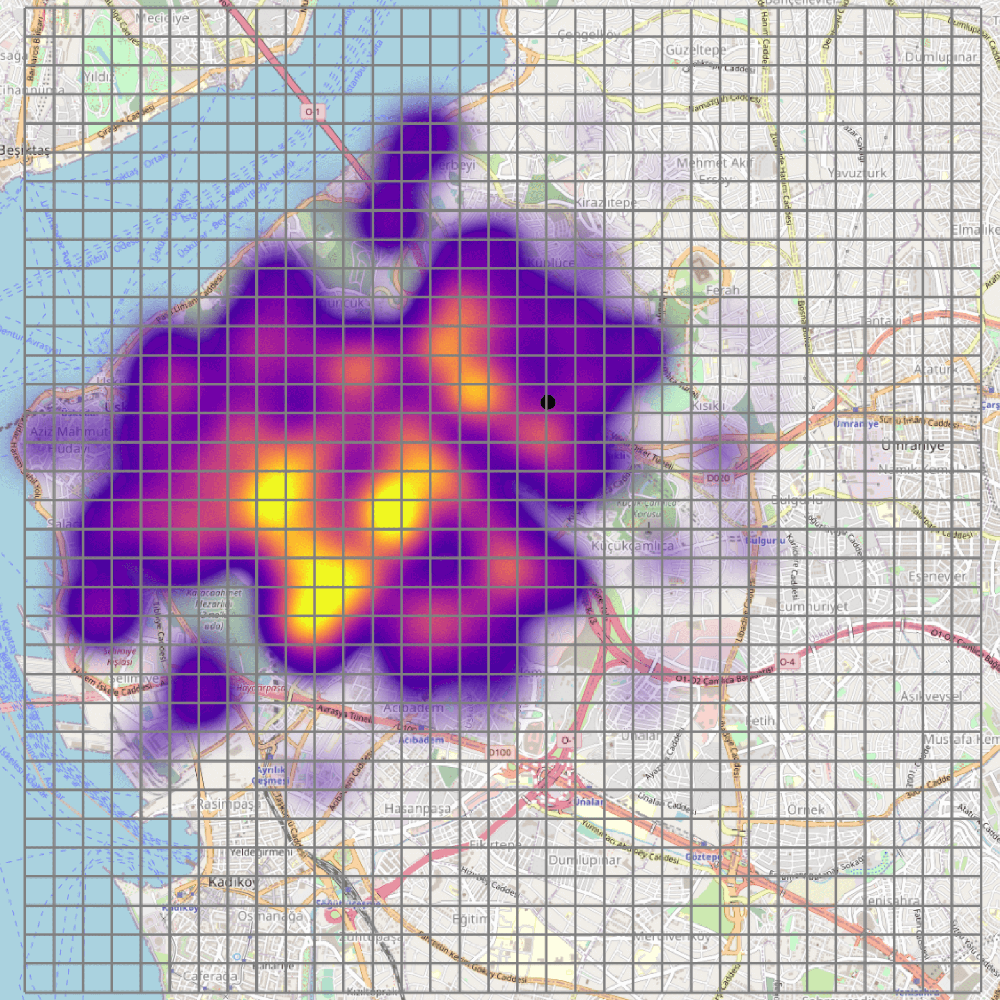}
    \caption{Uskudar} \label{fig:Uskudar_cust}
  \end{subfigure}\quad
  \begin{subfigure}[t]{.3\textwidth}
    \centering\includegraphics[width=\textwidth]{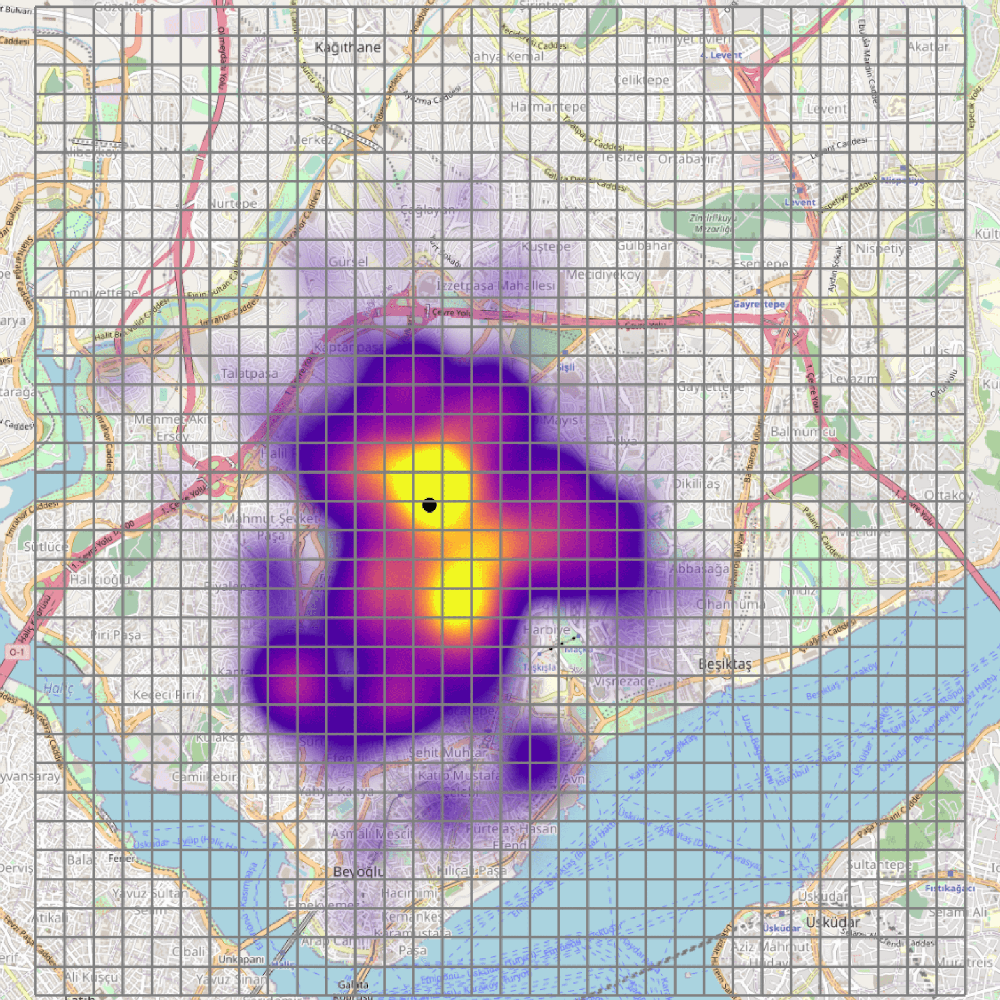}
    \caption{Bomonti} \label{fig:Bomonti_cust}
  \end{subfigure}
  ~
    \caption{The spatial distribution of orders (based on customers' locations) in each region. The black dot represents the depot's location.} \label{fig:regions_cust_and_depot}
\end{figure*}

Figure~\ref{fig:regions_cust_and_depot} presents the order distribution of each region as a heatmap to show the order frequency in each cell. 
In the Hisarustu region, customers are mainly located in the middle of the region. 
In the Bomonti region, order density is very high compared to the Hisarustu and Uskudar regions. 
Customers are clustered in one area in the middle of the region. 
In the Uskudar region, customer density is lower; however, they are distributed to a larger area. 

\begin{figure}[!ht]
\centering
    \includegraphics[width=0.85\linewidth]{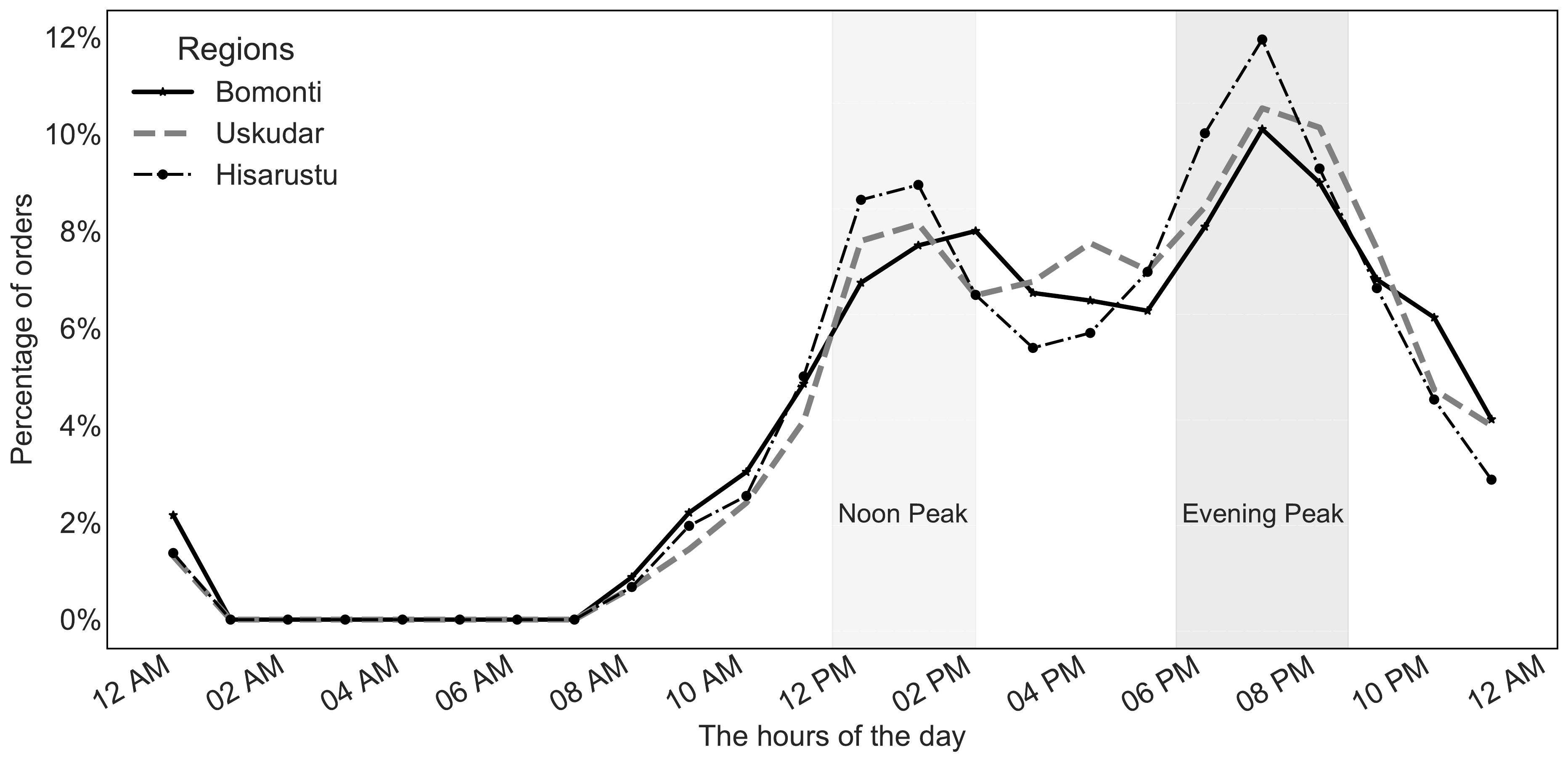}
    \caption{Hourly distribution of the orders during an arbitrary day}
    \label{fig:order_percentage}
\end{figure}

Figure~\ref{fig:order_percentage} shows the percentage of the orders received by the company for three different regions. 
The values in the $x$-axis specify the start time for counting orders; for instance, 11 p.m. includes all orders between 11 p.m. and midnight. 
For each region, ordering behavior follows a similar pattern. 
There are two peaks during the day. 
The first one starts at noon and ends at 3 p.m., and the second one is from 5 p.m. to 9 p.m. 
The order frequency in the second peak between 5 p.m. and 9 p.m is higher compared to the first peak. 
In the Uskudar region, the duration of the first peak is shorter (between noon to 1 p.m.), and unlike other regions, there is one more peak at 4 p.m.

\section{Results}\label{results}
In our numerical study, we first experiment with eight DQN extensions using our synthetic dataset to determine the best performing algorithm for our MDP model for the meal delivery model. 
We compare these extensions with respect to cumulative collected reward, the number of rejected orders, and the statistics regarding the delivery times. 
We also perform hyperparameter tuning to identify the best configuration for the identified deep RL algorithm. 
\tcolB{Note we also conducted preliminary analysis with standard RL methods such as Q-learning and SARSA, however, these methods did not perform well for our model, and hence omitted from the numerical study.}
Next, we provide results with Getir dataset for three different regions using the best performing model. 
We examine the cumulative collected reward, the number of rejected orders, and the statistics of delivery times for different numbers of available couriers. 
Lastly, we present sensitivity analysis results for various parameters in our models and discuss the algorithmic complexity and convergence. 
We implement all algorithms in Python and perform numerical experiments on a \textcolor{black}{PC with an i7-8700 CPU and 32 GB RAM}. 

\subsection{Comparison of DQN extensions}
We compare eight DQN extensions, namely, DQN, DDQN, DDQN-PER, and Dueling DDQN-PER, considering both hard and soft updates for each algorithm. 
We set the training period as 500 days and test each algorithm for 100 days to prevent possible bias due to randomness in the dataset.
The average number of orders for one day is 163. 
Note that the policy used by Getir in practice resembles the P$_{45}$ policy, which is considered as a baseline in this work.
Table~\ref{tab:duration_stats} presents training times of the algorithms, the average cumulative rewards, the percentage of the delivered orders, and the descriptive statistics for the delivery times that are obtained via simulation over 100 days. 

\textcolor{black}{For this experiment, we consider two approaches for training and testing.
First, a single courier policy is learned via the single courier version of our MDP model, and the corresponding policy is used for each courier in the testing phase in the multiple courier cases.
Second, training is done over the multiple courier version to learn the multi-courier policy, which is then used in the testing phase for the corresponding multiple courier cases. 
We employ five couriers for delivering the orders. 
Table~\ref{tab:duration_stats} shows that the average cumulative reward for the policy generated by DDQN$_{H}^+$ is higher than others for both single courier and multi-courier versions.
When we have 5 couriers for 163 average daily orders, multi-courier policies generated by each deep learning algorithm do not lead to any rejected orders.
On the other hand, DDQN$_{H}^+$ outperforms others thanks to better policies (e.g., by assigning the orders to more suitable couriers).
Overall, all deep learning algorithms outperform the rule-based baselines in terms of the accumulated reward, which shows that a more intelligent system accepts only the orders that contribute the most to the reward (e.g., by minimizing travel distances for the couriers). Also, they assign orders to the most appropriate couriers such that the delivery times are reduced, thus increasing the cumulative reward. 
}

\begin{table}[!ht]
    \centering\caption{\textcolor{black}{Summary results of algorithm performances
    on $10\times10$ synthetic data with five couriers.}}
    \resizebox{\textwidth}{!}{
    \begin{tabular}{l|c|r|r|rrrrr|R{1.3cm}R{1.3cm}R{1.3cm}}
    \toprule
    \multirow{2}{*}{\textbf{Algorithm}} & \textbf{\multirow{2}{*}{\begin{tabular}[c]{@{}c@{}}\textbf{Policy}\\ \textbf{type}\end{tabular}}} &  \multicolumn{1}{c|}{\multirow{2}{*}{\begin{tabular}[c]{@{}c@{}}\textbf{Avg. cum.}\\ \textbf{reward}\end{tabular}}} & \multicolumn{1}{c|}{\multirow{2}{*}{\begin{tabular}[c]{@{}c@{}}\textbf{\% of \tcolK{Rej}.}\\ \textbf{orders}\end{tabular}}} & \multicolumn{5}{c|}{\textbf{Delivery times}} & \multicolumn{3}{c}{\textbf{Order distribution (\%)}} \\ \cline{5-12} 
      & & & & \textbf{Min}&\textbf{Max} &\textbf{Mean} & \textbf{Med.} & \textbf{Std.}&$\leq$ \textbf{25} &\textbf{25-45} &\textbf{45-60} \\
    \hline
    P$_{45}$ & --- & 4249.2 & 0.2 & 6 & 45 & 18.9 & 18 & 6.4 & 87 & 13 & 0 \\
    P$_{60}$ & --- & 4229.6 & 0.0 & 6 & 48 & 19.1 & 18 & 6.7 & 87 & 12 & 1 \\ 
    \hline
    DQN$_H$ & Single & 4445.8 & 4.6 & 6 & 29 & 16.5 & 16 & 4.2 & 99 & 1 & 0 \\
    DQN$_S$ & Single & 4488.1 & 3.2 & 6 & 27 & 16.6 & 17 & 4.2 & 99 & 1 & 0 \\
    DDQN$_H$ & Single & 4449.9 & 4.5 & 6 & 28 & 16.5 & 16 & 4.2 & 99 & 1 & 0 \\
    DDQN$_S$ & Single & 4201.5 & 12.1 & 6 & 27 & 15.7 & 16 & 3.8 & 100 & 0 & 0  \\
    DDQN$_H^+$ & Single & \textbf{4514.9} & 2.0 & 6 & 29 & 16.8 & 17 & 4.4 & 98 & 2 & 0 \\
    DDQN$_S^+$ & Single & 4322.7 & 8.5 & 6 & 28 & 16.1 & 16 & 4.0 & 99 & 1 & 0 \\
    D3QN$_H^+$ & Single & 4457.0 & 3.0 & 6 & 29 & 16.9 & 17 & 4.3 & 99 & 1 & 0 \\
    D3QN$_S^+$ & Single & 4462.5 & 3.1 & 6 & 29 & 16.8 & 17 & 4.3 & 98 & 2 & 0 \\
    \hline
    DQN$_H$ & Multi & 4528.7 & 0.0 & 6 & 35 & 17.3 & 17 & 4.7 & 95 & 5 & 0  \\
    DQN$_S$ & Multi & 4533.8 & 0.0 & 6 & 35 & 17.2 & 17 & 4.7 & 95 & 5 & 0  \\
    DDQN$_H$ & Multi & 4555.3 & 0.0 & 6 & 35 & 17.1 & 17 & 4.7 & 95 & 5 & 0  \\
    DDQN$_S$ & Multi & \textbf{4556.2} & 0.0 & 6 & 35 & 17.1 & 17 & 4.8 & 95 & 5 & 0  \\
    DDQN$_H^+$ & Multi & \textbf{4556.2} & 0.0 & 6 & 35 & 17.1 & 17 & 4.8 & 95 & 5 & 0  \\
    DDQN$_S^+$ & Multi & 4555.9 & 0.0 & 6 & 35 & 17.1 & 17 & 4.7 & 95 & 5 & 0  \\
    D3QN$_H^+$ & Multi & 4532.7 & 0.0 & 6 & 35 & 17.2 & 17 & 4.7 & 95 & 5 & 0  \\
    D3QN$_S^+$ & Multi & 4525.3 & 0.0 & 6 & 33 & 17.3 & 17 & 4.8 & 95 & 5 & 0  \\
     \bottomrule 
    \end{tabular}}
    \label{tab:duration_stats}
    \scriptsize
    \begin{tablenotes}
    \item  \quad Order distributions are reported in minutes; \quad $H$: Hard Update;  \quad $S$: Soft Update;  \quad $+$: with PER; 
    \end{tablenotes}
\end{table}

\textcolor{black}{
The descriptive statistics for the delivery times in Table~\ref{tab:duration_stats} show that P$_{60}$ incurs the most delays in the system by over-assigning a few couriers. 
We observe few overdue orders in P$_{60}$, i.e., delivered after 45 minutes, resulting in a negative penalty. 
}
A more cautious policy, P$_{45}$, accepts all orders whose delivery time is below the threshold of 45 minutes; however, it leads to the highest number of last-minute order deliveries, with 13 percent of them being delivered after exceeding the 25 minutes mark. 
On the other hand, deep learning algorithms, learning that there will be new orders with shorter delivery times in the system in the near future, reject any order that may keep the courier busy for an extended amount of times while resulting in a negligible reward. 
\textcolor{black}{Specifically, DDQN with PER and hard update ($\text{DDQN}_H^+$), which has the best cumulative reward in both types, assign the orders to the couriers such that in the long run, the delivery time decreases while the rejection ratio remains small.}
Overall, based on our analysis with the synthetic dataset, DDQN$_H^+$ is the best performing algorithm for our problem, considering both the cumulative reward and the delivery time statistics. 
Accordingly, we proceed with only DDQN$_H^+$ throughout our remaining numerical study.

We note that having a single courier in model training as an alternative to using multiple couriers leads to an approximation mechanism for the multiple courier cases. 
In the multi-courier approach, the number of couriers in the training and testing phase should be equal. 
On the other hand, in the single courier approach, we can train our model for a single courier, and, in the testing phase, we can use the resulting policy for all couriers in the system as a shared policy. 
We provide summary statistics on the performances of single courier and multi-courier policies 
in Table~\ref{tab:MultiVsSingle}, which presents the collected reward, the percentage of rejected orders, and the descriptive statistics of the delivery times (min, max, mean, median, and std.) obtained from each policy for different grid and order sizes. 
For $10\times 10$ grids with 163 average daily orders, the collected reward of the single courier policy is greater than that of the multi-courier policy for two, three, and four courier cases.
This can be due to the smaller state space of the single courier case that enhances the learning.
That is, considering the same number of days for training (500 days), the single courier approach can better explore and learn its smaller state space. 
We verified our premise by increasing the training epochs for the multi-courier approach and, consequently, bridging the gap between the collected rewards. 
Furthermore, due to a higher number of rejections, delivery times for the single courier approach are shorter than those of the multi-courier ones. 
We note a trade-off between the full exploration of the state space and information loss related to courier interactions. 
We also find that, after three couriers, the multi-courier policy becomes superior in terms of various delivery statistics for $10\times 10$ grids with 163 average daily orders.
The single-courier policy fails to capture the importance of couriers' interaction, leading to a higher cumulative reward for the multi-courier policy despite its larger state space. 
\tcolB{Our analysis with a larger area ($15\times 15$ grid) and higher number of orders (203 daily orders on average) provide similar insights regarding the performance of single courier and multi courier approaches.
We find that single courier policies are superior to those of the multi courier model in terms of collected rewards and various delivery statistics for a varying number of couriers in the system.
We observe that a larger grid size leads to a substantial increase in the number of states, rendering learning higher-quality policies more difficult for the multi courier model.
}

\begin{table}[!ht]
    \centering
    \caption{\textcolor{black}{
    DDQN$_H^+$ results for single and multi courier models for different data instances
    } 
    \label{tab:MultiVsSingle}
    } 
    \resizebox{\textwidth}{!}{
        \begin{tabular}{lccrrrrrrrrr}
        \toprule
            \multirow{3}{*}{\textbf{\begin{tabular}[c]{@{}l@{}}Training\\approach\end{tabular}}} & \multirow{3}{*}{\textbf{\begin{tabular}[c]{@{}c@{}}\# of couriers\\in the testing \\ phase\end{tabular}}} & \multirow{3}{*}{\textbf{\begin{tabular}[c]{@{}c@{}}Time\\(hrs.)\end{tabular}}} & \multirow{3}{*}{\textbf{\begin{tabular}[c]{@{}c@{}}\# of\\orders\end{tabular}}} & \multirow{3}{*}{\textbf{\begin{tabular}[c]{@{}c@{}}Grid\\size\end{tabular}}} & \multirow{3}{*}{\textbf{Reward}} & \multirow{3}{*}{\textbf{\begin{tabular}[c]{@{}c@{}}\% Rej.\\ orders\end{tabular}}} & 
            \multicolumn{5}{c}{\multirow{2}{*}{\textbf{Delivery times}}}\\ \\
            \cmidrule(lr){8-12} 
             &  &  &  &  & & \textbf{} & \textbf{Min} & \textbf{Max} & \textbf{Mean} & \textbf{Median} & \textbf{Std.}\\
        \midrule
            \textbf{\multirow{5}{*}{\begin{tabular}[c]{@{}l@{}}Single\\Courier\end{tabular}}} & 2 couriers & 4.2 & 163 & $10\times10$ & 2,830 & 21.6 & 8 & 40 & 22.9 & 23 & 6.7\\
             & 3 couriers & \dittotikz & 163 & $10\times10$ & 3,870 & 8.1 & 6 & 33 & 19.2 & 19 & 5.3\\
             & 4 couriers & \dittotikz & 163 & $10\times10$ & 4,307 & 3.7 & 6 & 31 & 17.6 & 18 & 4.7\\
             & 5 couriers & \dittotikz & 163 & $10\times10$ & 4,515 & 2.0 & 6 & 29 & 16.8 & 17 & 4.4\\
             & 6 couriers & \dittotikz & 163 & $10\times10$ & 4,590 & 1.6 & 6 & 27 & 16.4 & 16 & 4.3\\
         \hline
        \textbf{\multirow{5}{*}{\begin{tabular}[c]{@{}l@{}}Multi\\Courier\end{tabular}}} & 2 couriers & 5.2 & 163 & $10\times10$ & 1,770 & 18.2 & 8 & 52 & 31.7 & 32 & 10.8\\
         & 3 couriers & 6.3 & 163 & $10\times10$ & 3,359 & 2.6 & 6 & 48 & 23.9 & 22 & 9.3\\
         & 4 couriers & 6.9 & 163 & $10\times10$ & 4,234 & 0.0 & 6 & 48 & 19.1 & 18 & 6.8\\
         & 5 couriers & 7.6 & 163 & $10\times10$ & 4,556 & 0.0 & 6 & 35 & 17.1 & 17 & 4.8\\
         & 6 couriers & 8.3 & 163 & $10\times10$ & 4,610 & 0.0 & 6 & 32 & 16.8 & 17 & 4.5\\
         \toprule
            \textbf{\multirow{5}{*}{\begin{tabular}[c]{@{}l@{}}Single\\ Courier\end{tabular}}} & 2 couriers & 7.7 & 203 & $15\times15$ & 2,239 & 46.6 & 8 & 43 & 24.4 & 24 & 6.1\\
             & 3 couriers & \dittotikz & 203 & $15\times15$ & 3,238 & 30.7 & 7 & 43 & 22.0 & 22 & 5.2\\
             & 4 couriers & \dittotikz & 203 & $15\times15$ & 3,879 & 21.9 & 7 & 34 & 20.6 & 21 & 4.8\\
             & 5 couriers & \dittotikz & 203 & $15\times15$ & 4,330 & 15.8 & 6 & 30 & 19.7 & 20 & 4.6\\
             & 6 couriers & \dittotikz & 203 & $15\times15$ & 4,590 & 13.5 & 6 & 29 & 18.9 & 19 & 4.5\\
         \hline
           \textbf{\multirow{5}{*}{\begin{tabular}[c]{@{}l@{}}Multi\\Courier\end{tabular}}} & 2 couriers & 9.3 & 203 & $15\times15$ & 1,372 & 43.3 & 8 & 50 & 33.1 & 33 & 8.9\\
             & 3 couriers & 11.3 & 203 & $15\times15$ & 1,791 & 24.7 & 8 & 50 & 33.3 & 35 & 9.5\\
             & 4 couriers & 12.6 & 203 & $15\times15$ & 2,352 & 11.2 & 7 & 55 & 32.0 & 32 & 10.8\\
             & 5 couriers & 13.8 & 203 & $15\times15$ & 3,263 & 3.7 & 7 & 64 & 28.3 & 25 & 11.9\\
             & 6 couriers & 15.1 & 203 & $15\times15$ & 4,218 & 0.4 & 7 & 60 & 24.2 & 22 & 9.0\\
         \bottomrule
        \end{tabular}
        }
\end{table}

We observe that the training times of the single courier approach are significantly shorter (e.g., 4 hours and 12 minutes for $10\times 10$ grid) than those of the multi-courier case (approximately from 6 to 10 hours for the same grid size).
Moreover, in the multi-courier case, the training needs to be performed whenever the number of couriers changes. 
In other words, we cannot directly use the policy generated for two couriers and test it for three couriers due to incompatibility in their state spaces. 
Thus, the single courier approach can be beneficial in terms of generating the policies faster without a substantial compromise in rewards. 
It also facilitates handling a varying number of couriers throughout a day for the company without the need for retraining. 
As such, we proceed with the single courier approach in the rest of the experiments, noting that it is an approximation and provides a theoretical lower-bound for the reward obtained by the multi-courier approach. 

\subsection{Hyperparameter tuning}
We originally trained all DQN extensions on the synthetic data with the default hyperparameters (see Table~\ref{tab:parameters}). 
We next provide our observations from the hyperparameter tuning experiments with DDQN$_H^+$.
We consider tuning the following hyperparameters: experience replay memory ($\ERMemory$), discounting factor ($\discount$), batch size in the network ($\batchSize$), and the number of steps after which the target network is updated ($\stepsToUpdateTargetNet$).
Experience replay memory is a variable that stores the agent's experiences (i.e., state, action, reward, and next state) at each time step.
The discount factor affects the weight given to the future rewards in the subsequent states relative to maximizing the immediate rewards.
The batch size corresponds to the number of samples propagated through the network. 
Lastly, the target network weights are constrained to change slowly, thus improving the stability of the learning and convergence results. Parameter $\stepsToUpdateTargetNet$ control the update frequency for the target network.
If the soft update is used as an update strategy, the degree of the weights coming from the target network is $\softUpdateParam = 0.5$. 
For each hyperparameter, we chose two values; therefore, we have 16 different configurations in total. Table~\ref{tab:hyper} presents the hyperparameter configurations along with their cumulative reward, number of rejected orders, and delivery times statistics. 
Training times for different hyperparameter configurations are very similar; therefore, we did not report them.

\begin{table}[!ht]
    \centering\caption{Hyperparameter tuning results for DDQN$_H^+$ using synthetic $10 \times 10$ grid.}
    \resizebox{\textwidth}{!}{\begin{tabular}{cccc|rrrrrrr}
    \toprule
    \multicolumn{4}{c|}{\textbf{Hyperparameters}} &&\multicolumn{1}{c}{\multirow{2}{*}{\begin{tabular}[c]{@{}c@{}}\textbf{\% Rej.}\\ \textbf{Orders}\end{tabular}}}&\multicolumn{5}{c}{\textbf{Delivery Times}}\\
\cline{7-11}
         $\mathcal{M}$ & $\gamma$ & $\mathcal{B}$ & $\mathcal{U}$ & \textbf{Reward} &  & \textbf{Min} & \textbf{Max} & \textbf{Mean} & \textbf{Median} & \textbf{Std. Dev.} \\
         \midrule
          \textbf{20,000}& \textbf{0.9}& \textbf{128}& \textbf{100}&\textbf{2795.7}&21&7&31&19.9&\textbf{20}&4.7\\
          20,000& 0.9& 128& 200&2780.4&18&7&38&21.4&21&5.7\\
          20,000& 0.9& 64& 100&2783.9&20&7&37&20.6&21&5.3\\
          20,000& 0.9& 64& 200&2780.1&19&7&41&21.1&21&5.4\\
          30,000& 0.9& 128& 100&2741.2&18&7&37&21.8&22&5.8\\
          30,000& 0.9& 128& 200&2778.5&20&7&36&20.5&21&5.0\\
          30,000& 0.9& 64& 100&2775.4&19&7&41&20.9&21&5.4\\
          \textbf{30,000}&\textbf{0.9}&\textbf{64}&\textbf{200}&2779.0&21&7&\textbf{30}&\textbf{19.5}&\textbf{20}&\textbf{4.5}\\
          \midrule
          20,000& 0.1& 128& 100&1001.2&12&7&60&36.8&39&15.5\\
          20,000& 0.1& 128& 200&998.4&12&7&60&36.8&39&15.6\\
          20,000& 0.1& 64& 100&980.3&12&7&60&37.0&40&15.6\\
          20,000& 0.1& 64& 200&1005.2&12&7&60&36.7&39&15.5\\
          30,000& 0.1& 128& 100&998.6&12&7&60&36.8&39&15.6\\
          30,000& 0.1& 128& 200&1013.5&12&7&60&36.7&39&15.5\\
          30,000& 0.1& 64& 100&974.8&12&7&60&37.0&39&15.7\\
          30,000& 0.1& 64& 200&1084.1&12&7&59&36.1&38&15.1\\
         \bottomrule
    \end{tabular}}
    \begin{tablenotes}
      \footnotesize
      \item \quad $\mathcal{M}$: Experience replay memory size
      \item \quad $\gamma$: Discount factor in Bellman Equation
      \item \quad $\mathcal{B}$: Batch size in neural network 
      \item \quad $\mathcal{U}$: The number of steps after which the target network is updated
    \end{tablenotes}
    
    \label{tab:hyper}
    
\end{table}   

We chose 20,000 and 30,000 for the experience replay memory size in order to observe whether increasing the memory size also leads to an increase in the performance. 
However, we did not observe any general relation in terms of the change in memory size. 
We used 0.9 and 0.1 as the discount factor. 
A discount factor close to zero means that the immediate reward is much more important than the next states' reward. 
If it is close to one, then it means that the algorithm weighs the future rewards more. 
We noted a significant effect of using a higher discount factor in our case. While its impact on the rewards is evident (i.e., lower $\discount$ leads to lower rewards), we observed a substantial effect on other outcomes such as \% rejected orders and delivery times.
Furthermore, we chose 128 and 64 for the batch size parameter. 
We did not observe any significant trend for the batch size value on its own; however, its effect should be investigated together with other parameters. 
Lastly, we set the step size in which the target network is updated to 100 and 200. 
Again, the step size by itself did not seem to contribute to the higher reward, and its value becomes more important when considered with other hyperparameters. 
For instance, if the memory size is 30,000, we may obtain higher rewards if we update the target network every 200 steps. 

We determined that the hyperparameter values of $\mathcal{M} = 20,000$, $\gamma = 0.9$, $\mathcal{B} = 128$, and $\mathcal{U} = 100$ collected the highest reward (2795.7) among all other configurations. The hyperparameter values 30,000, 0.9, 64, and 200 for $\mathcal{M}$, $\gamma$, $\mathcal{B}$, and $\mathcal{U}$, respectively, attained the shortest delivery times among all other configurations. However, the delivery time statistics of the first configuration are almost identical to this configuration. Therefore, we use the first hyperparameter configuration to assess the performance of DDQN$_H^+$ on the Getir dataset.

\subsection{Model outcomes for the Getir dataset}
We trained the DDQN$_H^+$ algorithm for 1000 days using the Getir dataset involving three regions of Istanbul, namely, Bomonti, Uskudar, and Hisarustu. We compared the policy generated by the DDQN$_H^+$ algorithm (P$_{\text{DDQN}_H^+}$) with the baseline policies (P$_{45}$ and P$_{60}$) based on the cumulative reward and the number of rejected orders (see Table~\ref{tab:Reward}). 

\begin{table}[!ht]
\centering
\caption{Performance comparison of the RL and baseline policies for the regions with different number of couriers} 
\resizebox{\textwidth}{!}{\begin{tabular}{llcrrrcrrrcrrr}
\toprule
&  & \multicolumn{3}{c}{\textbf{Bomonti}} &\ & \multicolumn{3}{c}{\textbf{Uskudar}} & \ & \multicolumn{3}{c}{\textbf{Hisarustu}} \\
\cline{3-13}
 &  & P$_{\text{DDQN}_H^+}$ & P$_{45}$ & P$_{60}$ && P$_{\text{DDQN}_H^+}$ & P$_{45}$ & P$_{60}$ && P$_{\text{DDQN}_H^+}$ & P$_{45}$ & P$_{60}$ \\
 \midrule
\textbf{\multirow{5}{*}{\begin{tabular}[c]{@{}l@{}}Cum.\\ reward\end{tabular}}}
& 3 couriers & 3144.8 & 887.8 & -957.6 && 2215.2 & 958.6 & -422.0 && 2557.9 & 1794.8 & 830.7 \\
 & 4 couriers &3860.1  &2145.9  & 280.2 && 2702.6 & 1918.1 & 726.0 && 2945.1 & 2551.6 & 2135.1 \\
 & 5 couriers & 4309.1 & 3279.2 & 1911.0 && 3063.9 & 2739.1 &1990.5  &&  3113.1& 2988.0 & 2909.5 \\
 & 6 couriers & 4631.6 &4187.6  &3473.4  && 3238.4 & 3104.2 & 2693.5 && 3286.5 & 3210.9 & 3191.1 \\
 & 7 couriers & 4864.7 &4724.2  & 4502.6 && 3348.5 & 3315.1 & 3193.9 && 3326.1 & 3316.6 &3322.3  \\
 \midrule
\textbf{\multirow{5}{*}{\begin{tabular}[c]{@{}l@{}}Rej.\\ orders\end{tabular}}} & 3 couriers & 71.0  &63.5 & 61.3 && 46.6 & 45.8 & 42.6 && 31.0 & 20.1 & 15.7 \\
 & 4 couriers &46.0  & 35.8 & 33.1 && 29.5 & 25.4 & 19.7 && 20.1 & 7.5 & 3.6 \\
 & 5 couriers &31.9  & 22.1 &14.4  && 19.5 & 13.3 & 7.6 &&16.3  & 3.2 &0.7 \\
 & 6 couriers & 20.5 & 7.8 & 4.0 && 15.8 & 7.9 & 2.4 && 14.7 & 1.9 &  0\\
 & 7 couriers & 14.3 &3.3  & 0.7 && 13.0 & 5.2 & 0.3 && 13.7 & 1.0 & 0 \\
 \bottomrule
\end{tabular}}
\label{tab:Reward}
\end{table}

The results reported in Table~\ref{tab:Reward} obtained by evaluating the policies for a given number of couriers over a test dataset for 100 days. 
Note that each region has different daily order frequencies. 
Specifically, we tested each policy on average daily order amounts of 141, 172, and 220 for the Hisarustu, Uskudar, and Bomonti regions, respectively. 

We observe that the policy generated by $\text{DDQN}_H^+$ generates higher rewards than baseline policies for each region. 
As the number of couriers increases, we observe a plateau effect in the collected reward. 
For all the regions, after five couriers, the marginal gain of adding one more courier to the system considerably decreases. 
Considering the employment costs of the couriers, we can observe that adding an extra courier might result in more losses than earnings. 
For instance, in Hisarustu, if we use seven couriers, they would dispatch only one more order compared to six couriers, and the increase in the collected reward is less than 2\%.
Consequently, finding the ideal number of couriers is still an open question in such problems.

\begin{figure*}[!ht]
  \centering
\begin{subfigure}[t]{.9\textwidth}
    \centering\includegraphics[width=\textwidth]{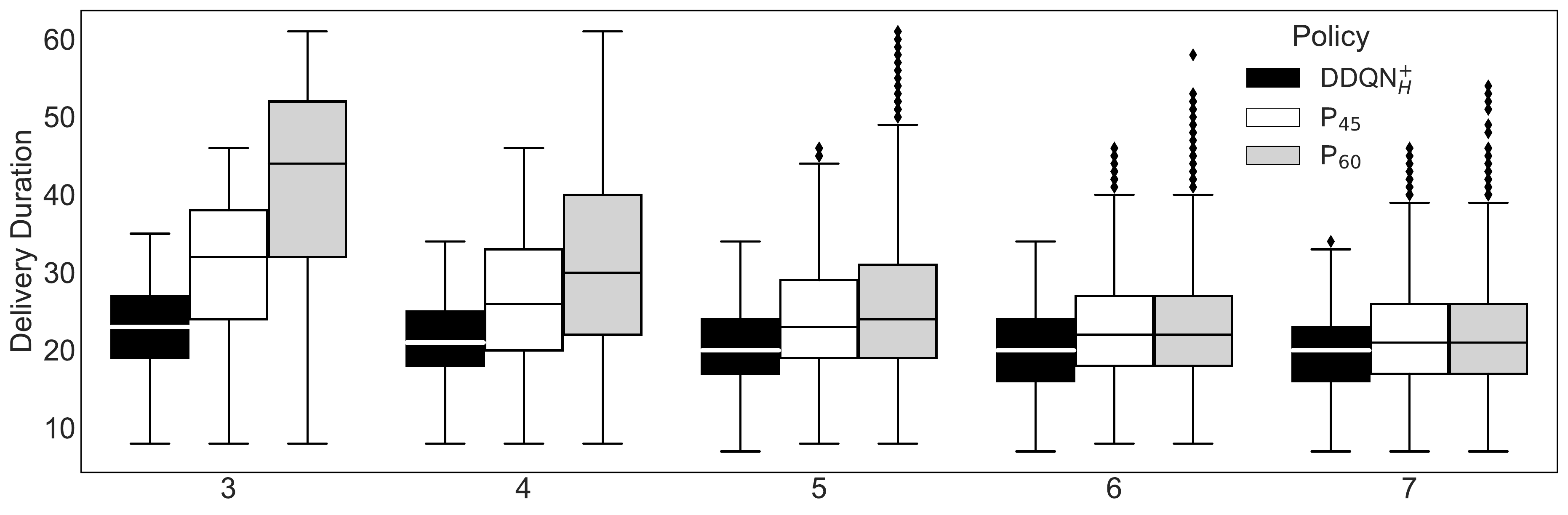}\vspace{-0.3cm}
    \caption{Bomonti} \label{fig:hisarustu_del}
    \vspace{0.2cm}
  \end{subfigure}\quad
  \begin{subfigure}[t]{.9\textwidth}
    \centering\includegraphics[width=\textwidth]{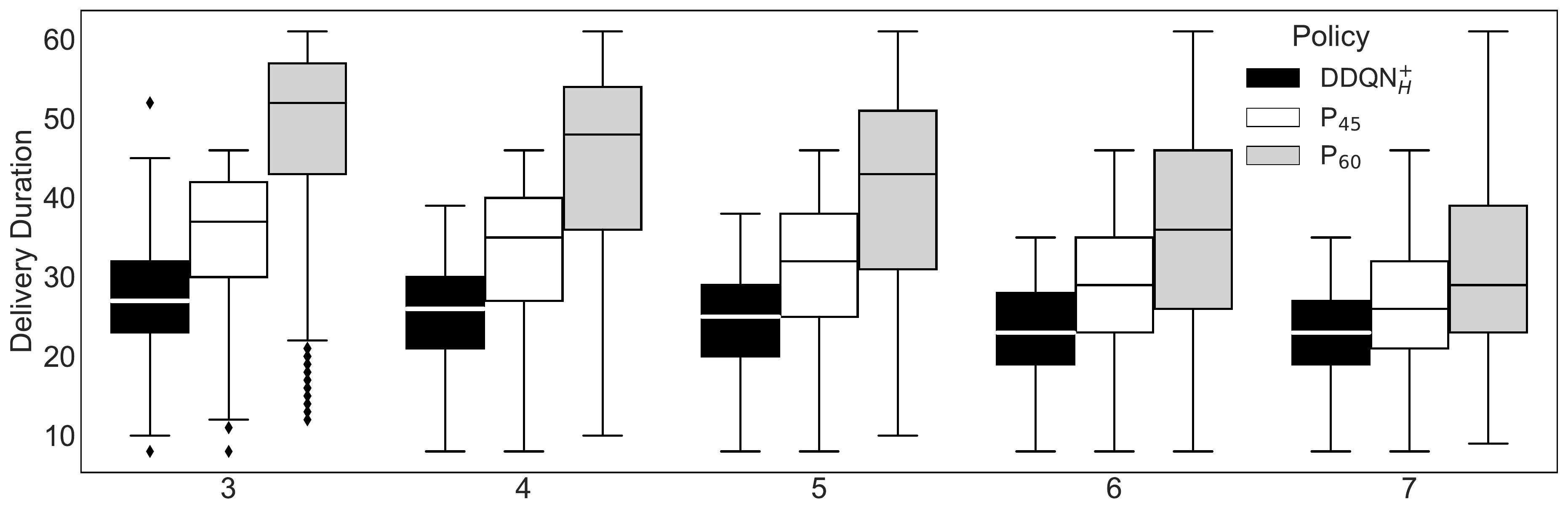}\vspace{-0.3cm}
    \caption{Uskudar} \label{fig:uskudar_del}
    \vspace{0.2cm}
  \end{subfigure}
  \begin{subfigure}[t]{.9\textwidth}
    \centering\includegraphics[width=\textwidth]{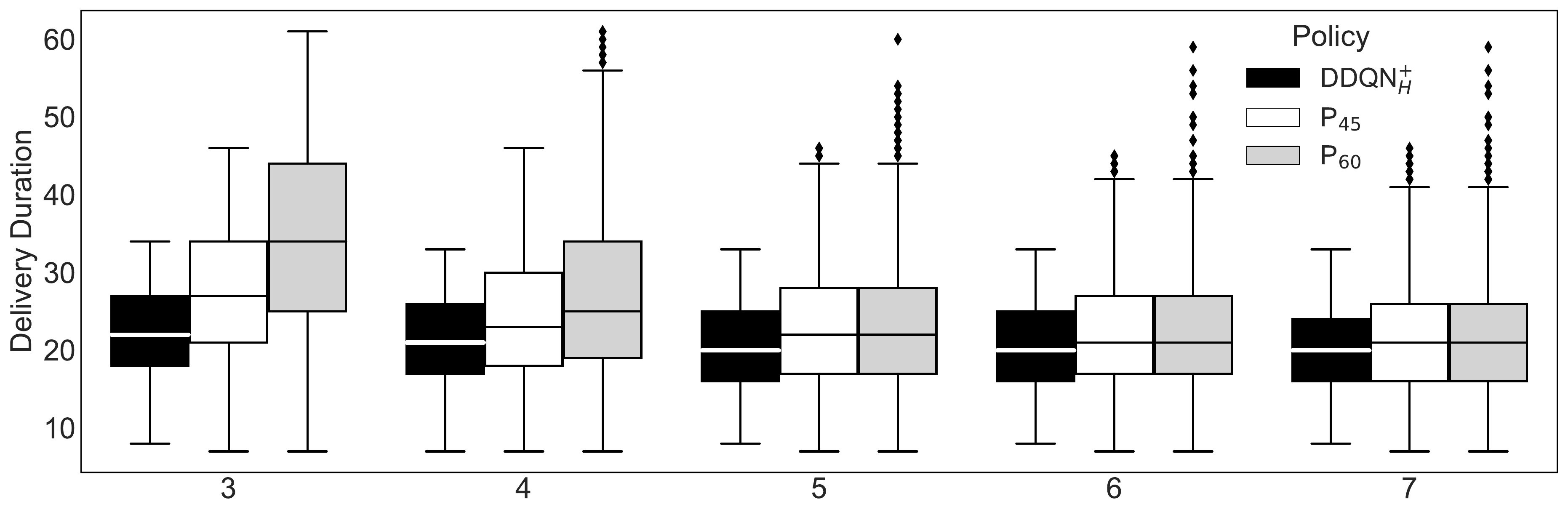}\vspace{-0.3cm}
    \caption{Hisarustu} \label{fig:bomonti_del}
  \end{subfigure}
  ~
    \caption{Delivery times of the orders in minutes for different numbers of couriers per policy.} \label{fig:regions_del_times}
\end{figure*}

We present delivery times of the orders in Figure~\ref{fig:regions_del_times} for each region. 
In the case of fewer couriers, such as three or four, we observe that the differences between the median delivery times are significant among the policies. 
For instance, in Bomonti, when there are three couriers in the system, the median delivery times are 23, 32, and 44 for $\text{DDQN}_H^+$, P$_{45}$, and P$_{60}$, respectively. 
Especially for the regions having a higher number of daily orders, this result is more noticeable. 
However, when we have six or seven couriers in the system, even the median values get closer to each other, noting that it is not justifiable to use these many couriers because they cause an increase in the overall cost. 

\begin{figure}[!ht]
\begin{center}
\subfloat[3 Couriers]{\includegraphics[width=0.32\textwidth]{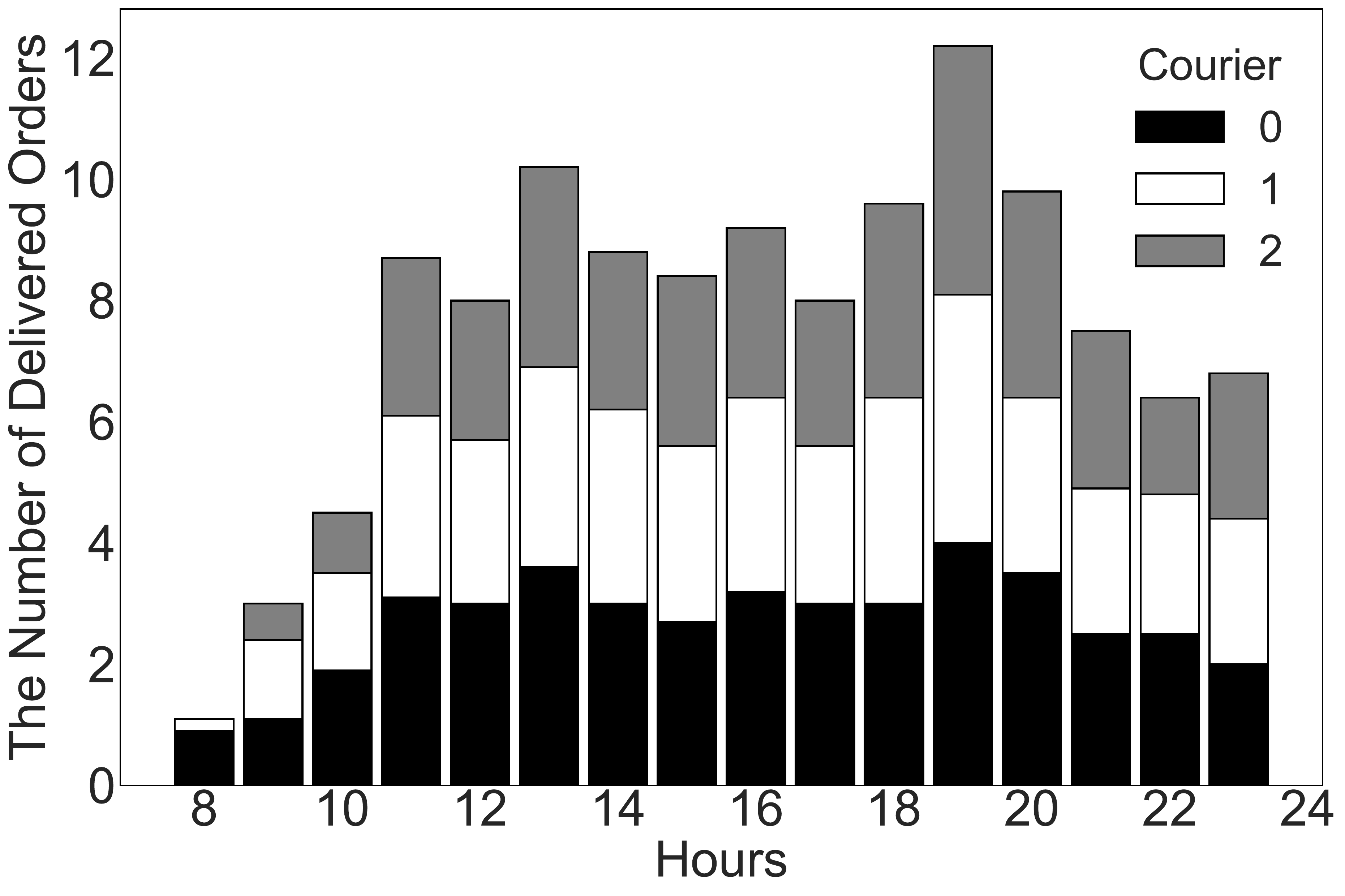}}
\subfloat[4 Couriers]{\includegraphics[width=0.32\textwidth]{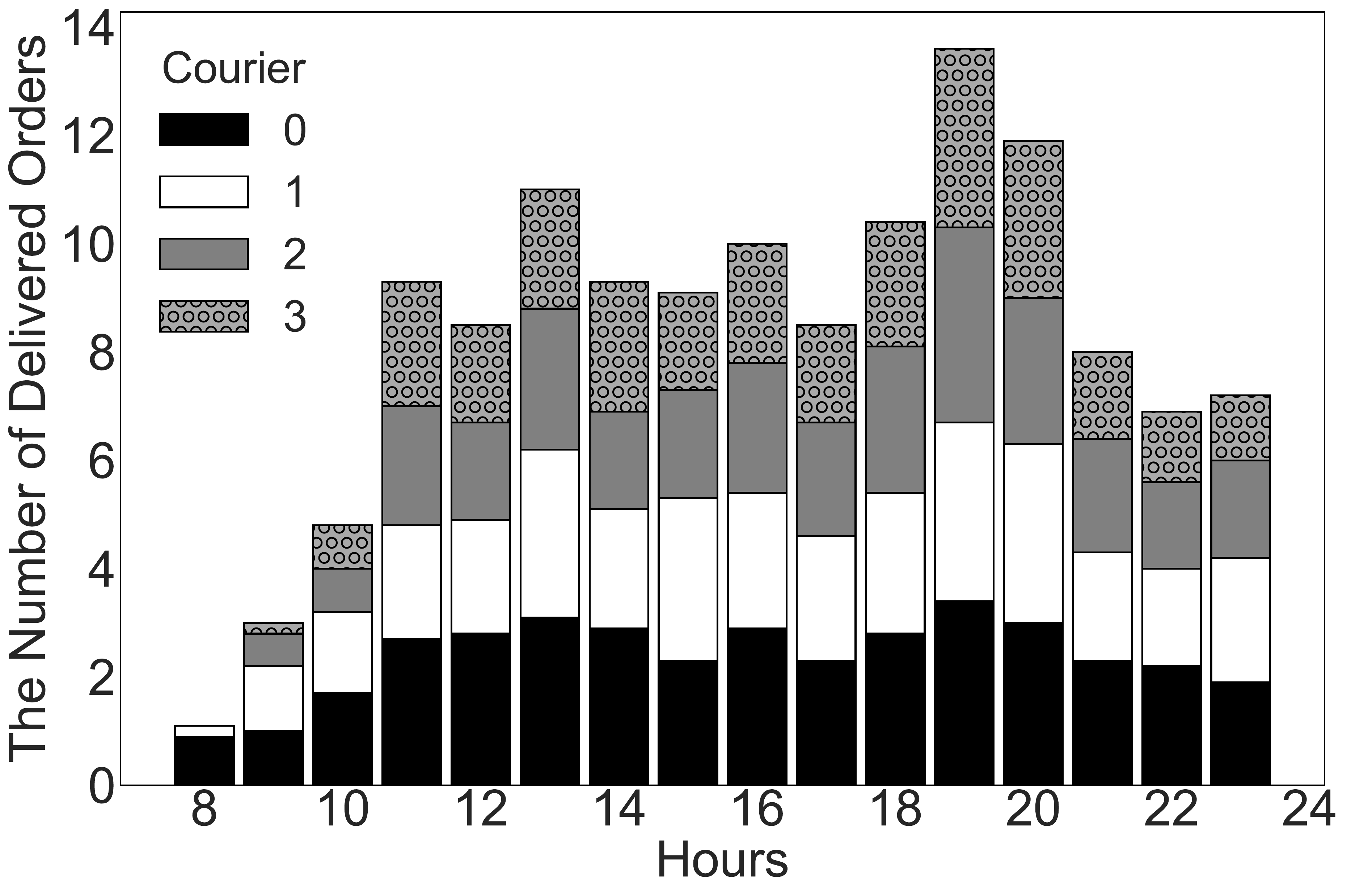}}
\subfloat[5 Couriers]{\includegraphics[width=0.32\textwidth]{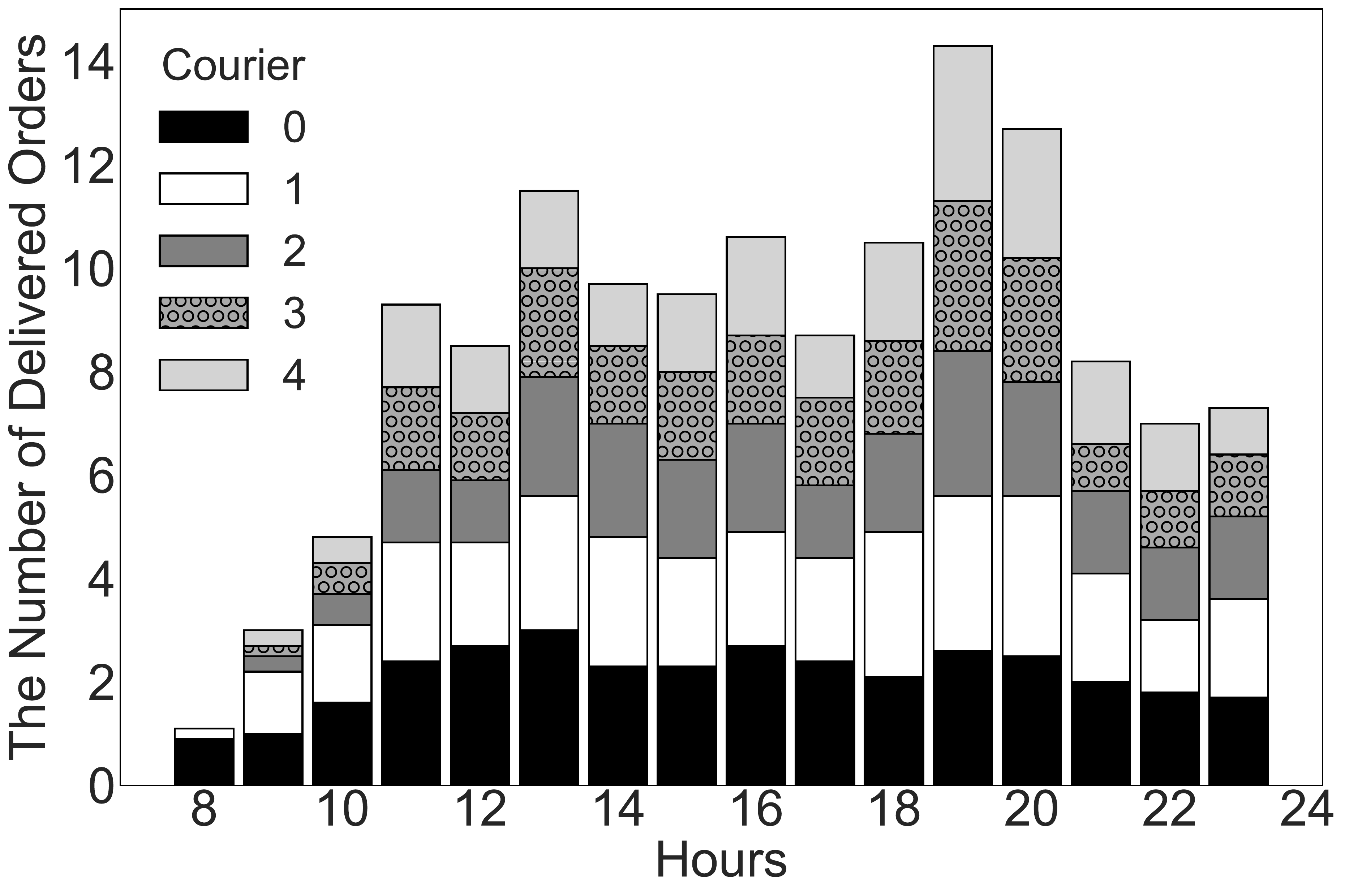}}\\
\subfloat[6 Couriers]{\includegraphics[width=0.32\textwidth]{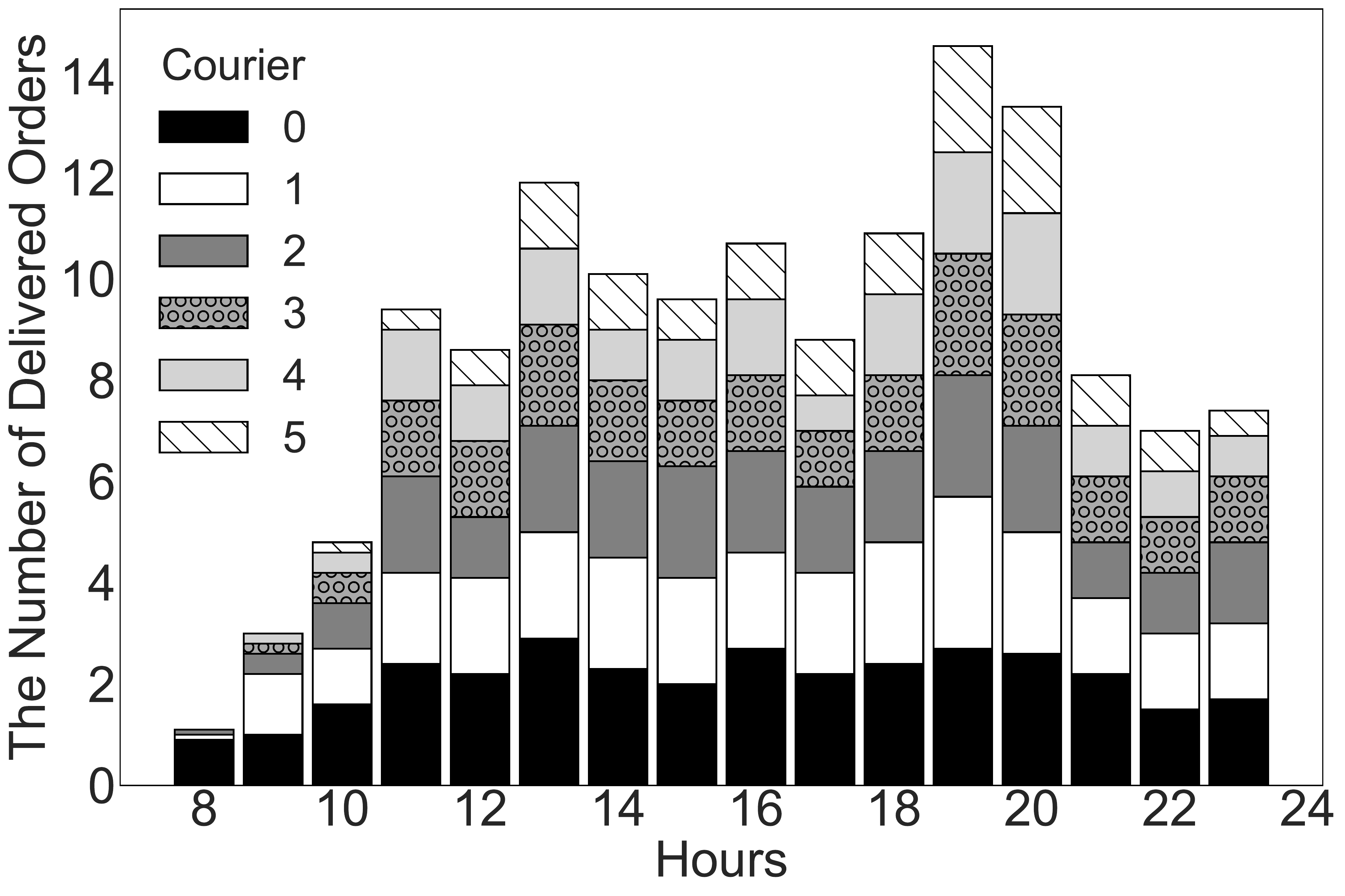}}\quad
\subfloat[7 Couriers]{\includegraphics[width=0.32\textwidth]{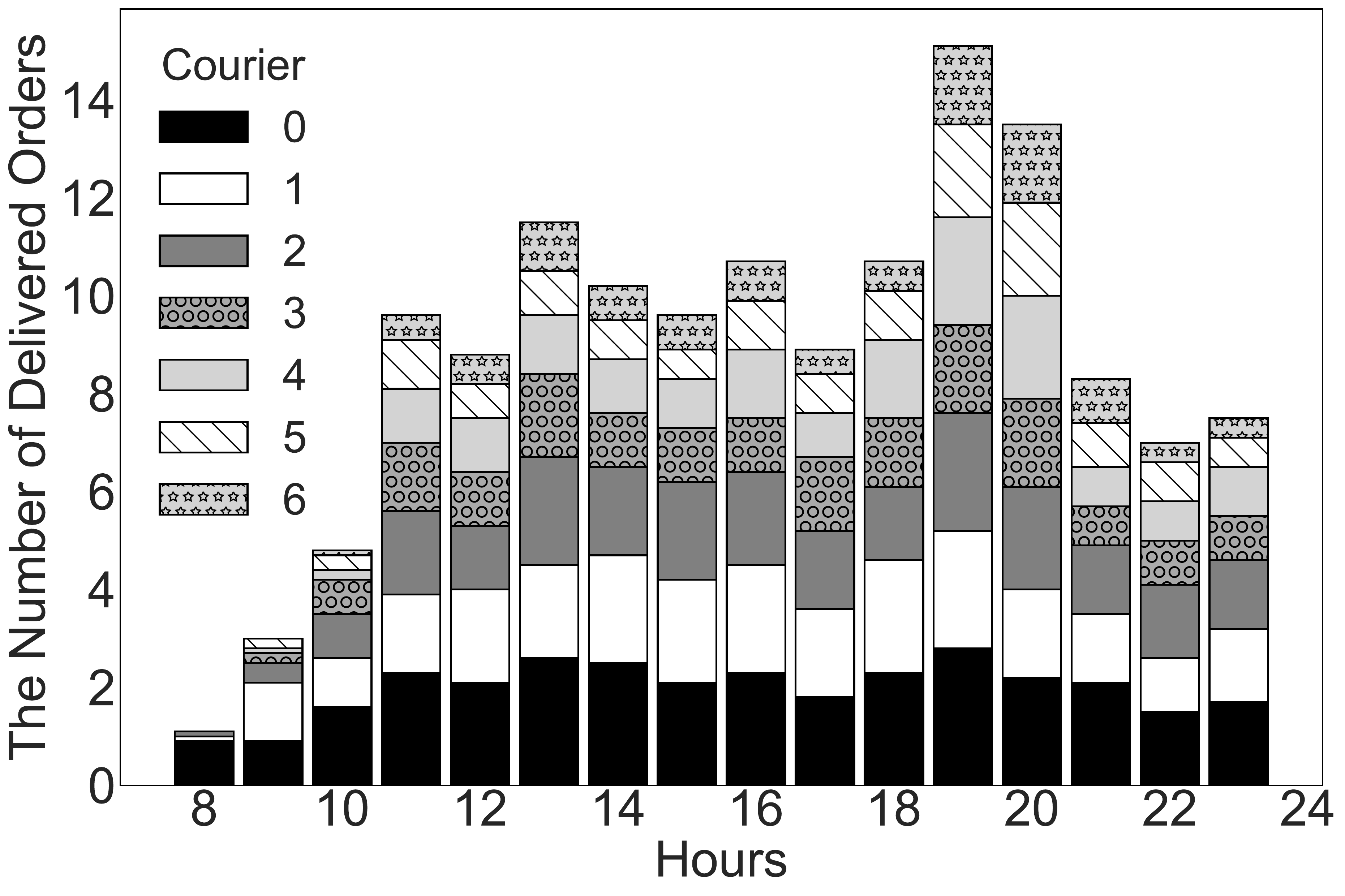}}
\caption{The hourly number of delivered orders in Bomonti region}
\label{fig:hourly_del_orders}
\end{center}
\end{figure}

We also investigate the number of delivered orders in each hour of a day by changing the number of couriers (see Figure~\ref{fig:hourly_del_orders}). 
We note that the system does not use all couriers at each hour. 
For instance, when the maximum hourly number of couriers is seven, it does not assign orders to all of them between 8 a.m. and 11 a.m. 
Even when there are at most three couriers in the system, it only uses one of them to dispatch the orders when the number of orders is small. 
Hence, these findings may help decision-makers to decide on the optimal number of couriers in each hour. 
Another issue is the maximum number of delivered orders for one day for different numbers of couriers. 
When the number of couriers is six or seven in the system, the maximum number of delivered orders in one hour does not change significantly, which means that after five couriers, hiring a new courier cannot increase the number of delivered orders in one hour. 
In other words, the maximum orders received in an hour can be delivered with fewer couriers.

\begin{figure}[!ht]
\begin{center}
\subfloat[3 Couriers]{\includegraphics[width=0.31\textwidth]{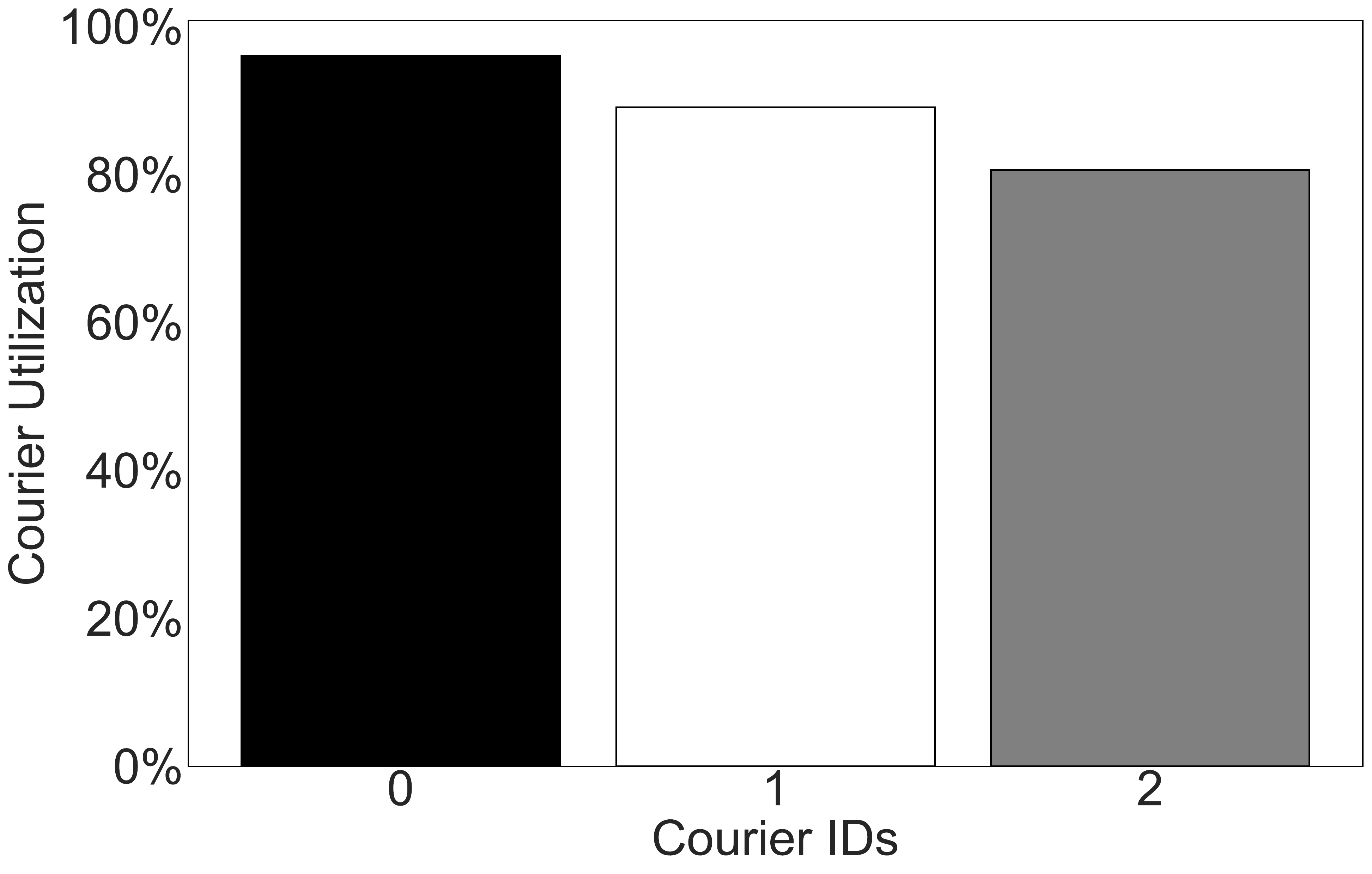}}
\subfloat[4 Couriers]{\includegraphics[width=0.31\textwidth]{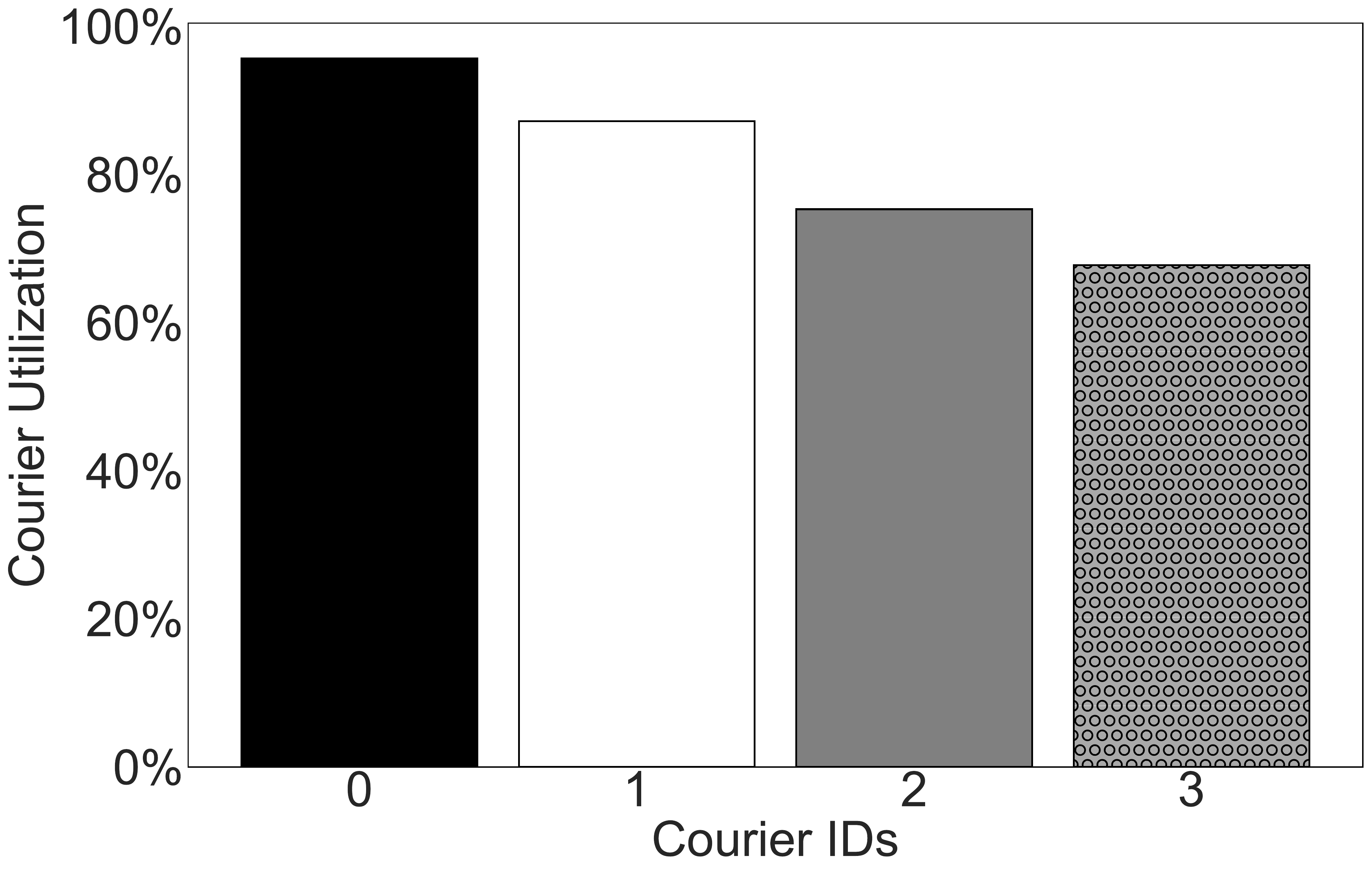}}
\subfloat[5 Couriers]{\includegraphics[width=0.31\textwidth]{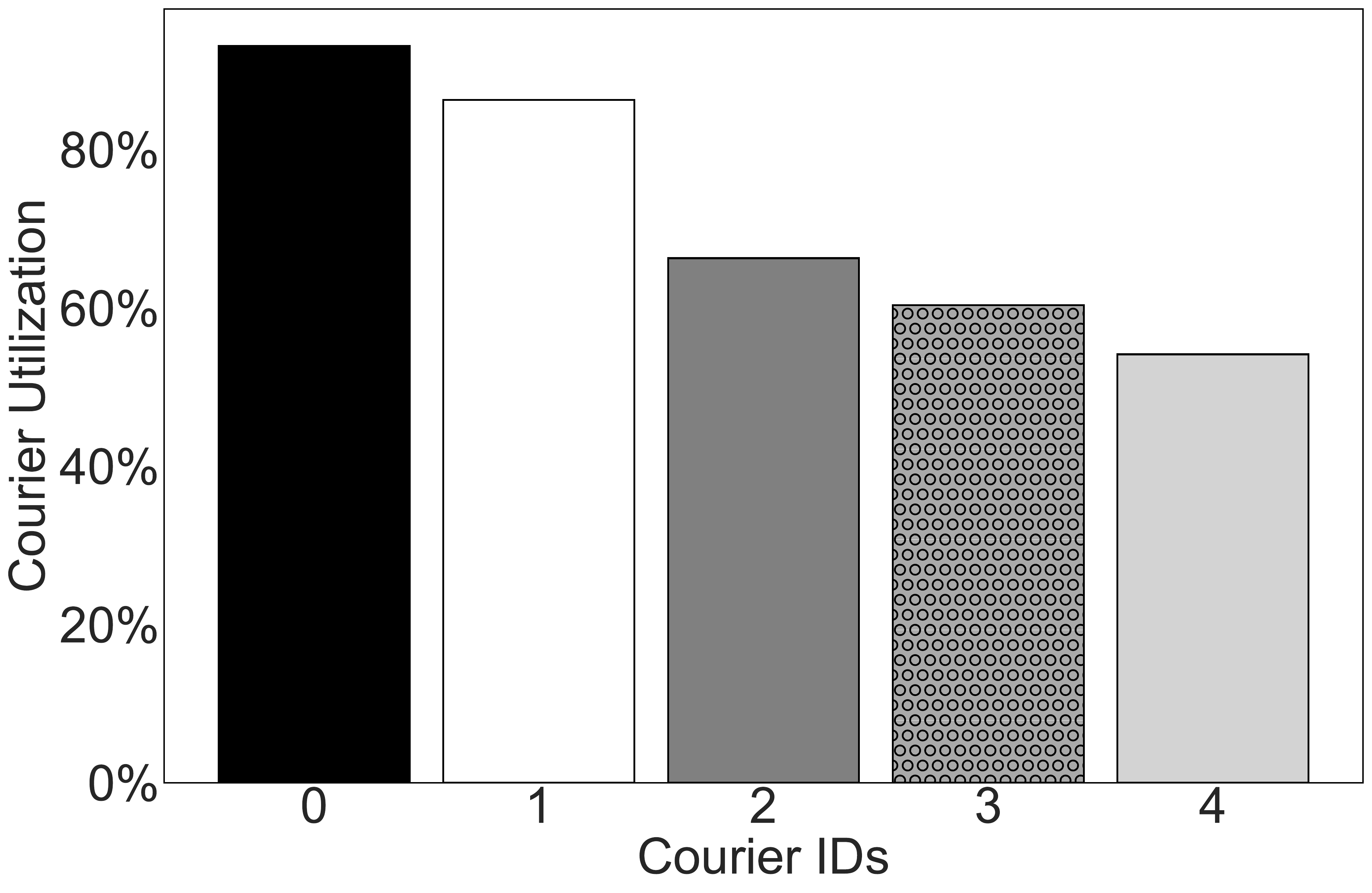}}\\
\subfloat[6 Couriers]{\includegraphics[width=0.31\textwidth]{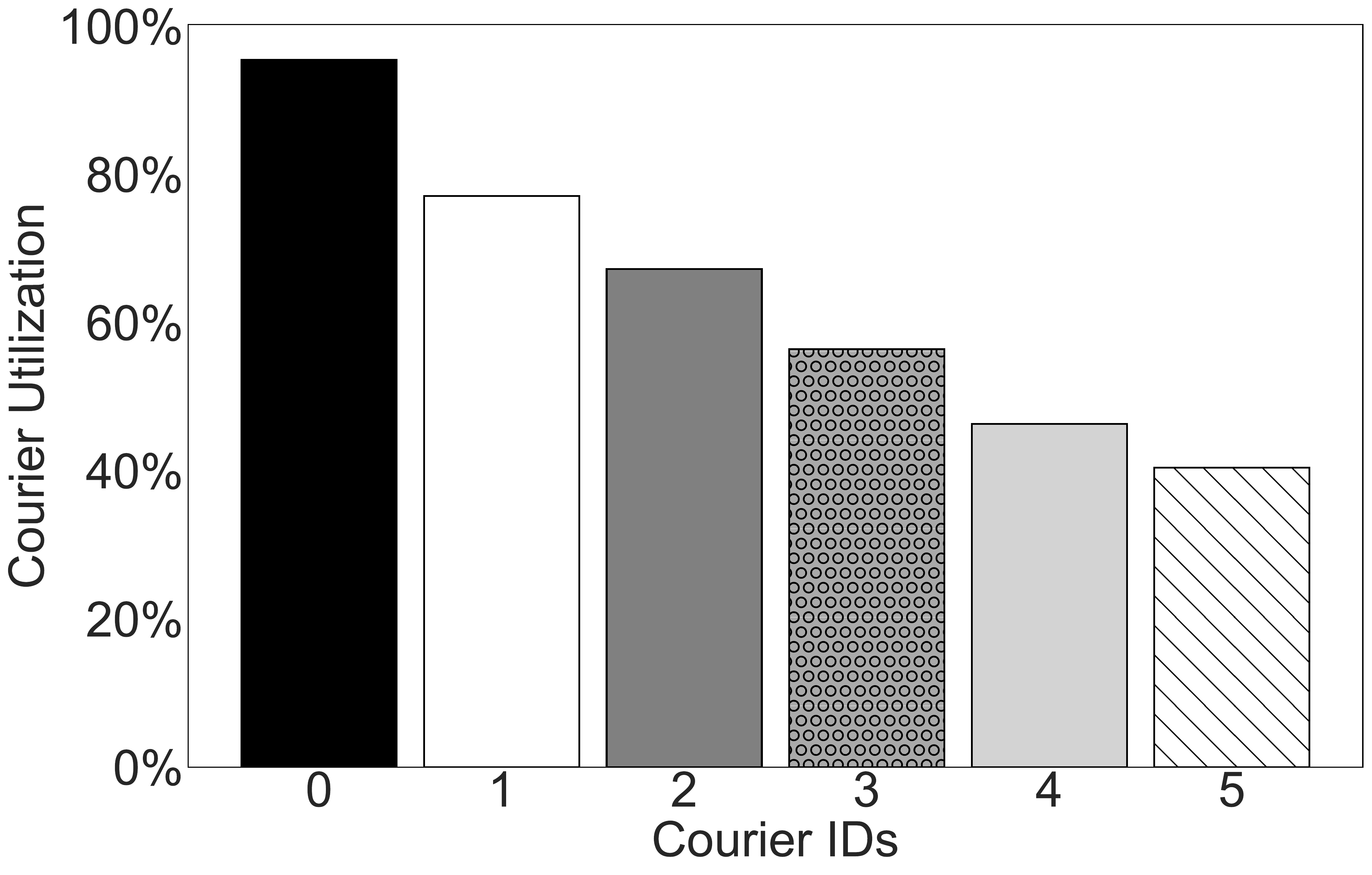}}\quad
\subfloat[7 Couriers]{\includegraphics[width=0.31\textwidth]{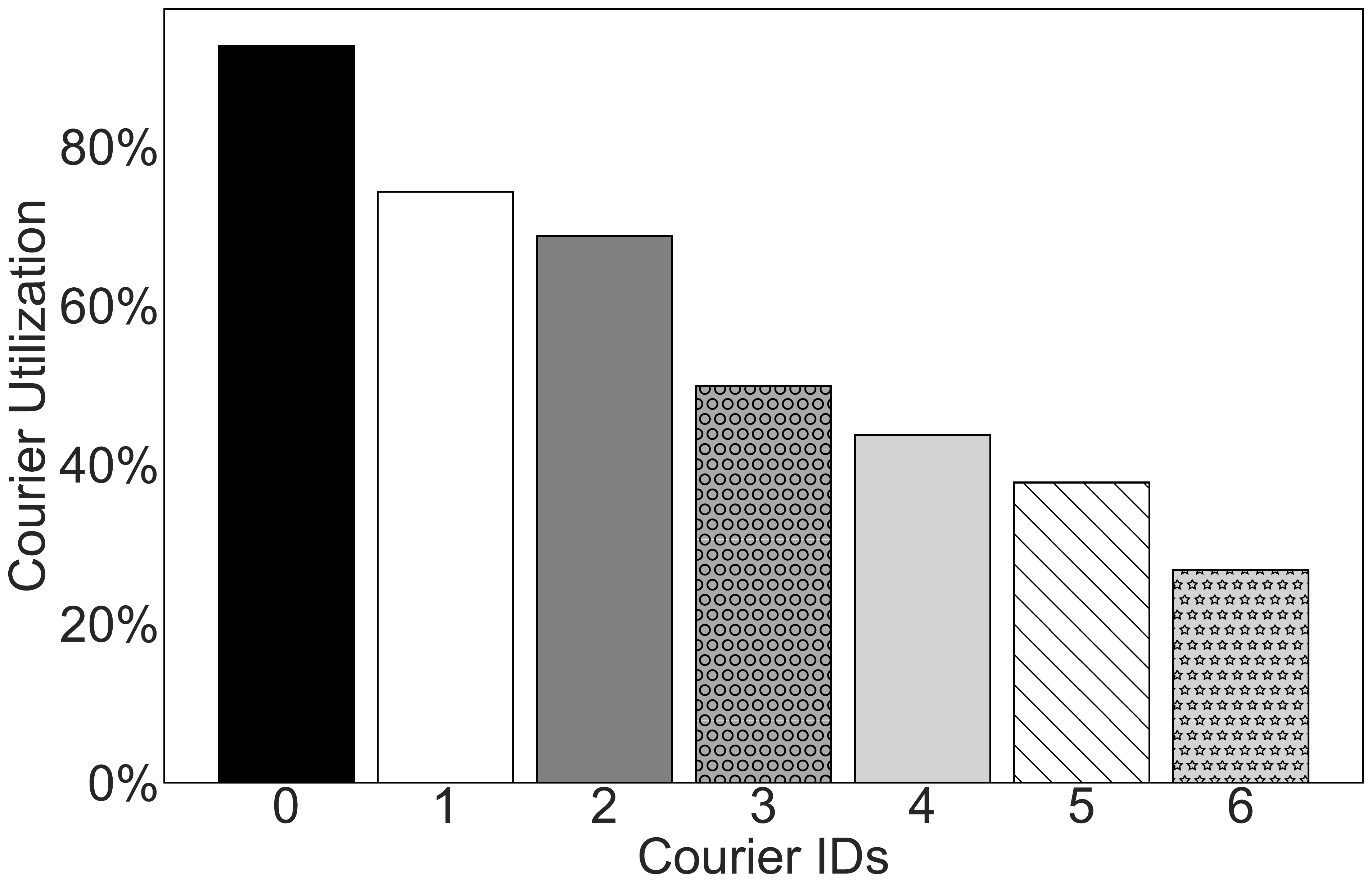}}
\caption{Courier utilization in Bomonti region}
\label{fig:utilization}
\end{center}
\end{figure}

Figure~\ref{fig:utilization} presents the utilization of each courier in one of the regions (Bomonti) when the company adopts the policies generated by ${DDQN_H^+}$. 
When we have up to four couriers in the system, the minimum utilization is more than 70\%, which is the utilization threshold for the company. 
However, the average utilization decreases dramatically for a higher number of couriers.
Besides, we observe an imbalanced distribution of orders among couriers, which implies that some couriers cannot be used efficiently and are not on duty for a significant amount of time, leading to lower overall utilization rates. 
\tcolB{Note that we also examined whether using the multi courier version of the model would lead to a noticeable difference in courier utilizations.
Our preliminary analysis showed that the mean and standard deviation values for the work hours of the couriers are largely the same for both models.}

\subsection{Sensitivity analysis}
We perform a sensitivity analysis to assess the robustness of our model with respect to the order arrival rate and reward function parameters.
Specifically, the orders are generated using a Poisson distribution with parameter $\paramOrderDistn_{\indHour}$ for each hour $\indHour$, which are predetermined as in synthetic dataset or obtained from the Getir dataset (see Figure~\ref{fig:order_percentage}). Let $\orderCountDaily$ be the daily order count, and $\paramOrderDistnPercentage_{\indHour}$ percentage of daily orders at hour $\indHour$. Then, the hourly arrival rate can be obtained as 
\begin{equation*}
   \paramOrderDistn_{\indHour} = \lfloor\paramOrderDistnPercentage_{\indHour} \times \orderCountDaily\rfloor.
\end{equation*}
We consider varying daily order counts as $\orderCountDaily \in \{120, 170, 220\}$ in our sensitivity analysis.
Note that distribution of randomly generated orders mimic the behavior of actual orders and follow the same peak times and low density hours.


In our reward function (see Equation~\ref{eq:reward}), the main component is defined as ${+45-\timeDeliver_{\indCourier}^{\indOrder}}$ if $\indAction = \actionAssign_{\indCourier}^{\indOrder}$. That is, if the order is assigned to a courier, then the collected reward is 45 minutes (target upper bound for delivery time) minus the actual delivery time. We experiment with three different values, i.e., 30, 45 and 60, for the target delivery times. 
Note that if we use the target delivery time of 30 minutes for the reward function parameter, $RP$, instead of 45 minutes, the reward of the orders having a delivery time of more than 30 minutes will be negative. 
Similarly, if we take this parameter as 60 minutes, then the orders having a delivery time of more than 60 minutes will be penalized. 

We conduct a two-way sensitivity analysis with these parameters.
We train $\text{DDQN}_H^+$ by using each configuration of these parameters. 
We compare resulting policies according to the cumulative reward, the number of rejected orders, and the descriptive statistics of the delivery times, as shown in Table~\ref{tab:sensitivity} for the representative region, Bomonti.

\begin{table}[!ht]
    \centering\caption{Sensitivity analysis results}
    \resizebox{0.9\textwidth}{!}{
    \-\hspace{1.3cm}\begin{tabular}{ccrrrrrrr}
    \toprule
    &&\multicolumn{1}{c}{\multirow{2}{*}{\begin{tabular}[c]{@{}r@{}}\textbf{Avg. cum.}\\ \textbf{reward}\end{tabular}}}&\multicolumn{1}{c}{\multirow{2}{*}{\begin{tabular}[c]{@{}r@{}}\textbf{Avg. \% of}\\ \textbf{rej. orders}\end{tabular}}}&\multicolumn{5}{c}{\textbf{Delivery Times}}\\
\cline{5-9}
         \textbf{$RP$} & $\orderCountDaily$ & &  & \textbf{Min}&\textbf{Max}&\textbf{Mean}&\textbf{Median}&\textbf{Std.}\\
         \midrule
            30 & 120 & 2577.0 & 15 & 8 & 34 & 20.4 & 20 & 5.1 \\ 
            30 & 170 & 3178.2 & 32 & 7 & 27 & 19.7 & 20 & 4.3 \\
            30 & 220 & 3956.7 & 37 & 8 & 31 & 20.6 & 21 & 4.5 \\
            \midrule
            45 & 120 & 2578.8 & 10 & 8 & 35 & 21.4 & 21 & 5.9 \\
            45 & 170 & 3270.3 & 16 & 8 & 35 & 22.2 & 22 & 5.7 \\
            45 & 220 & 3885.4 & 29 & 8 & 32 & 22.5 & 23 & 5.2 \\
            \midrule
            60 & 120 & 2497.1 & 8  & 8 & 38 & 22.6 & 22 & 6.5 \\
            60 & 170 & 3162.3 & 13 & 8 & 38 & 23.8 & 23 & 6.5 \\
            60 & 220 & 3886.7 & 29 & 8 & 37 & 22.5 & 23 & 5.4\\
         \bottomrule
    \end{tabular}
   
    }
     \begin{tablenotes}
      \footnotesize
      \item \qquad \qquad ${RP}$: Reward function parameter
      \item \qquad \qquad ${Settings}$: Four couriers, Bomonti region, $DDQN_H^{+}$ model, Batch size: 128.
      
    \end{tablenotes}
    \label{tab:sensitivity}
  \end{table} 

We observe a significant correlation between the changes in the maximum acceptable delivery time and the percentage of the rejected orders. 
The models with lower $RP$ values learn to reject more orders to maintain shorter delivery times. 
On the other hand, in a day with an average number of orders in the system, $RP=30$ is too conservative in accepting orders. 
In that case, the threshold of 45 performs better. 

In general, we observe that if the target delivery time is relaxed, e.g., to 60 minutes, the accumulated returns diminish. 
At the same time, it increases the average delivery times and loses the capability of smart rejection. 
We observe that increased daily order counts ($\orderCountDaily$) lead to increased rejection rates, which is expected. 
Intuitively, we expect the reward to improve with $\orderCountDaily$ for a fixed target delivery time because there is a higher amount of orders to deliver (and collect rewards for). 
On the other hand, rejecting orders has associated penalties as well. 
To find the optimal threshold, the company (or delivery service) may consider the trade-off between the number of declined orders and the expected reward.
Although being too strict in accepting the orders leads to higher returns, it may reduce customer satisfaction due to unserved orders. 
Therefore, we consider $RP=45$ as a good compromise as it has an acceptable rejection rate and cumulative rewards.

\subsection{\textcolor{black}{Discussion on algorithmic complexity and convergence}}

    \textcolor{black}{Based on our MDP model formulation, in an $n \times n$ grid with $r$ restaurants, $c$ couriers, and a single depot, given the maximum $f$ simultaneous order assignments, the number of possibilities for different cases in the state-action space can be obtained as follows:}
    \begin{subequations}
    \begin{align}
        (s=\text{new order},a=\text{accept})                    & \rrw f\times (n+n-2)\times c \label{eq:1}\\ 
        (s=\text{new order},a=\text{reject})                    & \rrw f\times (n+n-2)\times c \label{eq:2} \\
        (s=\text{delivered order},a=\text{return to the depot}) & \rrw (n+n-2)\times c  \label{eq:3}\\ 
        (s=\text{delivered order},a=\text{go to a restaurant})  & \rrw r\times(n+n-2)\times c  \label{eq:4} 
    \end{align}
    \end{subequations}
    
    \textcolor{black}{These calculations can be explained as follows. The maximum distance between two points in an $n \times n$ grid is $n+n-2$. 
    If a courier has at most $f$ orders and all of them are from one corner to another, $f\times(n+n-2)$ is their maximum trip length. 
    If we have $c$ couriers in the system, whether a new order is rejected or assigned to each courier, we will have $f\times (n+n-2)\times c$ state-action pairs. 
    On the other hand, when a courier delivers the order, they may either return to the depot or move towards one of the restaurants in the grid. 
    Accordingly, Equations~\eqref{eq:3} and \eqref{eq:4} show the number of state-action pairs for these cases.}
    
    \textcolor{black}{
    Tabular model-free RL methods such as Q-learning and SARSA have convergence guarantees under certain conditions, whereas there is no theoretical convergence guarantee for deep RL methods~\citep{sutton2018reinforcement}.
    On the other hand, these tabular methods might also have slow convergence issues even for very small problem instances depending on the problem setting (e.g., Q-learning is shown to suffer from maximization bias, which is mitigated by double Q-learning~\citep{sutton2018reinforcement}).
    As such, we started by training the standard RL algorithms (e.g., Q-learning and SARSA) for our MDP formulation for a large number of episodes, however, we observed that many state-action pairs were neither visited nor updated. 
    Thus, the long-term reward of many state-action pairs could not be determined, and, in the testing phase, those algorithms demonstrate a performance even worse than the rule-based baselines. 
    Accordingly, we substitute these traditional RL algorithms with deep RL algorithms. 
    We note that many deep RL applications in different domains compared DQN variants against tabular methods, and empirically showed the superiority of the DQNs~\citep{lin2018efficient, luo2020dynamic}. 
    }
    
\begin{figure}[!ht]
        \centering
        \subfloat[DDQN$^+_H$, 2 couriers, $5\times5$ \label{fig:2c5P}]{\includegraphics[width=0.33\textwidth]{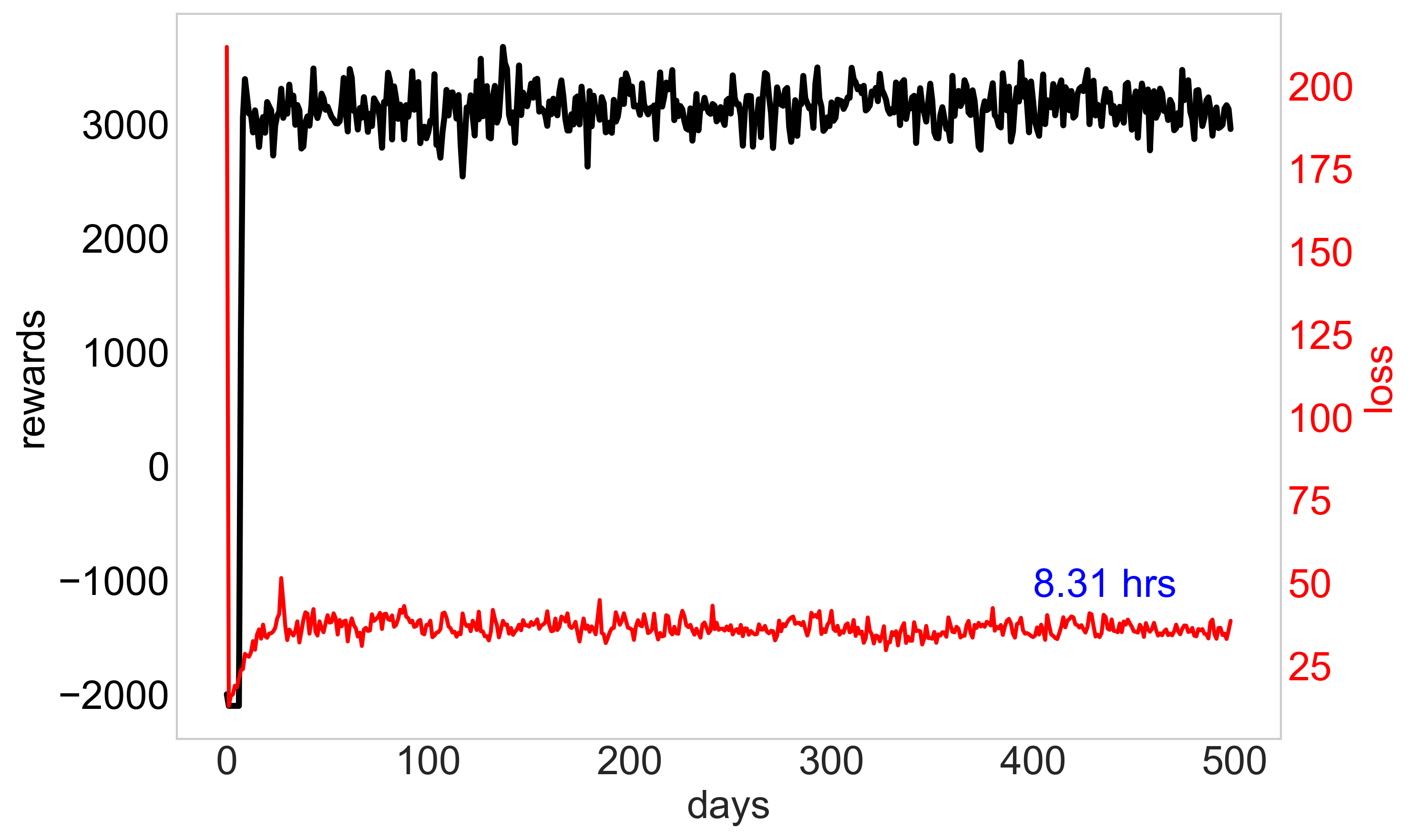}}
        \hfill
        \subfloat[DDQN$^+_H$, 2 couriers, $10\times10$ \label{fig:2c10P}]{\includegraphics[width=0.33\textwidth]{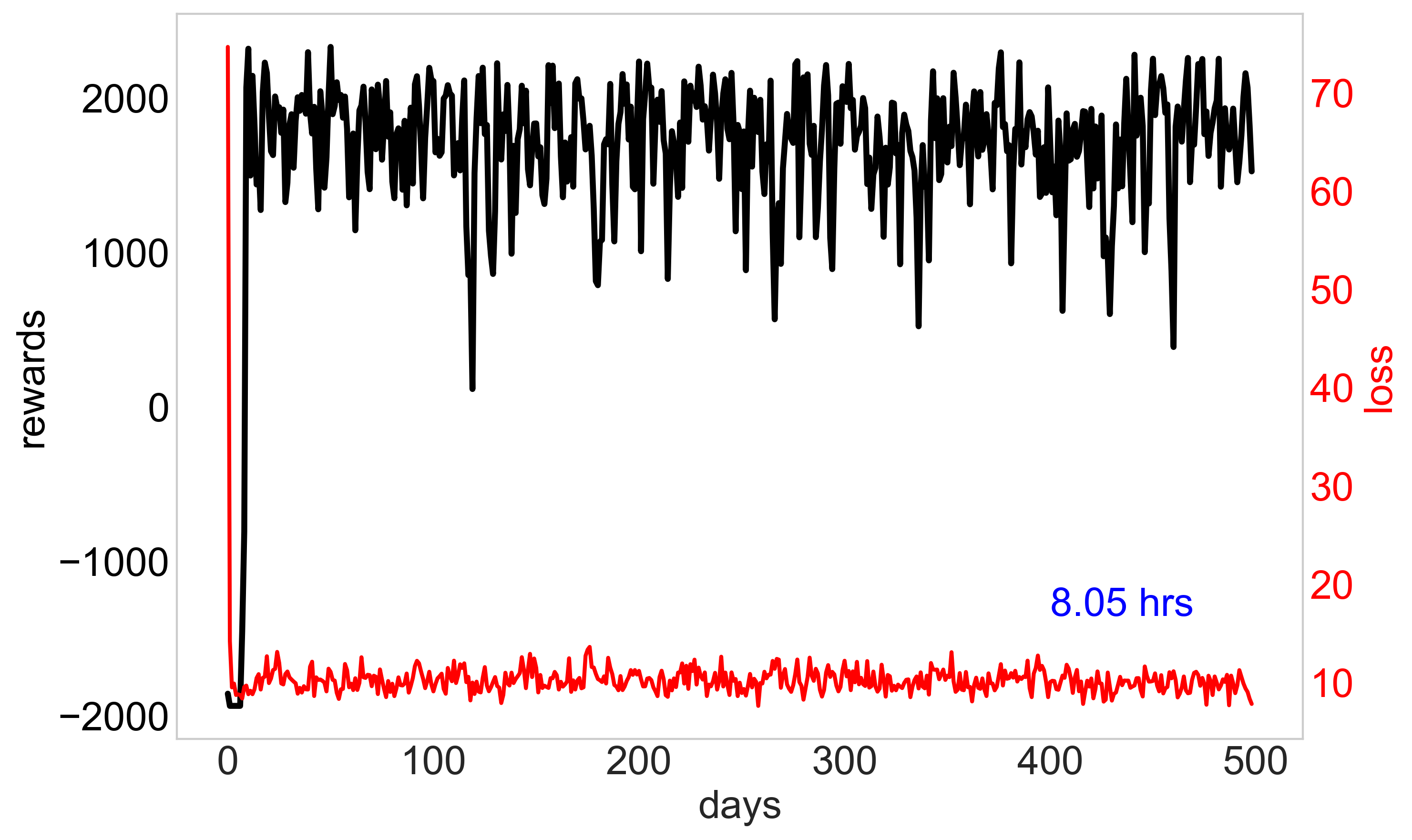}}
        \hfill
        \subfloat[DDQN$^+_H$, 2 couriers, $20\times20$ \label{fig:2c20P}]{\includegraphics[width=0.33\textwidth]{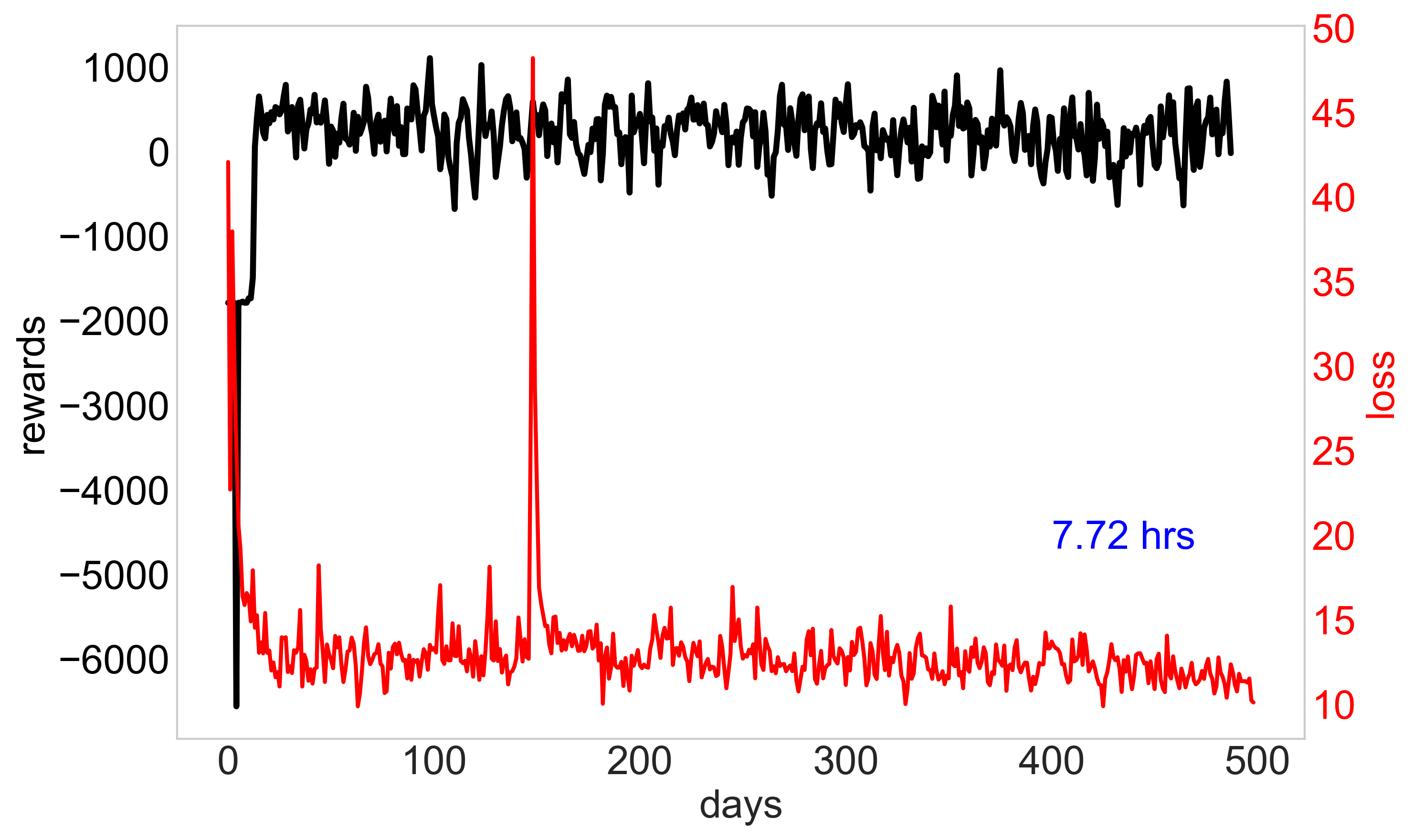}}\\
        \subfloat[DQN$_H$, 2 couriers, $5\times5$\label{fig:2c5NP}]{\includegraphics[width=0.33\textwidth]{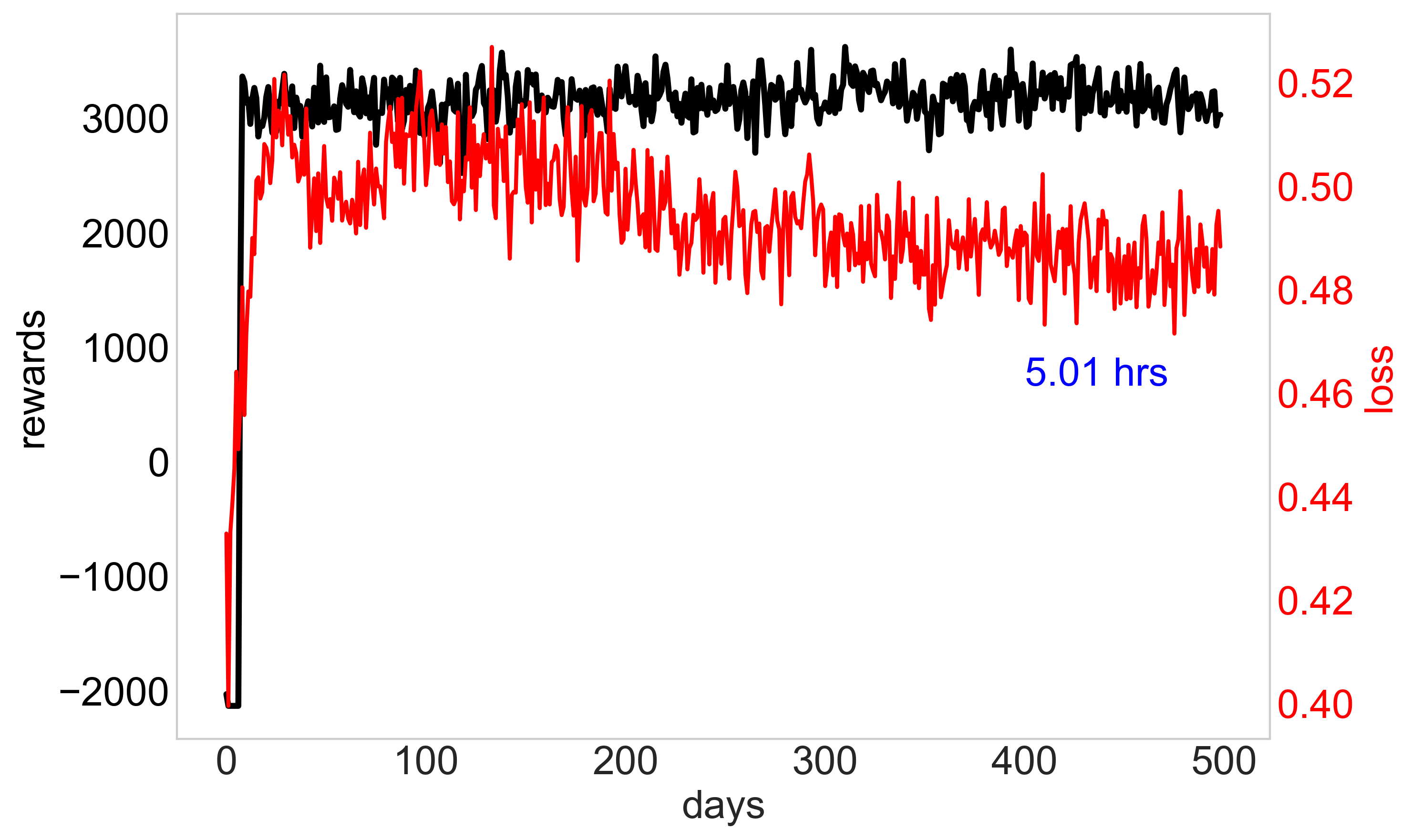}}
        \hfill
        \subfloat[DQN$_H$, 2 couriers, $10\times10$\label{fig:2c10NP}]{\includegraphics[width=0.33\textwidth]{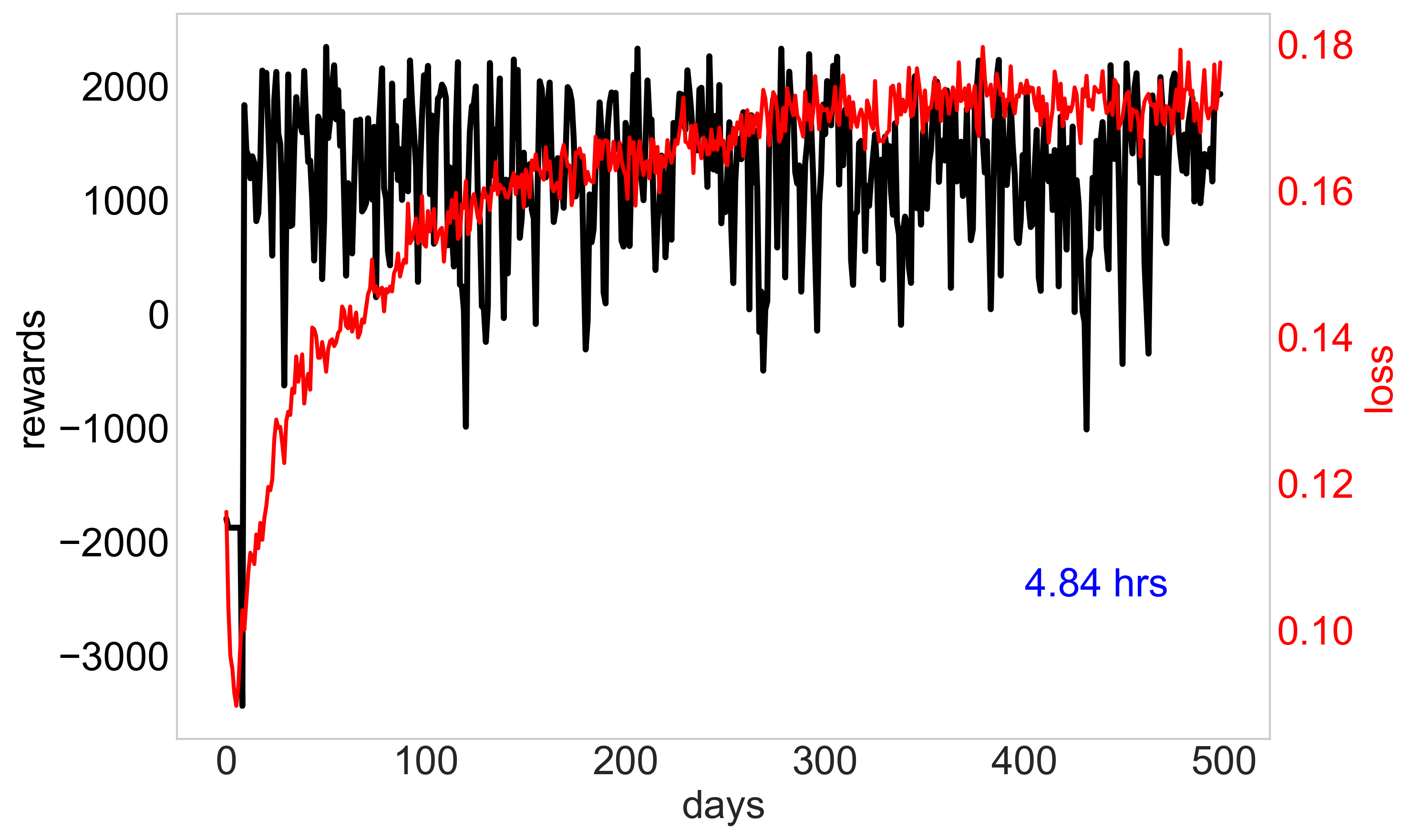}}
        \hfill
        \subfloat[DQN$_H$, 2 couriers, $20\times20$\label{fig:2c20NP}]{\includegraphics[width=0.33\textwidth]{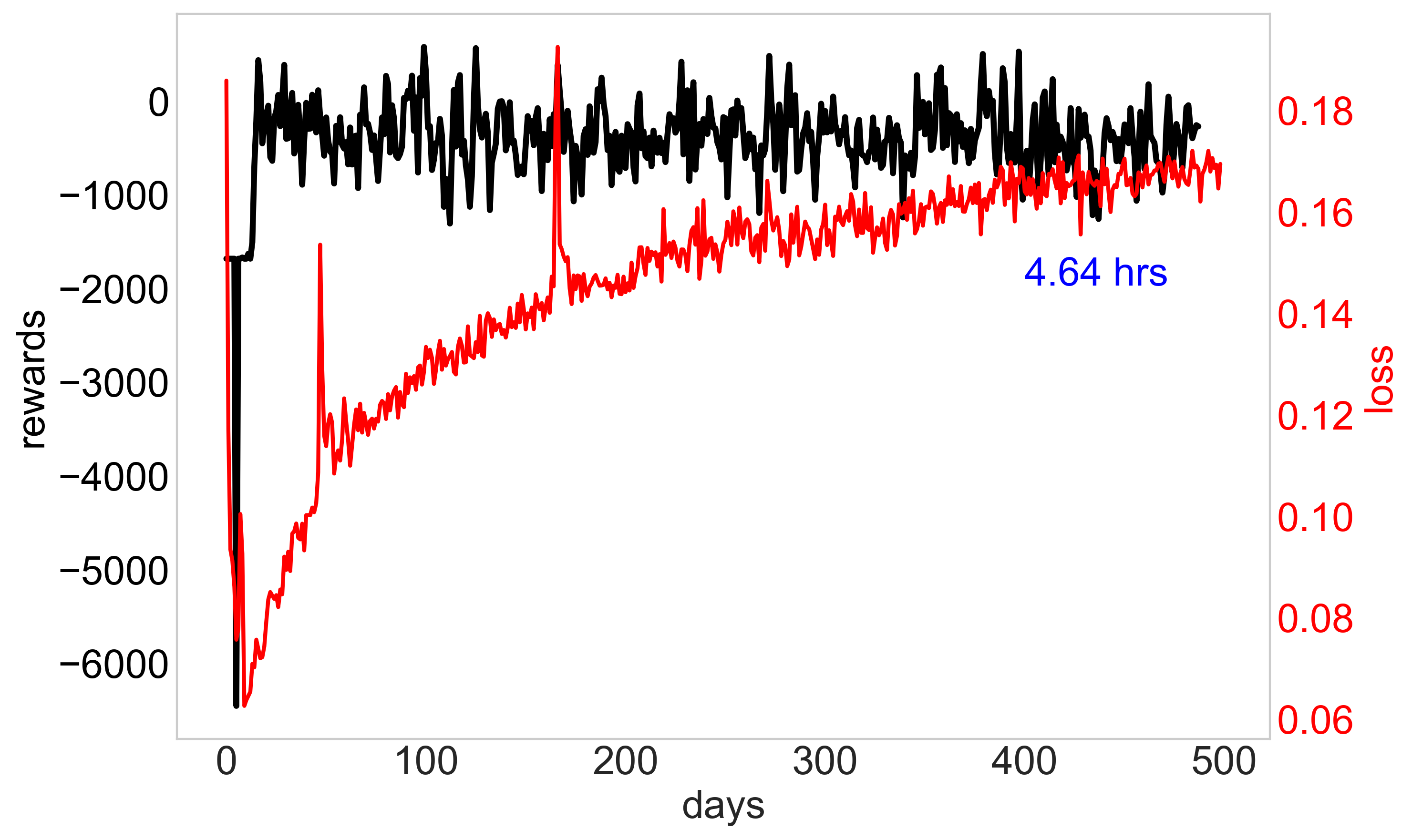}}\\
        \subfloat[DDQN$^+_H$, 5 couriers, $5\times5$ \label{fig:5c5P}]{\includegraphics[width=0.33\textwidth]{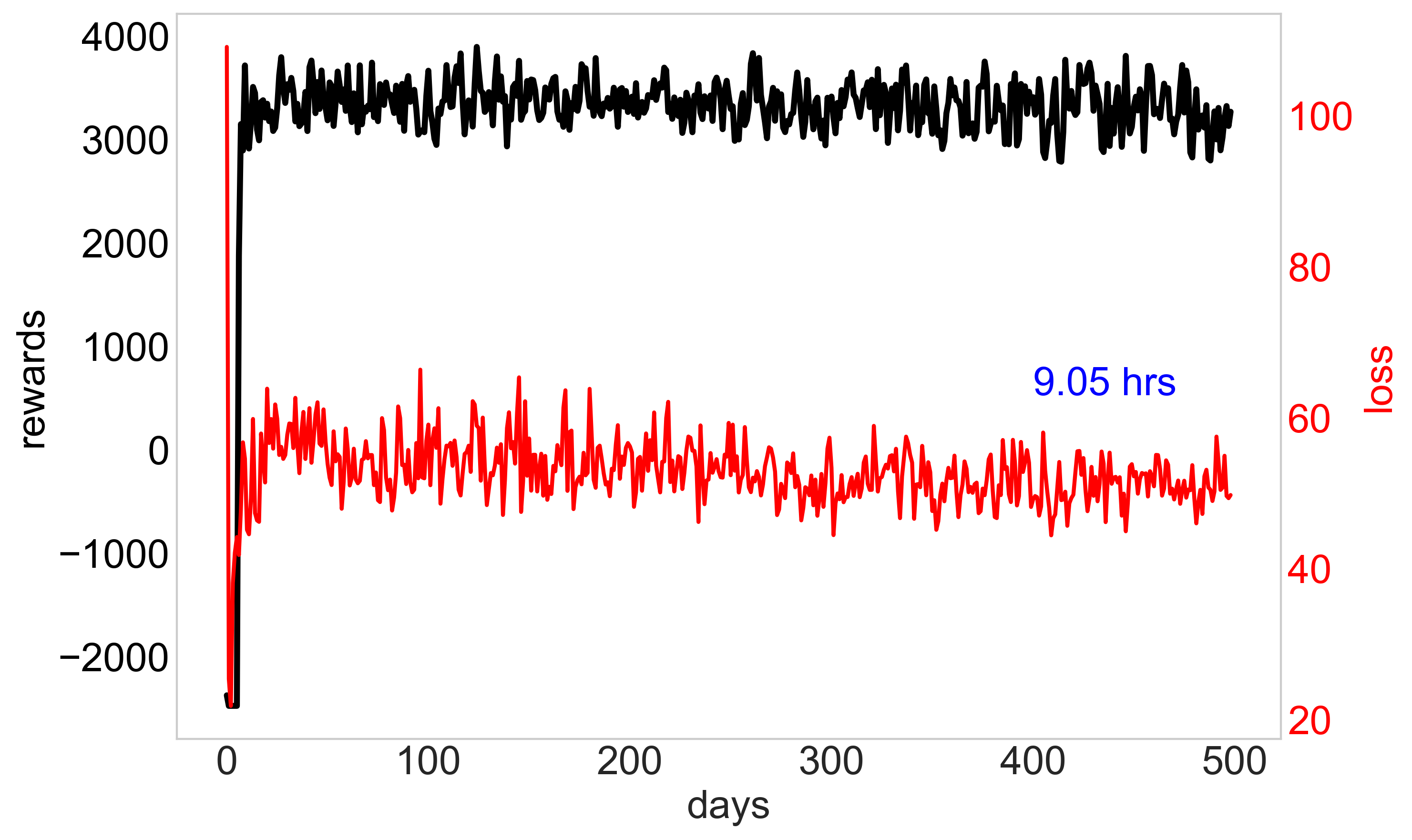}}
        \hfill
        \subfloat[DDQN$^+_H$, 5 couriers, $10\times10$ \label{fig:5c10P}]{\includegraphics[width=0.33\textwidth]{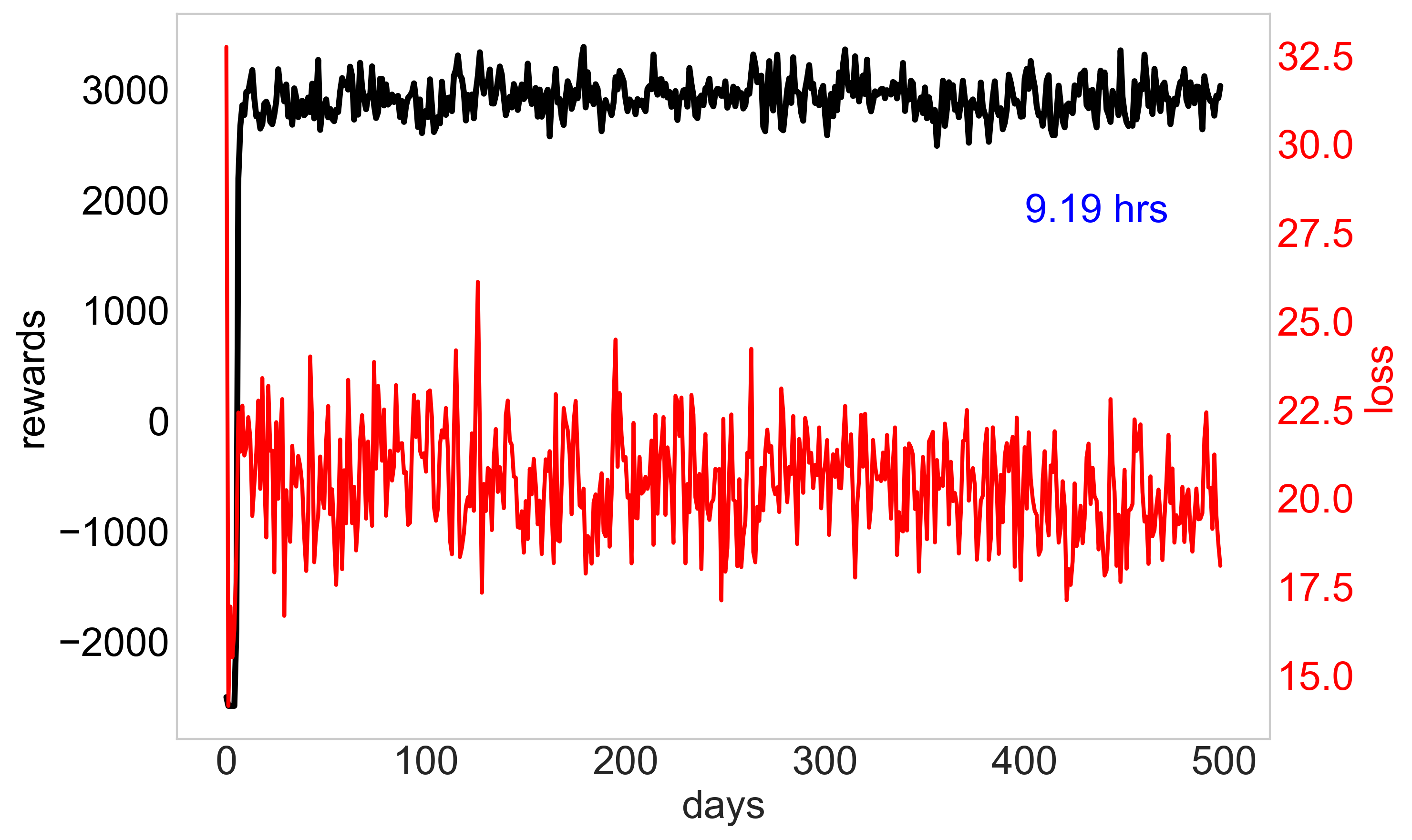}}
        \hfill
        \subfloat[DDQN$^+_H$, 5 couriers, $20\times20$ \label{fig:5c20P}]{\includegraphics[width=0.33\textwidth]{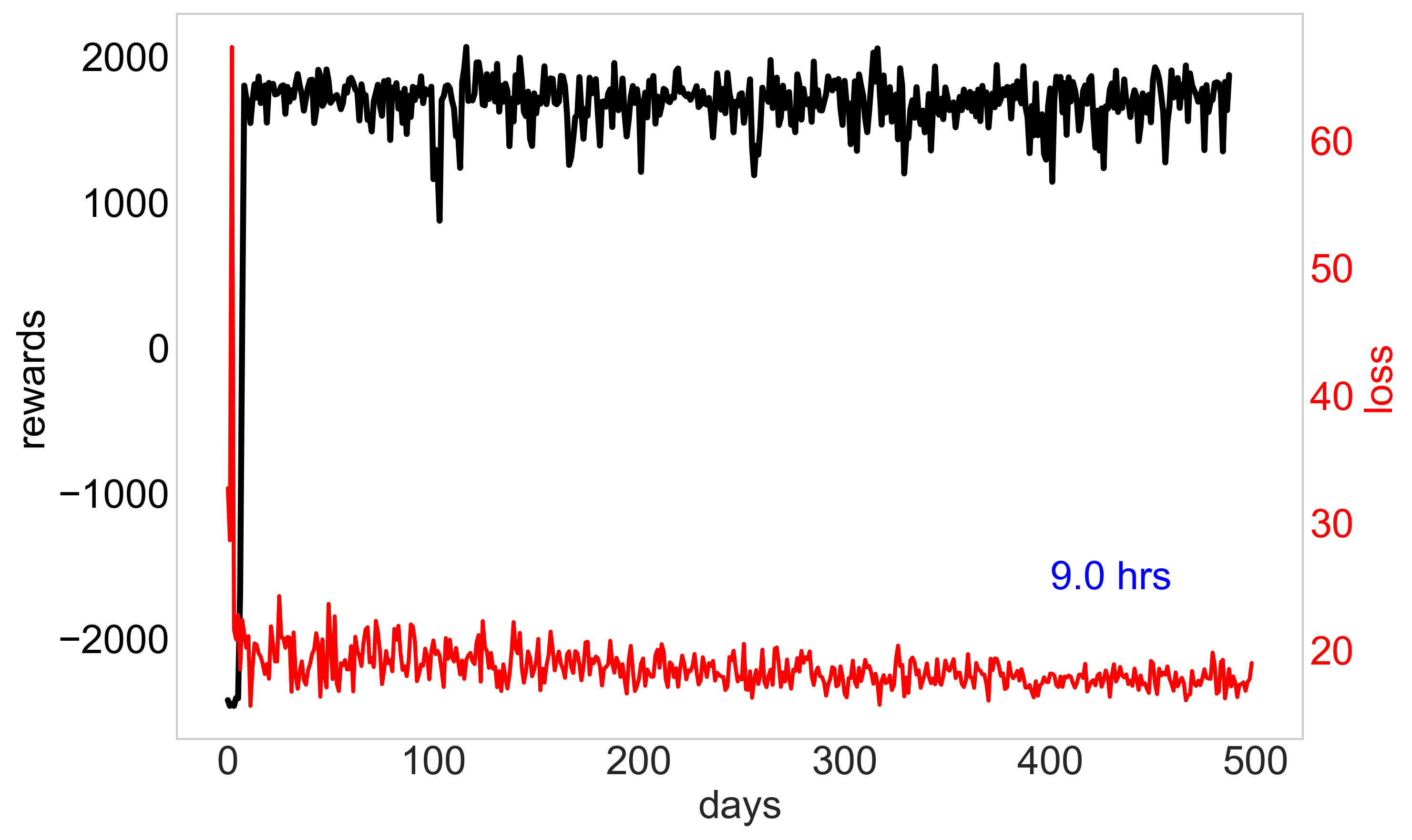}}\\
        \subfloat[DQN$_H$, 5 couriers, $5\times5$\label{fig:5c5NP}]{\includegraphics[width=0.33\textwidth]{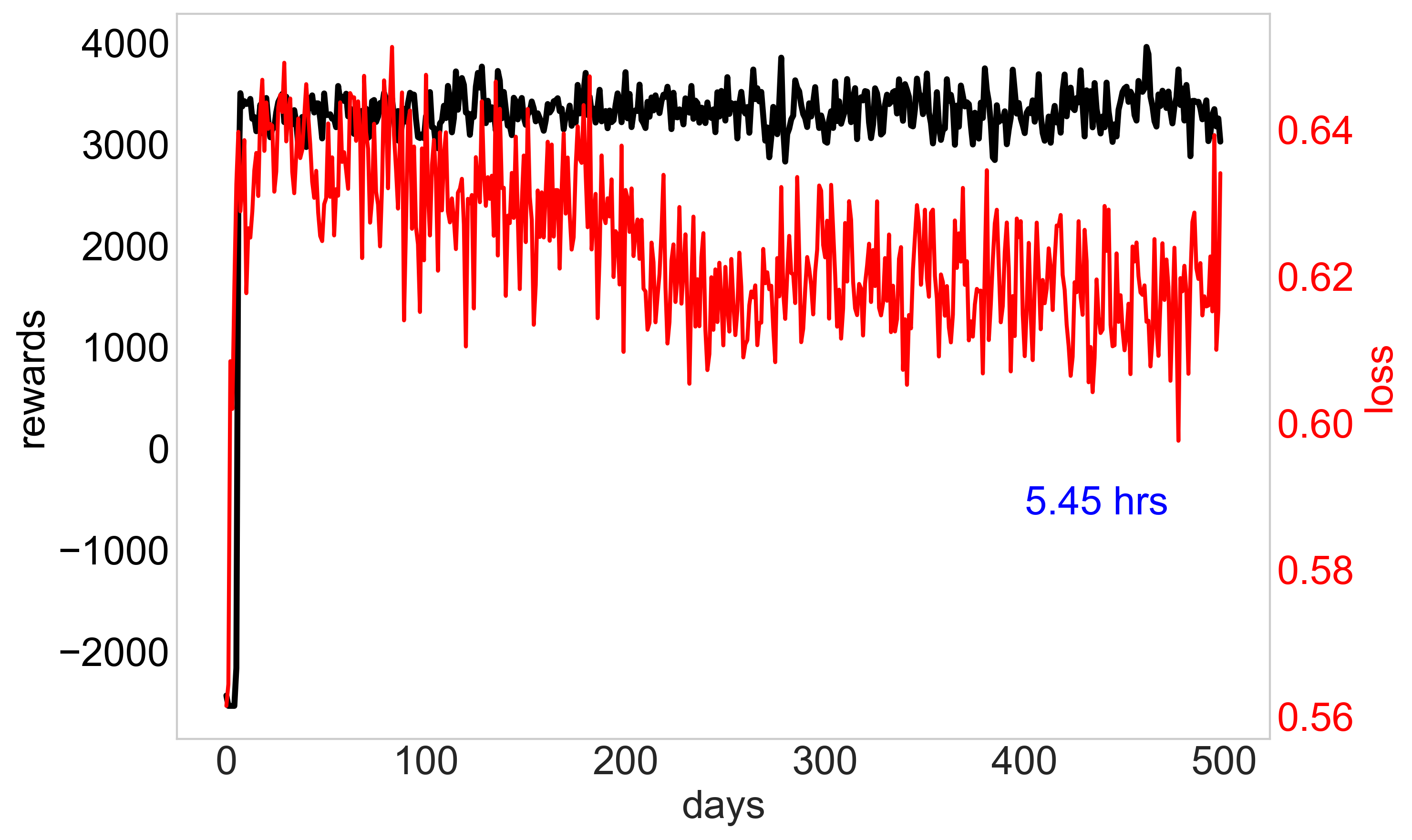}}
        \hfill
        \subfloat [DQN$_H$, 5 couriers, $10\times10$\label{fig:5c10NP}]{\includegraphics[width=0.33\textwidth]{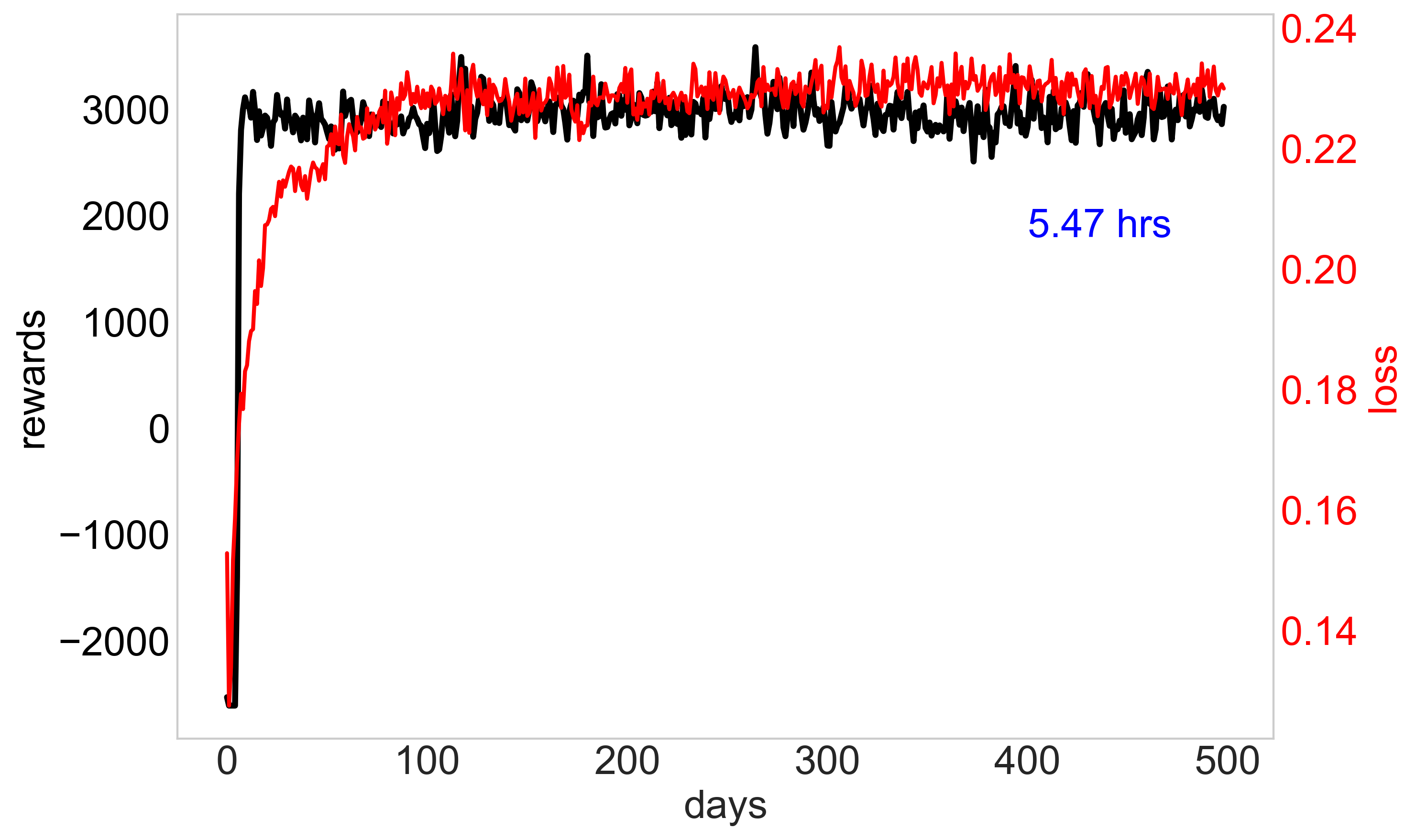}}
        \hfill
        \subfloat[DQN$_H$, 5 couriers, $20\times20$\label{fig:5c20NP}]{\includegraphics[width=0.33\textwidth]{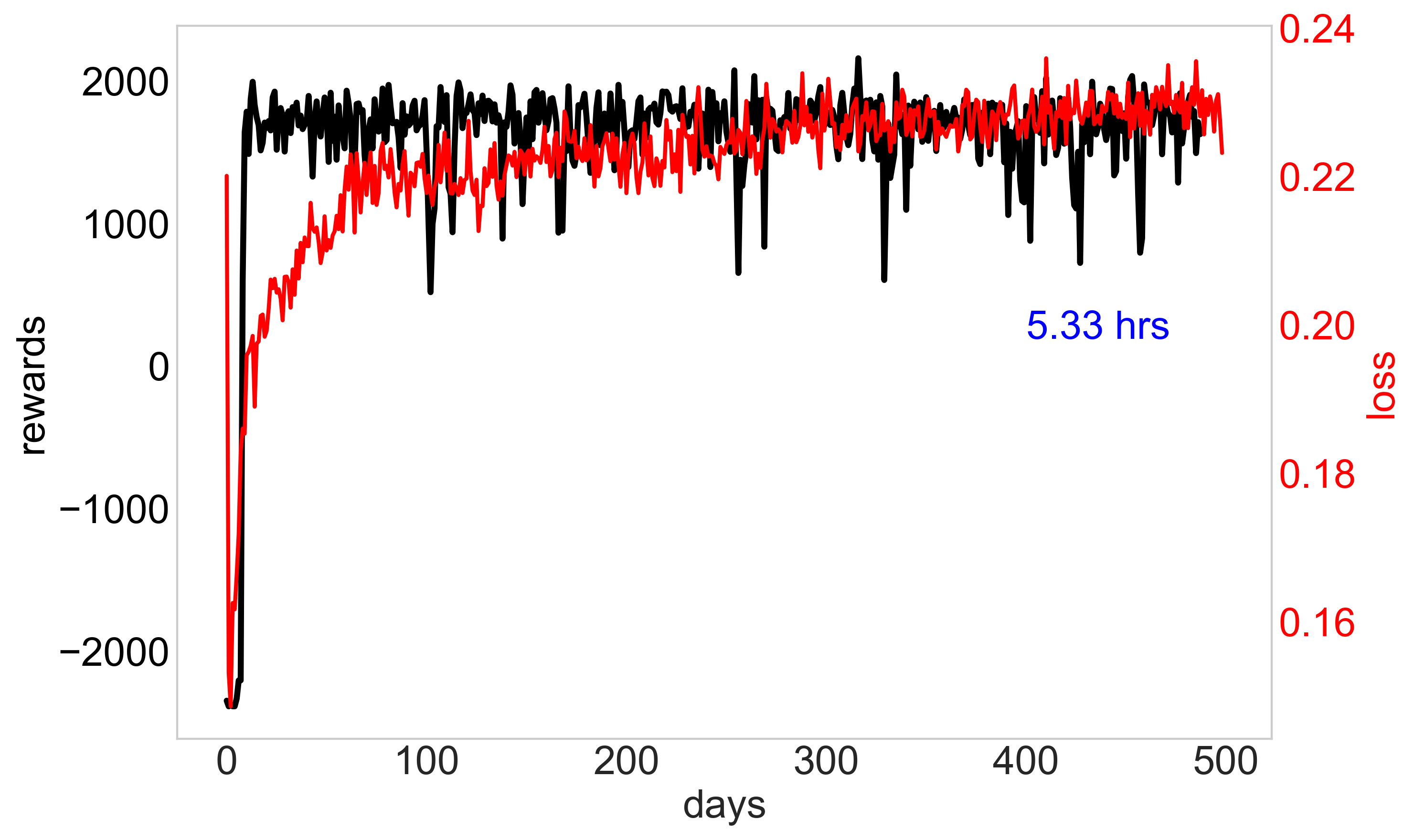}}
        \caption{{Model convergence under different settings (i.e., different numbers of couriers and grid sizes with or without dueling and prioritized experience replay). Note that y-axis ranges may differ.}}
        \label{fig: Loss_reward}
    \end{figure}

    \textcolor{black}{Previous studies consider the convergence of Q-values as an indicator for the convergence of the DQNs and the variants~\citep{tampuu2017}. 
    Accordingly, we define the convergence as a mean-squared error (MSE) between the Q-value estimates and the ground truth. 
    Additionally, we monitor the average loss of the model in the training phase. 
    Figure~\ref{fig: Loss_reward} shows the cumulative daily reward and the average daily loss of the deep learning model for different numbers of couriers and grid sizes with or without dueling and prioritized experience replay in the DQN architectures. 
    We set the same daily order distribution (120 orders), same restaurant and depot distribution, and identical algorithm settings (e.g., batch size). 
    We observe that DDQN$^+_H$ shows reasonable convergence performance. That is, in the first few days when the couriers start exploring, the model demonstrates a lower performance. 
    However, the collected reward increases and stabilizes after almost 15 days (roughly 2,300 episodes). 
    Similarly, the model fluctuates around the same loss values after a few days of training. 
    As expected, when the problem size gets larger (i.e., having either more couriers or greater grid size), we observe more fluctuation in the loss and reward values. 
    We do not observe a significant difference in the convergence behaviour while using prioritized experience (with PER). Regarding the average daily reward, having more couriers results in higher rewards as they deliver more orders and have fewer reject decisions, which impact the convergence plots.
    In some cases, we observe more fluctuations in the reward and loss values during the training phase, especially for DQN$_H$ (as can be seen by comparing Figures~\ref{fig:2c20P} and \ref{fig:2c20NP}). 
    Regarding the run times, we find that dueling and PER increase the training time by almost 66\%.
    }
    
\section{Conclusion}\label{conc}

In this paper, we have modeled a meal delivery problem in which a set of couriers deliver orders from restaurants to customers in a timely manner.
Our objective is to increase the cumulative reward in a given time interval and decrease the overall delivery times. 
\textcolor{black}{We note that this problem involves sequential decision-making under uncertainty, which requires decision making in real-time (e.g., via pre-computed policies).
Moreover, we do not have access to full characterisation of the underlying stochastic process in our meal delivery problem, hence we resorted to model-free reinforcement learning methods such as deep reinforcement learning algorithms to solve the problem.
}
We compared eight DQN extensions on the synthetic data to find the best performing algorithm for this model. 
We observed that DDQN with a prioritized experience replay with hard updates ($\text{DDQN}_H^+$) performed better than other algorithms with respect to both collected reward and the delivery times. 
Furthermore, we applied hyperparameter tuning to identify the ideal hyperparameters for this algorithm, which was then used in the subsequent experiments.

We trained this algorithm on order data of three regions of Istanbul. 
We compared the resulting policy with two baseline policies according to the collected reward and order delivery times for a different number of couriers. 
The policies generated by $\text{DDQN}_H^+$ outperform two baseline policies in terms of collected rewards and delivery times for a varying number of couriers. 
We also investigated the number of delivered orders in each hour of the day and the courier utilization for a different number of couriers. 
As we increase the number of couriers in the system, the collected reward increases, and the delivery time decreases.
However, after a certain number of couriers, the return of adding one more courier to the system decreases. 
In other words, considering the order arrival rates and the costs associated with hiring each courier, we recommend the practitioners balance the number of couriers and the order rate, as confirmed by our analysis. 
Moreover, we observe that the courier utilization decreases with an increasing number of couriers in the system, emphasizing the importance of finding the ideal number of couriers.

Although we concur that there is no best algorithm for all circumstances, we chose the one that performs the best under different conditions using a synthetic dataset.
Moreover, in terms of bias in the experiment design, we repeat the process multiple times and report the average performance of the models. 
We repeat the process for different synthetic grid sizes and different geographic regions to ensure the generalizability of the model. 
We also explore different order rates in the system to mimic both busy and slow days. 
\textcolor{black}{Regarding the delivery fees, they might differ from one location to another in a real-world scenario. 
Additionally, instead of rejecting the orders from distant locations (i.e., based on the delivery time estimates), the customers can be charged extra to fulfill the order.
However, our model assumes the feasibility of the orders to be based on their distance to the available couriers. 
We acknowledge this limitation and leave it for future research where 
dynamic delivery fees
could be incorporated into the model.}
Another noteworthy caveat in our analysis is that we exclude different characteristics of the couriers. 
For instance, one courier can be faster and more diligent in delivering orders than others, or there might be some irresponsible couriers in the system who take detours in their delivery. 
However, we assume that the delivery times for different couriers to be the same if they are assigned to an identical order. 
Given the paucity of current information on the delivery time, we used the grid approach and Manhattan distance for the shortest path calculations. 
Based on the spatial configuration of Istanbul and its network pattern, we considered it as a proper approximation of the real-world situation.

To the best of our knowledge, no other previous studies investigated the same problem settings for the meal delivery problems, making it nontrivial to directly compare our model against others'. In future work, we aim to investigate ADP-based formulations for similar meal delivery problems as they provide another suitable modeling mechanism for such problems. Accordingly, we can perform a comparative analysis between ADP and deep RL-based approaches.
Another relevant venue for future research would be to augment environmental mediators such as couriers' fatigue, break times, fuel cost, and couriers' shifts and possible overtime. 
Another consideration that merits further investigation is path updates after we assign a busy courier multiple orders. 
In the current system, although we can assign more than one order to couriers, we assume that they deliver the new order after finishing the current one(s). 
However, an important consideration in actual scenarios would be updating the path accordingly while assigning a new order. 
Future research efforts should also focus on the fair distribution of orders among couriers. 
That is, what we denote as ``courier utilization'' in this paper requires more attention in courier assignment tasks to address the equity and fairness considerations in workload management.
\textcolor{black}{Lastly, similar to other reinforcement learning approaches, we ensure satisfying various meal delivery problem constraints through our reward mechanisms.
Accordingly, the proposed model treats these constraints as soft constraints, which is the common strategy to handle constraints in large scale problems via reinforcement learning techniques.
We note that this constitutes a limitation of the proposed method when hard constraints exist in the problem setting, and constrained reinforcement learning methods (e.g., constrained MDPs) can be developed to handle the hard constraints (e.g., strict delivery duration targets) in the meal delivery problem.}

\section{Acknowledgements}
The authors would like to thank the Getir company for supporting this study and providing data and feedback throughout.
\bibliographystyle{elsarticle-harv}

\bibliography{cas-refs}


\end{document}